\PassOptionsToPackage{table}{xcolor}
\documentclass{article}
\usepackage{geometry}
\usepackage{times}

\usepackage[utf8]{inputenc} 
\usepackage[T1]{fontenc}    
\usepackage{xr-hyper}
\usepackage{hyperref}       
\usepackage{url}            
\usepackage{booktabs}       
\usepackage{amsfonts}       
\usepackage{amsmath}
\usepackage{nicefrac}       
\usepackage{microtype}      
\usepackage{natbib}
\usepackage{graphicx}
\usepackage{enumitem}
\usepackage{diagbox}
\usepackage{soul}
\usepackage{adjustbox}
\usepackage{svg}
\usepackage{wrapfig}
\usepackage{tabularx}
\usepackage[position=above]{caption}
\usepackage{subcaption}

\usepackage{floatrow}
\usepackage[title]{appendix}

\definecolor{lightgray}{gray}{0.94}
\let\oldtabular\tabular
\let\endoldtabular\endtabular
\renewenvironment{tabular}{\rowcolors{2}{lightgray}{white}\oldtabular}{\endoldtabular}

\newfloatcommand{capbtabbox}{table}[][\FBwidth]
\makeatletter
\newcommand*{\addFileDependency}[1]{
  \typeout{(#1)}
  \@addtofilelist{#1}
  \IfFileExists{#1}{}{\typeout{No file #1.}}
}
\makeatother

\newcommand*{\myexternaldocument}[1]{%
    \externaldocument{#1}%
    \addFileDependency{#1.tex}%
    \addFileDependency{#1.aux}%
}

\myexternaldocument{appendix}
\usepackage{pdfpages}
\title{The Robustness Limits of SoTA Vision Models to Natural Variation}

\author{Mark Ibrahim$^1$, Quentin Garrido$^{1,2}$, Ari Morcos$^{1*}$, Diane Bouchacourt$^{1*}$}
\date{$^1$Fundamental AI Research (FAIR), Meta AI\\
$^2$Univ Gustave Eiffel, CNRS, LIGM, F-77454 Marne-la-Vallée, France}

\begin{document}

\def\thefootnote{*}\footnotetext{equal contribution}\def\thefootnote{\arabic{footnote}}
\maketitle

\begin{abstract}
Recent state-of-the-art vision models have introduced new architectures, learning paradigms, and larger pretraining data, leading to impressive performance on tasks such as classification.
While previous generations of vision models were shown to lack robustness to factors such as pose, the extent to which this next generation of models are more robust remains unclear.
To study this question, we develop a dataset of more than 7 million images with controlled changes in pose, position background, lighting color, and size.
We study not only how robust recent state-of-the-art models are, but also the extent to which models can generalize to
variation in each of these factors.
We consider a catalog of recent vision models, including vision transformers (ViT), self-supervised models such as masked autoencoders (MAE), and 
models trained on larger datasets such as CLIP.
We find that even today's best models are not robust to common changes in pose, size, and background.
When some samples varied during training, we found models required a significant portion of instances seen varying to generalize---though eventually robustness did improve.
When variability is only witnessed for some classes however, we found that models did not generalize to other classes unless the classes were very similar to those seen varying during training. 
We hope our work will shed further light on the blind spots of SoTA models and spur the development of more robust vision models.

\end{abstract}

\section{Introduction}

\begin{figure}
    \centering
    \includegraphics[width=\textwidth]{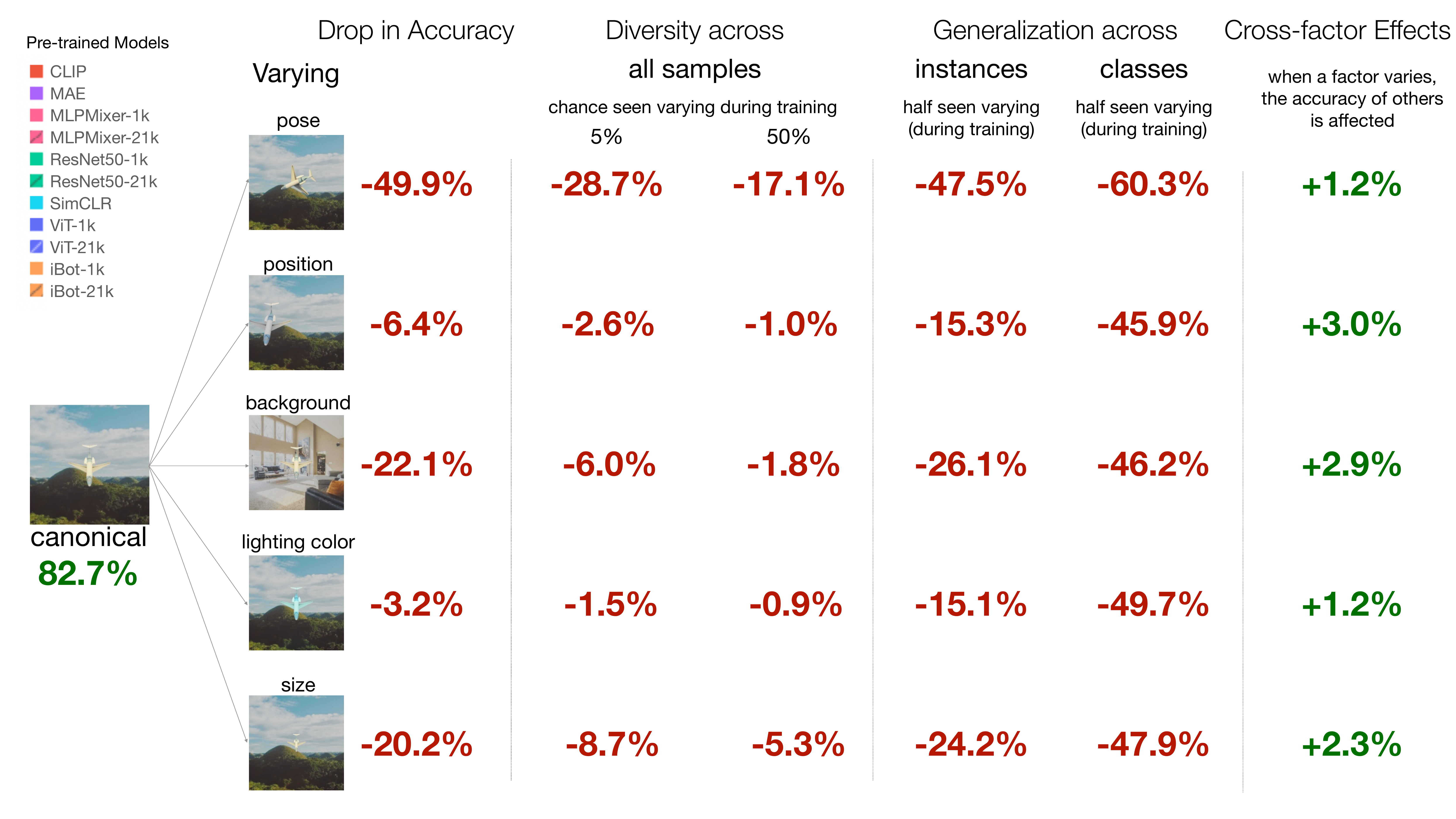}
    \caption{\textbf{SoTA models are not robust to and struggle to generalize common variations in pose, background, size}.
    We show the average drop in accuracy across models when we vary each factor. 
    We find if we vary all samples during training, models require a significant portion of variation ($\geq 50\%$) to close the robustness gaps. When variation is only seen for some classes or instances, models struggle 
    to generalize variation across instances or classes. Finally, when a factor varies during training the robustness of other factors is also affected.
    }
    \label{fig:fig1}
\end{figure}

A dataset of natural images can be described by a set of factors of variations which characterize the main axes along which samples sample vary; for example pose, position, illumination, size, etc \citep{bengio_representation_2013,bouchacourt_grounding_2021}. Importantly, test-time unseen data samples may exhibit different variability across factors than those seen during training \citep{Quinonero-Candela2009}. It is thus desirable for state-of-the-art (SoTA) models to be robust to changes in these factors \citep{bengio_representation_2013}. However, previous work has shown that vision models such as Convolutional Neural Networks (CNNs) or Vision Transformers (ViTs; \citet{dosovitskiy_image_2021}) are quite brittle to changes in pose, illumination, or even slight rotations and translation transformations \citep{engstrom_2019, alcorn_strike_2019, Abbas2022}. Yet, much of the existing work focuses either on the effect of a single transformation or analyzes toy settings where variability can be controlled. If we aim to deploy models in more realistic and challenging applications, however, we need to study their brittleness to more natural variations on more realistic data which can potentially appear together (e.g. multiple factors at the same time). \\
\\
Here, we extend existing work to study models’ susceptibility to changes in position, size, spot hue, background, and pose independently, as well as \emph{changes in all factors in conjunction}. To do so, we develop a dataset allowing based on 3d warehouse objects~\citep{3dwarehouse} that we place in non-uniform backgrounds and for which we vary the aforementioned factors. Using the typical evaluation procedures of self-supervised models \citep{caron_unsupervised_2021,dosovitskiy_image_2021,chen_simple_2020} (finetuning and linear evaluation), we examine robustness across a catalog of state-of-the-art vision architectures, such as CLIP \citep{radford2021learning} that have significantly outperformed earlier models on robustness benchmarks such as ObjectNet \citep{NEURIPS2019_97af07a1}, Masked AutoEncoders (MAE, \citep{he2022masked}), or ViTs \citet{dosovitskiy_image_2021}) among others. This allows us to compare common inductive biases such as architectures, training paradigm or the amount of pre-training data. Furthermore, we examine the effect of \emph{variability} for the factors, that is (i) seeing some instances varying for a \emph{single factor affects the other factors} and how (ii) seeing some \emph{some classes varying for factors affect other classes}. To the best of our knowledge, generalization of robustness across classes has not previously been studied.
\\
Our main findings and contributions, summarized in Figure \ref{fig:fig1},  are:
\begin{enumerate}
    \item We study the robustness of a wide range of SoTA models to variations in naturally occurring factors, examining single-factor and all-factors variations. In general, we found that \textbf{SoTA pre-trained models fine-tuned with little or no variability are not robust to factor variations} (Section \ref{sec:not_robust}).
    \item We compare the effect of different inductive biases as realized through different architectures, training paradigms, quantity of pre-training data, and finetuning vs. linear evaluation. We found that differences in \textbf{architecture and training paradigm have minor impacts on robustness, but that more training data helps} and that finetuning generally leads to worse robustness (Section \ref{sec:not_robust}).  
      \item \textbf{Increasing the amount of variability of all instances for each factor during training helps generalization} (Section \ref{sec:singlefactorimages}). However, increasing variability only for some instances can hurt if not enough variability is introduced (Section \ref{sec:singlefactorinstance}). Nonetheless, \textbf{variability in single factors tends to improve robustness to other factors too} (Section \ref{sec:crossfactoreffects}).  
    \item By studying the effect of variability across classes, we find that if a \textbf{class is seen varying for some factors during training, it helps to generalize to very similar classes that were not encountered varying, but generalizes worse to all classes that are even little dissimilar} and much worse to those classes which are highly dissimilar (Section \ref{sec:cross-class}).
\end{enumerate}

\section{Related work}
Recently, there have been much interesting work studying models' brittleness. Our work falls in this body of literature, yet we aim to provide a more extensive analysis by (i) varying different factors alone and in combination (ii) studying the effect of different amounts of variability seen in the training data (iii) on a extended list of standard vision models. \citet{alcorn_strike_2019} focuses on models' robustness to pose changes, while \citet{engstrom_2019} studies rotation and translation changes (both together and alone) and find that explicitly augmenting the data with such variations does not fix the problem, a conclusion shared by \citep{Azulay19} who study small image changes e.g. translating / scaling. \citet{Madan2021} also find that ResNets and CLIP networks are brittle to pose and lighting changes. \citet{madan_when_2021} investigate the generalization of only CNNs to combinations of two factors (object category and 3D viewpoint), finding that increasing the number of combinations seen at training helps generalization and also that separate networks outperform shared ones.

Some works explicitly study the invariance of models, e.g. \citet{lenc18understanding,bouchacourt_grounding_2021} where the latter also found that data augmentation does not bring the expected invariance. A similar conclusion was also drawn in~\citep{bordes2021high} where both supervised and self-supervised representations were found to not be invariant to the training augmentations. In~\citep{von_kugelgen_self-supervised_2021} the preservation of natural transformation information was studied in a self-supervised setting, where they found that the pretraining data augmentation policy plays an important role. \\
\\
Perhaps closest to our work, \citep{Abbas2022} studied the sensitivity of a large body of models and a variety of effects (architectures, data augmentation, dataset modalities), albeit only to pose changes (orientation and scale). 
In this work, we extend the set of factors studied, including cross-factors effects as well as a larger catalog of more recent SoTA models and architectures.

\section{Methods}

\subsection{The dataset suited to perform robustness analysis}

To study the brittleness of state-of-the-art models with respect to data factors of variation, several data properties are desirable. First, while there exist variants of real image datasets that present naturally occurring variations \citep{hendrycks_natural_2021}, control over the data generation process allows detailed factor metadata. Second, we want the dataset to vary sufficiently for us to draw consistent conclusions. Finally, we want the images to remain as close to realistic images in terms of image quality as possible. Existing datasets developed to show robustness of models often vary in just one or a few factors, and the images are not really realistic or span only a few classes (e.g. Shapes3D \citep{kim2018factorvae}, MPI3D \citep{gondal2019mpi3d}, dSprites \citep{higgins2017betavae} among others). Therefore we develop and release our own dataset based on 3d Warehouse~\citep{3dwarehouse} objects that we place in non-uniform backgrounds.

We use $54$ synsets from 3d Warehouse~\citep{3dwarehouse}, and  $50$ objects for each synset. For the first $4$ scalar factors (position, pose, size, lighting color), we use equally spaced scalar values. For the background, we use $5$ background types (\emph{sky, water, city, home, grass}) and $5$ different backgrounds per type, with natural images coming from~\cite{li2022bridging}. We define for each factor a canonical value, that is, the most represented value for that factor, to mimic the fact that in natural images we often see objects in a set of given factors (e.g. their upright position). We then vary these factors in three different manners: (i) each factor independently ($101$ scalar values equally spaced for scalar factors + $25$ backgrounds) (ii) factors varying in pairs ($11$ scalar values + $10$ backgrounds) (iii) all factors varying together (drawing $1000$ random combinations from the full grid of $11$ equally spaced values and $10$ backgrounds). This gives us roughly $7$ million (M) images in total, divided as follows: single factor (1.1M), paired factors (3.1M), and all factors (2.7M).

\subsection{Introduce pre-trained models}

We select a set of state-of-the-art (SoTA) vision models, with many achieving > $80\%$ top-1 accuracy on ImageNet, spanning learning paradigms, training dataset sizes, and architectures. We also include CLIP, a model trained with caption supervision on over 400M text-pair images 
that has show impressive performance on several OoD benchmarks \cite{radford2021learning}. We also evaluate the zero-shot performance of CLIP trained on 2B images from the LAION dataset \cite{ilharco_gabriel_2021_5143773}.

We select SoTA supervised models of varying architectures. For ResNet-50, a CNN-based model, we use a ImageNet-1k pre-trained model based on the an improved training recipe from \cite{wightman2021resnet} achieving $80.4\%$ top-1 accuracy on ImageNet and an ImageNet-21k weights from \cite{ridnik2021imagenet21k} achieving $82.0\%$ top-1 accuracy on ImageNet. For Vision Transformer (ViT), an attention-based model, we use an ImageNet-21k pre-trained ViT-B/16 achieving $83.97 \%$ and ImageNet-1k pretrained weights from \cite{ridnik2021imagenet21k}. For MLPMixer, a multi-layer percetron-based model,  we use ImageNet-21k pretrained weights from \cite{ridnik2021imagenet21k}
and ImageNet-1k weights from \cite{rw2019timm} using Base-16 architecture.

We also select several SoTA self-supervised learning models. 
For SimCLR \citep{chen_simple_2020}, a contrastive learning method, we select a ResNet-50 (CNN-based) backbone, trained on ImageNet-1k 
based on weights from \cite{falcon2020framework}. 
For MAE \citet{he2022masked}, a method based on a reconstruction objective, we select an attention-based ViT encoder. We use pre-trained weights from the official repo of \cite{he2022masked}. For iBot \citet{zhou2021ibot}, also a ViT-based model, we use ImageNet-1k and ImageNet-21k pre-trained weights from the official repo of \cite{zhou2021ibot}.

\section{SoTA vision models are not robust to natural variations} \label{sec:not_robust}

\begin{table}
\centering
\caption{\textbf{SoTA models are not robust to common factors}: we show the drop in accuracy relative to each model's held-out accuracy when an object is presented in its canonical setting for linear eval (a) and finetuning (b).
We notice especially large gaps for pose, background, and size factors.}
\label{tab:canonical}
\begin{subtable}[c]{\textwidth}
\subcaption{Linear evaluation gaps}
\centering
\label{linear eval_canonical}
\begin{adjustbox}{width=\textwidth}
\begin{tabular}{@{}lllllllll@{}}
\toprule
        &  Train accuracy &  Held-out accuracy &  Pose gap &  Background gap &  Size gap &  Position gap &  Lighting color gap &  Average gap \\
    \midrule
        CLIP &           80.65 &              72.22 &    -42.40 &          -25.43 &    -19.84 &         -5.70 &               -2.65 &       -19.20 \\
         MAE &           30.12 &              21.11 &    -13.77 &          -10.77 &     -6.79 &         -2.30 &               -2.36 &        -7.20 \\
  MLPMixer1k &           85.55 &              71.56 &    -44.19 &          -35.11 &    -26.31 &        -10.59 &               -4.92 &       -24.22 \\
 MLPMixer21k &           91.59 &              80.37 &    -43.25 &          -26.83 &    -20.32 &         -5.45 &               -1.53 &       -19.48 \\
 ResNet50-1k &           86.76 &              77.41 &    -42.30 &          -27.78 &    -25.28 &         -5.10 &               -3.33 &       -20.76 \\
ResNet50-21k &           92.89 &              76.22 &    -35.49 &          -23.20 &    -14.85 &         -0.51 &               -1.04 &       -15.02 \\
      SimCLR &           91.69 &              73.33 &    -51.05 &          -33.93 &    -28.01 &         -5.85 &                1.11 &       -23.55 \\
      ViT-1k &           93.47 &              79.63 &    -44.48 &          -24.43 &    -25.05 &         -6.53 &               -2.56 &       -20.61 \\
     ViT-21k &           91.82 &              78.89 &    -39.87 &          -20.38 &    -26.73 &         -6.97 &               -0.89 &       -18.97 \\
     iBot-1k &           93.75 &              81.11 &    -52.63 &          -28.38 &    -25.96 &         -7.54 &               -3.62 &       -23.63 \\
    iBot-21k &           93.60 &              82.96 &    -52.90 &          -31.66 &    -30.46 &         -7.14 &               -1.06 &       -24.64 \\
     Average &           84.72 &              72.26 &    -42.03 &          -26.17 &    -22.69 &         -5.79 &               -2.08 &       -19.75 \\
\bottomrule
\end{tabular}
 \end{adjustbox}
\end{subtable}

\begin{subtable}[c]{\textwidth}
\subcaption{Finetuning gaps}
\label{finetuning_canonical}
\begin{adjustbox}{width=\textwidth}
\begin{tabular}{@{}lllllllll@{}}
\toprule
        &  Train accuracy &  Held-out accuracy &  Pose gap &  Background gap &  Size gap &  Position gap &  Lighting color gap &  Average gap \\
\midrule
        CLIP &           94.36 &              81.85 &    -50.84 &          -16.67 &    -16.63 &         -5.32 &               -1.96 &       -18.28 \\
         MAE &           92.91 &              73.33 &    -50.50 &          -44.73 &    -24.74 &        -17.71 &              -14.62 &       -30.46 \\
  MLPMixer1k &           90.73 &              80.37 &    -51.10 &          -25.76 &    -23.13 &         -7.08 &               -3.17 &       -22.05 \\
 MLPMixer21k &           96.21 &              84.44 &    -46.44 &          -15.67 &    -16.53 &         -5.06 &               -2.48 &       -17.24 \\
 ResNet50-1k &           95.61 &              80.96 &    -50.63 &          -18.41 &    -20.64 &         -6.66 &               -4.06 &       -20.08 \\
ResNet50-21k &           95.76 &              86.67 &    -46.68 &          -29.87 &    -19.54 &         -3.74 &               -2.35 &       -20.44 \\
      SimCLR &           95.34 &              82.96 &    -56.38 &          -30.11 &    -24.36 &         -8.30 &               -0.72 &       -23.98 \\
      ViT-1k &           96.17 &              84.44 &    -46.61 &          -14.36 &    -16.98 &         -3.34 &               -1.66 &       -16.59 \\
     ViT-21k &           96.01 &              84.44 &    -46.95 &           -9.83 &    -18.01 &         -2.78 &               -0.19 &       -15.55 \\
     iBot-1k &           94.56 &              84.81 &    -53.22 &          -25.57 &    -23.99 &         -5.82 &               -3.01 &       -22.32 \\
    iBot-21k &           95.70 &              85.56 &    -50.55 &          -12.02 &    -17.64 &         -4.1 &               -1.18 &       -17.10 \\
     Average &           94.85 &              82.71 &    -49.99 &          -22.09 &    -20.20 &         -6.36 &               -3.22 &       -20.37 \\
\bottomrule
\end{tabular}
 \end{adjustbox}
 \end{subtable}
\end{table}

We evaluate pretrained model's ability to generalize natural variation using two common protocols: linear evaluation and finetuning. We measure models' generalization by each model's classification accuracy for ``canonical'' settings and the same objects varying by one or more of the natural factors. Note that the canonical value of each factor is chosen arbitrarily, but fixed across all experiments such that the canonical value is simply the value which is dominant in the training data. We then evaluate these models on held-out objects which have factor values not seen in training, varying the values of one factor at a time. To control for differences in the performance of models on canonical data, we report the gap between the model's accuracy on the canonical and varying held-out sets. 
\paragraph{SoTA models are not robust to changes in pose, background, and size}
While models reached strong performance on canonical data, Table \ref{tab:canonical} demonstrates that even SoTA models suffer considerable drops in performance when objects vary across factors. Models were particularly sensitive to changes in pose, background, and size, while models were largely robust to changes in position and lighting color. We hypothesize that this difference in robustness across factors may be related to how easily variation across a factor can be approximated by pixel-level augmentations. Both lighting color and position can be well approximated by color shift and translation, respectively. In contrast, pose, background and size (relative to a fixed background) all require 3D manipulation of the object itself, and are therefore very difficult to replicate with pixel-level augmentations. 

While finetuning consistently improved performance on canonical data (finetuned held-out canonical accuracy of 82.71\% vs. 72.26\% for linear), it actually hurt robustness relative to linear evaluation.  Performance gaps on varying held-out instances \textit{increased} after finetuning, demonstrating that while finetuning can improve in-distribution performance, it does so at the cost of generalization (Table \ref{tab:canonical}).


\subsection{Do architectural inductive biases matter?}


\paragraph{Learning objective is more impactful than architecture for robustness} In general, we found that robustness was similar across models with the notable exception of MAE. As shown in Table \ref{tab:canonical} (b), the MAE model is especially susceptible to changes in background, with a-44.7\% drop compared to an average -19.4\% for other models. MAE is also substantially more sensitive to position and lighting color. This sensitivity was not observed in other ViT based models, suggesting that it stems from differences in the training objective rather than the architecture. While all other models use either supervised or InfoNCE based objectives, MAE uses a reconstruction objective. This focus on reconstruction may cause the model to pay closer attention to factors like background, position, and lighting color, as it is likely necessary to learn these correlations to effectively reconstruct. 

Interestingly, the consistency across architectures also largely held for comparisons between CNNs and ViT based models, even for factors such as position (translation) for which CNNs are widely believed to be robust, although several recent works have suggested otherwise \citep{Kayhan_2020_CVPR, liu_2018, bouchacourt_grounding_2021, biscione_jmlr_2021, ruderman2018pooling, zhang2019shiftinvar}. We also found comparable  gaps when evaluating CLIP using zero-shot classification, including CLIP trained on 2B LAION images (see Appendix \ref{clip_zero_shot}).

\subsection{Are self-supervised models more robust?}

\begin{figure}
         \includegraphics[width=\textwidth]{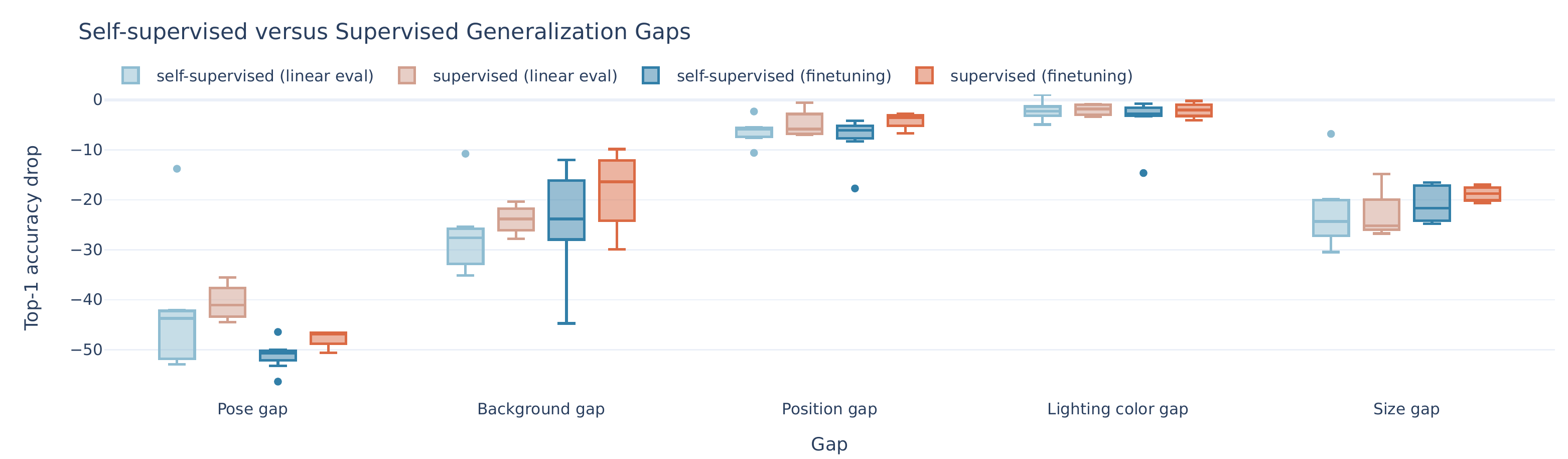}
     \caption{\textbf{Supervised models benefit more from finetuning than self-supervised models:} we compare generalization gaps 
     for self-supervised and supervised models using box-plots.}
     \label{fig:ssl_gaps}
\end{figure}

Several recent works have suggested that pre-training with self-supervision may lead to increased robustness \citep{hendrycks_ssl_robustness_2019, geirhos_2020_surprising_sim}. To test this, we compared the robustness of self-supervised models to supervised models in Figure \ref{fig:ssl_gaps}. For linear evaluation, supervised models slightly outperformed SSL models on average, though SSL models were able to achieve a higher ceiling. In the finetuning setting, however, this difference is far more striking, suggesting that the robustness of supervised models benefits far more from finetuning than SSL models. Previous works \citep{fan2021when, Chen_2020_CVPR} noted that regular finetuning of the full network weights does not preserve the robustness self-supervised might have learned during unsupervised pretraining (e.g. with adversarial pretraining).


\subsection{Can more training data improve robustness?}
\begin{figure}
     \centering
         \includegraphics[width=\textwidth]{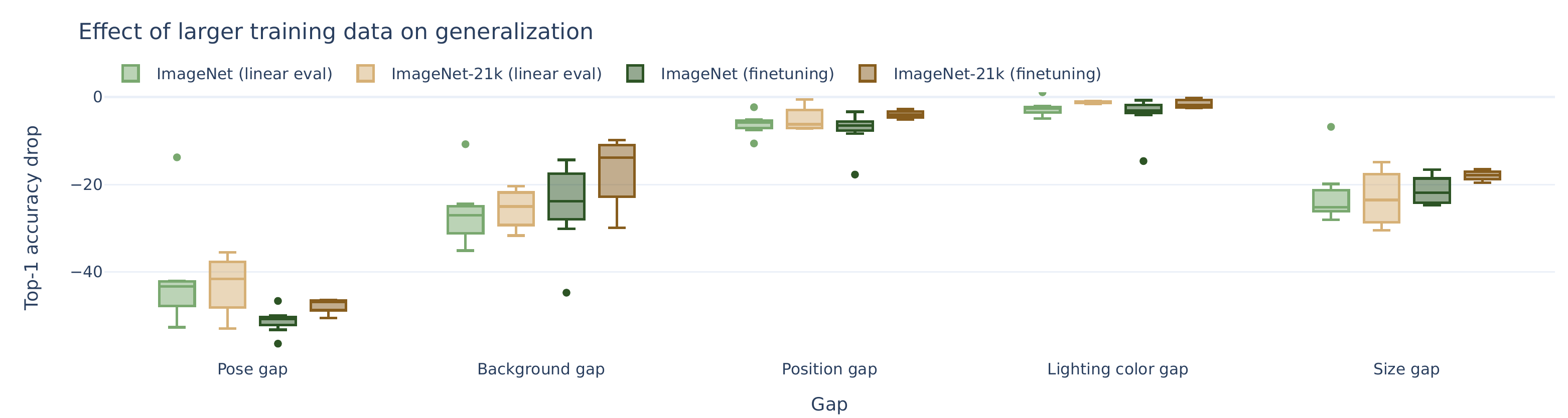}
     \caption{\textbf{Models trained on ImageNet-21k are more robust} compared to those trained on ImageNet-1k. We compare the effect training size for linear evaluation and finetuning.}
         \label{fig:21k_gaps}
\end{figure}

Recent works have shown that increasing the dataset size leads to substantial gains, especially for SSL models \citep{Zhai_2022_CVPR,GoyalWild,Hoffmann2022,Kaplan2020}. However, the effect of additional data on robustness remains unclear. To test this, in Figure \ref{fig:21k_gaps}, we focus on the comparison between ImageNet-21k (14 million training samples) and ImageNet-1k (1.2 million training samples).
We found that for both finetuning and linear evaluation, models trained on ImageNet-21k were substantially more robust than those trained on ImageNet-1k (Figure \ref{fig:21k_gaps}). 
Interestingly, this effect was more pronounced in the context of finetuning than linear evaluation, with pose, size, and position benefitting most.
Finetuning also led to less variance in accuracy drops across models, suggesting models robustness converges with finetuning compared to linear evaluation.



\section{Can models generalize variation from seeing variability in the training data?} \label{sec:diversity}

In the previous section, we demonstrated that SoTA vision models struggle to generalize across several common factors such as pose or size. We also observed that pre-training on larger datasets (ImageNet-21k vs. ImageNet-1k) led to improved robustness, consistent with other results demonstrating the impact of additional data \citep{radford2021learning,Kaplan2020, Hoffmann2022, Zhai_2022_CVPR}. 
Here, we study the extent to which models can generalize variability from training to held-out samples across three settings: 1) when all samples vary 2) when only some instances vary 3) when only some classes vary.

\subsection{How much training variability is needed to close the generalization gaps?}
\label{sec:singlefactorimages}

\begin{figure}[h!]
\includegraphics[width=\textwidth]{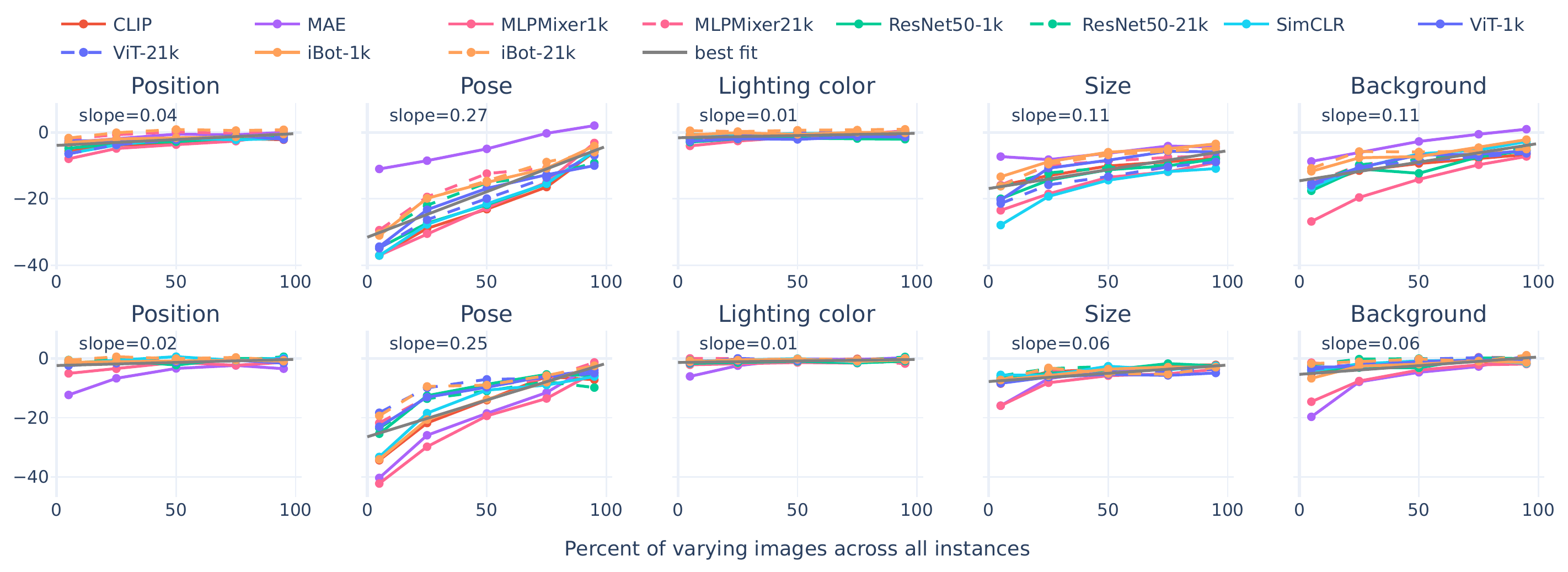}
\caption{\textbf{Models require significant variation across all samples to close generalization gaps.} 
We show the generalization gaps as the variability across all samples increases using linear evaluation (top) and finetuning (bottom). We find pose and size require an especially large portion of varying images to close the gap.}
\label{fig:data_diversity_all_instances}
\end{figure}

We first measure the extent to which seeing all instances varying during training can close the generalization gaps. In order to introduce variation, we ensure a particular fraction of samples per instance feature diverse values (i.e., departing from canonical). We increase the amount of variability from 5 to 95\% and evaluate how robustness to variability on novel instances at test time changes relative to the robustness of models trained only on data with canonical values for factors. To begin with, we analyze variation for each factor independently. Figure \ref{fig:data_diversity_all_instances} reports the effect of increasing variability on the generalization gaps for each factor. While all factors benefit from introducing variability, some factors such as pose and size still incur quite a large gap even with 50\% variability. This result demonstrates that while incorporating variability during training improves robustness, the magnitude of this effect varies substantially across factors.

\subsection{Can models generalize variation across instances?}
\label{sec:singlefactorinstance}

The previous experiment measured whether introducing variability across all training instances helped robustness, but it remains unclear whether models can generalize variability in one set of instances to a different set of instances. This is analogous to the experiments of \citet{alcorn_strike_2019} but extends their work to different levels of variability and additional factors. We thus introduce variability only for a subset of instances for each factor of variation. By contrast to Section \ref{sec:singlefactorimages}, variability in \% now refers to the percentage of the instances that are seen undergoing variations, while the rest of the instances are seen only with their canonical factors values during training. This is a substantially more difficult generalization problem, as exemplified by the larger gaps observed in this setting. However, while we found that models continue to struggle to generalize when the amount of instances seeing varying is low (<25\%), robustness improves with additional varying instances, with some factors reaching minimal gaps with as few as 50\% of the training instances are seen varying (Figure \ref{fig:data_diversity}). The pattern across factors was largely consistent, though both position and lighting color reached minimal gaps with comparatively less variability in training data, consistent with our previous observation that models are more robust to variance along these factors.\\
\\
Interestingly, varying only a portion of \emph{instances} led to substantial overfitting, especially when the proportion of varying instances is smaller than 50\%. Compared to the original gap with no diversity in Section \ref{sec:not_robust}, the gaps are higher when initially introducing variability, and only return to their baseline values once sufficient variability is reached. For example, while position and lighting have gaps of -5\% and -2\% respectively with no variability (Table \ref{tab:canonical}), their gaps when 5\% of instances vary are nearly -40\% (Figure \ref{fig:data_diversity}). This suggests models struggle to generalize variation across instances so much so that it can hurt generalization relative to seeing no variability.
\paragraph{Finetuning vs. linear evaluation as variability increases during training} To summarize these results across factors, for each subplot, we compute a linear fit to the average model curve and compute its slope. Models with higher slopes are more sensitive to the fraction of instances seen varying during training, while lower slopes indicate models which have the same generalization gap regardless of how much instances was presented varying during training. The average slope across factors and models for finetuning was $0.359 \pm 0.035$ vs. $0.441 \pm 0.020$ for linear evaluation (mean $\pm$ std). This result demonstrates that, while both benefit from increasing the percentage of instances seeing varying during training, this effect is much more pronounced for finetuning, providing further evidence that the impact of supervision is larger for finetuned models, likely because of the increased expressivity introduced by allowing all the weights to change.

\paragraph{Does training with instances varying for a single given factor improve robustness to variation in other factors?} 
\label{sec:crossfactoreffects}
\begin{figure}
    \centering
    \includegraphics[width=\textwidth]{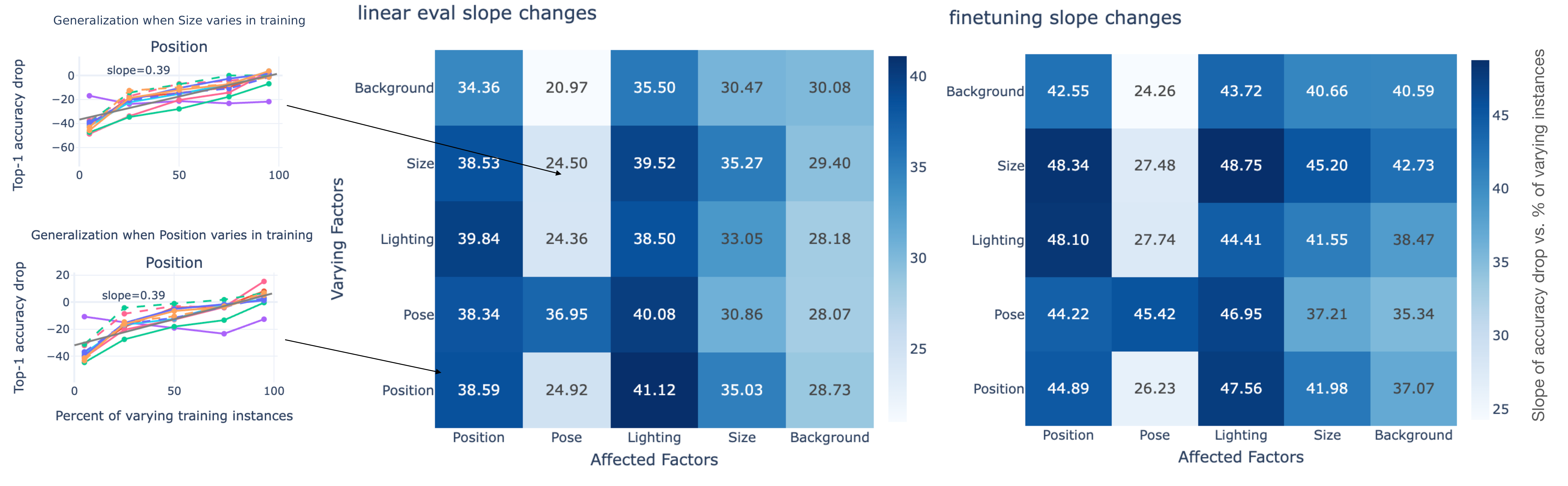}
    \caption{\textbf{Varying one factor can improve robustness to other factors}. We illustrate the cross-factor changes when a factor varies by plotting the change in gaps using the line of best fit as the number of varying instances increases.}
    \label{fig:cross_factor}
\end{figure}

Does robustness to a single factor provide broader robustness to other factors as well? 
To test this, we trained models with increasing amounts of variability for a single factor and evaluated the robustness of other factors. 
In Figure \ref{fig:fig1} (last column), we show the average change in gap for factors other than factor varying when we increase
the number of instances varying to 50\%. We find average effects of 1-3\%.
We further isolate this effect for each factor in the heatmaps show in Figure \ref{fig:cross_factor}
by plotting the slope of the line of best fit across models as we increase the portion of varying instances seen during training.
The diagonal of this heatmap represents cases where robustness evaluated for the same factor seen varying during training; off-diagonal entires measure changes for other factors. 
Inducing robustness to one factor consistently improved robustness for other factors by as much as 41\% for linear evaluation 
and 48\% for finetuning, though results varied across factor pairs. 
For example cross factor effects for pose are minor relative to those for position and lighting color.
In fact, we found position and lighting color most helped each other, suggesting that the impact of position and lighting color variability are somewhat entangled. 



\paragraph{Does larger pretraining data improve generalization to varying held-out instances?} 
To test the importance of pretraining data size, we compared models trained on ImageNet-1k to those trained on ImageNet-21k. As can be seen in figure~\ref{fig:agg-perf}, ImageNet-21k pretraining consistently improves robustness compared to ImageNet-1k pretraining, whether for finetuning or linear evaluation.


\subsection{Does training with instances varying in all factors improve robustness?} \label{sec:multi-factor}

\begin{figure}[t]
\includegraphics[width=\textwidth]{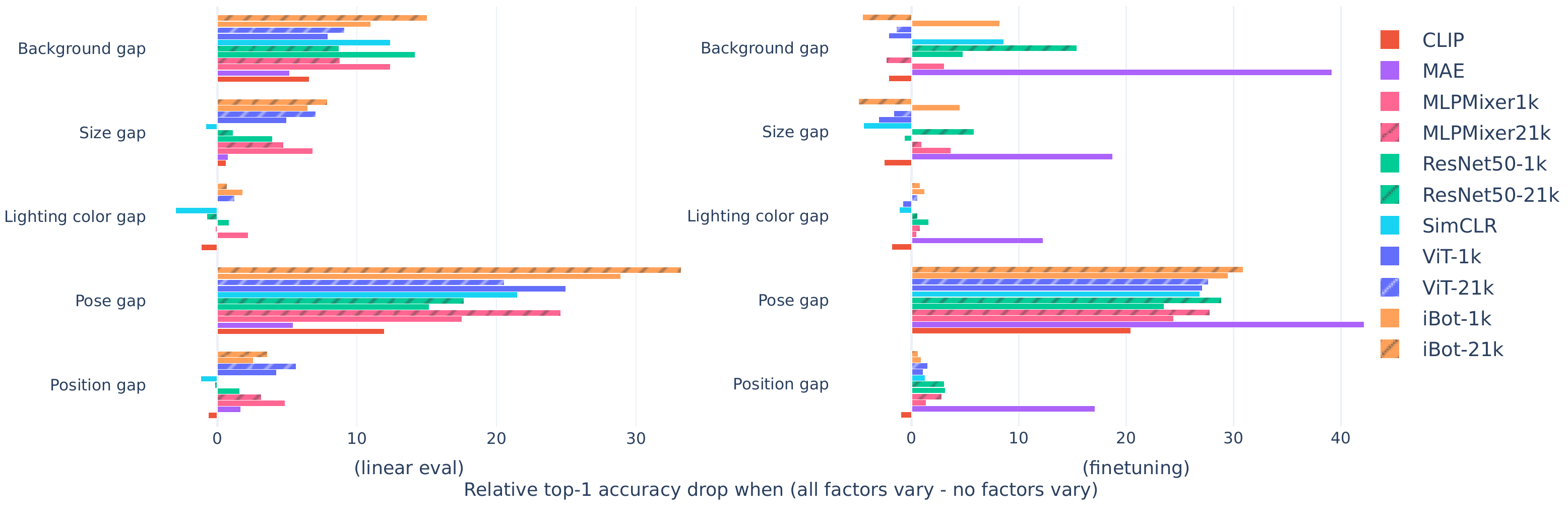}
    \centering
    \caption{\textbf{Varying all factors during training improves robustness} We show show relative generalization gaps when all factors vary during training relative to no instances seeing varying (no variability).}
    \label{fig:all_factor_relative}
\end{figure}

Training with variability for a single factor improves robustness both to the factor seen during training as well as other factors, but how does training with variability for all factors impact robustness? To test this, we selected random bases in the five-dimensional factor space and sampled images with random values along these bases during training. We report the change in the accuracy gap induced by incorporating factor variation during training (e.g., gap with no varying training factors - gap with all varying training factors). Positive values indicate an improvement in robustness, while negative values indicate a decrease. We found that training with variability across all factors led to substantially improved robustness for most factors, though lighting and color received no benefit, perhaps because its baseline robustness was already quite high (Figure \ref{fig:all_factor_relative}). Interestingly, pose benefited the most from training with variability across all factors, despite being helped the least from the individual cross-factor variability, suggesting that while variability in other factors can improve pose robustness, variability across multiple factors simultaneously is necessary to induce noticeable improvements. 


\subsection{Can models generalize variation across classes?} \label{sec:cross-class}
We have shown that introducing variability during training improves robustness for new instances, and in some cases, for entirely different factors of variation. However, in all prior experiments, the class distribution was held constant such that models were only asked to generalize to new instances from the same class. Can models generalize robustness across classes? To test this, we trained models with variability only present for a single factor for half of the classes (randomly selected). For classes trained with variability, half of the instances within that class were seen varying for the given factor. Results are summarized in Tables \ref{linear eval_class_generalization} and Table \ref{finetuning_class_generalization}.

\paragraph{Models are significantly less robust when variation is only seen for some classes}
 We found significant gaps in generalization when only half of classes were seen with variability for each of the factor, as shown in Table \ref{finetuning_class_generalization}. The average gap across all factors is -50\% more than double the gaps observed where no variability is seen during training at all. This implies that when variation is only observed for some classes, models generalize even more poorly and extends \cite{alcorn_strike_2019}'s results demonstrating lack of generalization across instances at the class level. Our finding suggests we should develop explicit mechanisms for improving model generalization across classes.

\begin{figure}[h]
\includegraphics[width=\textwidth]{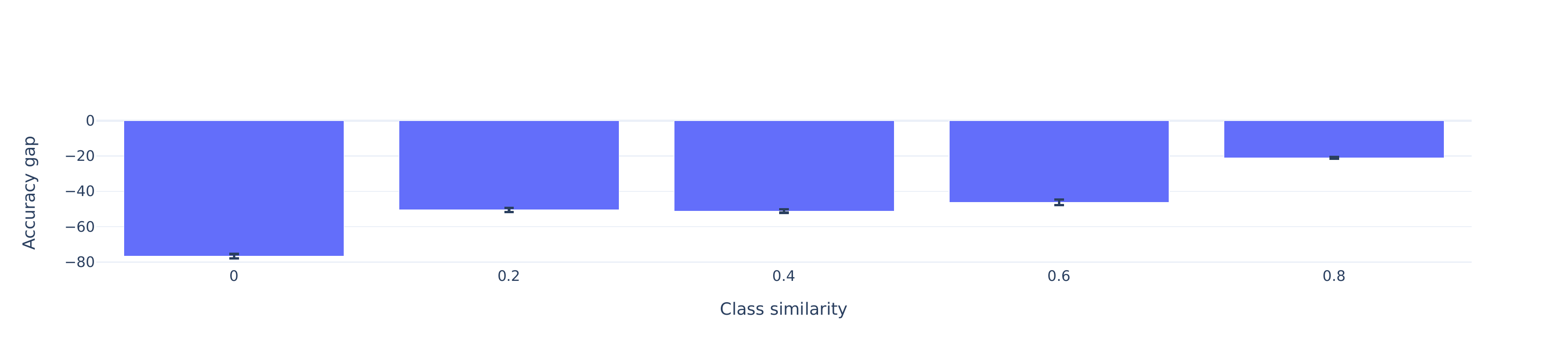}
    \caption{\textbf{Generalization gaps are smaller only for classes very similar to those seen during training and worse for classes that are very dissimilar.} We plot the generalization gaps as similarity to the nearest class seen varying during training increases using the mean accuracy gap with error bars indicating the standard error.}
    \label{fig:class_similarity}
\end{figure}

\paragraph{Models generalize equally poorly across classes, unless classes are very similar or very dissimilar to those seen varying during training}
It is possible that robustness can only be generalized across classes when the classes exceed some threshold similarity. To test this, we evaluated the cross-class robustness as a function of the distance between classes. Class distance was computed using a pre-trained word-embedding similarities \citep{spacy2}. While the most dissimilar classes were harmed more, the majority of classes exhibited a similar detrimental effect regardless of their similarity to the training classes that were varying (Figure \ref{fig:class_similarity}).

\begin{table}
\centering
\caption{\textbf{Models have significant gaps in generalization when only half of classes were seen varying.} Table shows generalization gap differences between classes (27 randomly selected) seen with diversity and those not when finetuning.}
\label{finetuning_class_generalization}
\begin{adjustbox}{width=0.9\textwidth}
\begin{tabular}{@{}lllllll@{}}
\toprule
       &  Position gap &  Pose gap &  Lighting color gap &  Size gap &  Background gap &  Average gap \\
\midrule
        CLIP &        -49.34 &    -62.91 &              -53.25 &    -50.30 &          -53.87 &       -53.93 \\
         MAE &        -37.17 &    -48.30 &              -51.14 &    -42.35 &          -49.21 &       -45.64 \\
  MLPMixer1k &        -47.33 &    -60.10 &              -51.26 &    -46.41 &          -52.36 &       -51.49 \\
 MLPMixer21k &        -46.18 &    -62.77 &              -50.28 &    -47.92 &          -47.79 &       -50.99 \\
 ResNet50-1k &        -45.66 &    -53.22 &              -43.62 &    -49.03 &          -35.44 &       -45.39 \\
ResNet50-21k &        -43.85 &    -54.09 &              -47.54 &    -47.12 &          -43.85 &       -47.29 \\
      SimCLR &        -45.49 &    -59.69 &              -46.59 &    -44.24 &          -29.39 &       -45.08 \\
      ViT-1k &        -48.13 &    -66.16 &              -51.90 &    -49.01 &          -48.39 &       -52.72 \\
     ViT-21k &        -47.17 &    -61.91 &              -49.93 &    -45.59 &          -47.56 &       -50.43 \\
     iBot-1k &        -46.53 &    -65.76 &              -49.82 &    -50.61 &          -51.47 &       -52.84 \\
    iBot-21k &        -48.22 &    -67.85 &              -51.46 &    -54.14 &          -49.33 &       -54.20 \\
     Average &        -45.91 &    -60.25 &              -49.71 &    -47.88 &          -46.24 &       -50.00 \\
\bottomrule
\end{tabular}
 \end{adjustbox}
\end{table}

\section{Discussion} \label{sec:discussion}
In order to develop robust, trustworthy models which do not fail when presented with distribution shift, we much characterize the generalization capabilities of our current best approaches. In this work, we provided an extensive study of the robustness of SoTA models to naturally occurring variations, extending on previous work in a number of ways. Our experiments show that models fail to generalize to variations of a set factors on held-out instances unless a reasonable amount of variability is seen during training. Surprisingly, we found that providing the model with training variability on a single factor can help generalize to other factors which were not varied during training. However, models struggle to transfer their knowledge of variations across classes: when only some classes are seen undergoing variability in training, only very similar classes (not seen varying at training) benefited at evaluation. Finally, we found that inductive biases such as architecture and training paradigm had minimal impact on models' converged robustness, in contrast to the pre-training data size and the method of downstream training. We hope that our work, by shedding further light on the blind spots of state-of-the-art models, can help practitioners develop robust models that can confidently and safely be deployed at large.

\section{Acknowledgements}

We thank Léon Bottou for feedback and helpful discussions.

\bibliography{biblio_diane}
\bibliographystyle{iclr2023_conference}

\newpage

\appendix
\section{Appendix}
\renewcommand{\thefigure}{A\arabic{figure}}
\setcounter{figure}{0}
\renewcommand\thetable{A\arabic{table}}
\setcounter{table}{0}

\section{Dataset samples}

\begin{figure}[!thbp]
    \centering
    \includegraphics[width=1\textwidth]{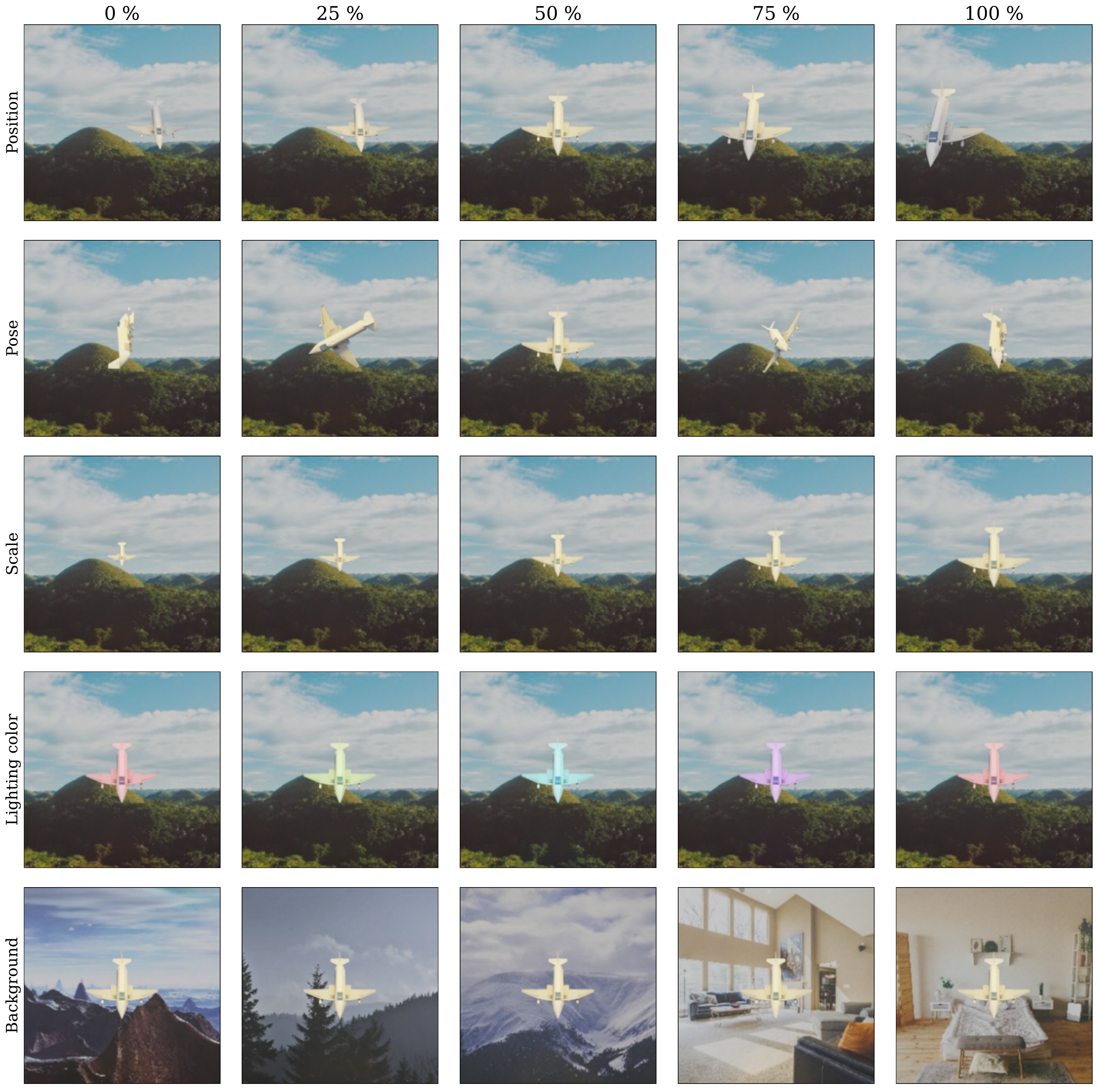}
    \caption{Examples from the dataset illustrating the different factors of variation.}
    \label{fig:data-samples-0}
\end{figure}
\begin{figure}[!thbp]
    \centering
    \includegraphics[width=1\textwidth]{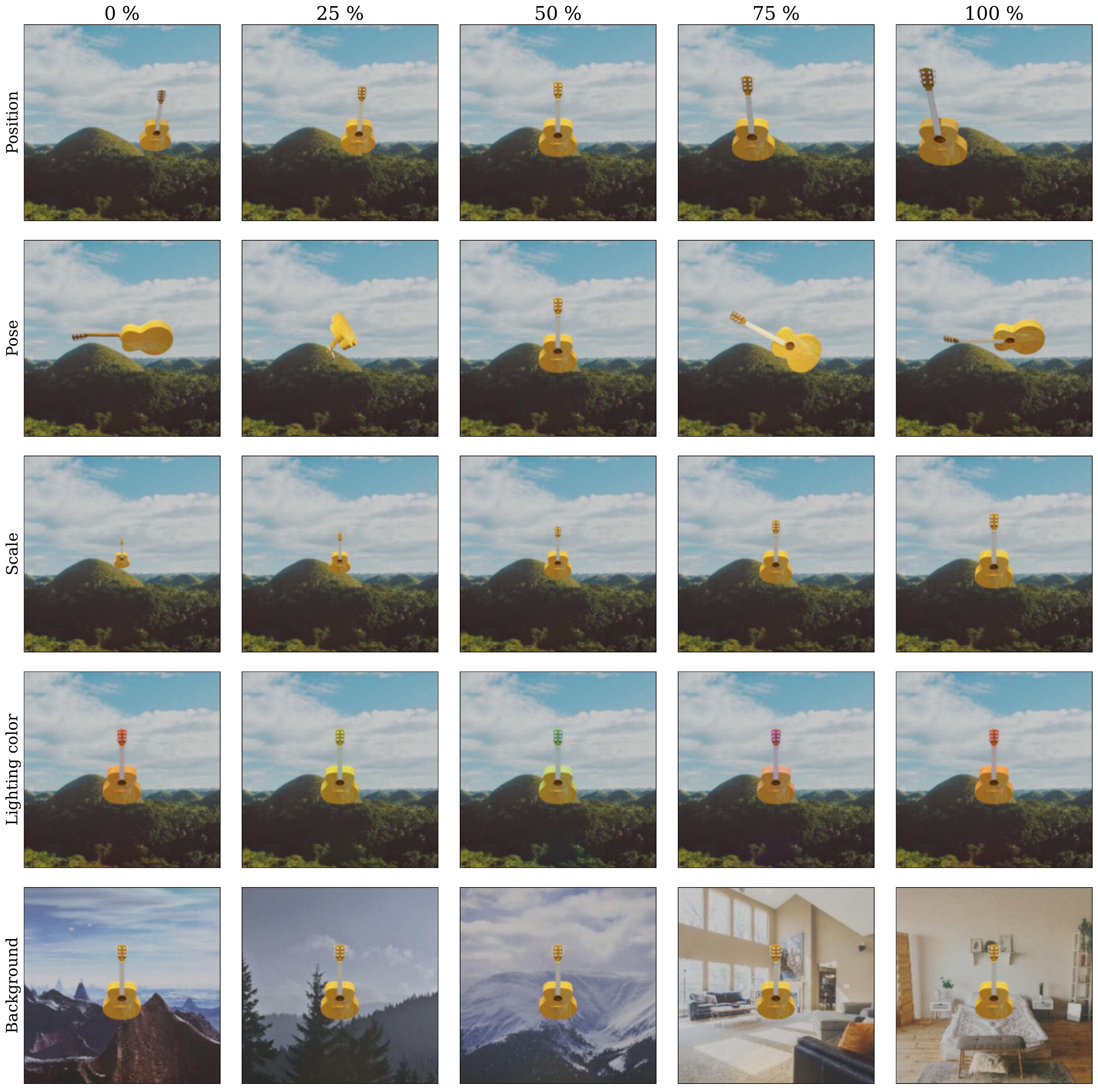}
    \caption{Examples from the dataset illustrating the different factors of variation.}
    \label{fig:data-samples-1}
\end{figure}
\begin{figure}[!thbp]
    \centering
    \includegraphics[width=1\textwidth]{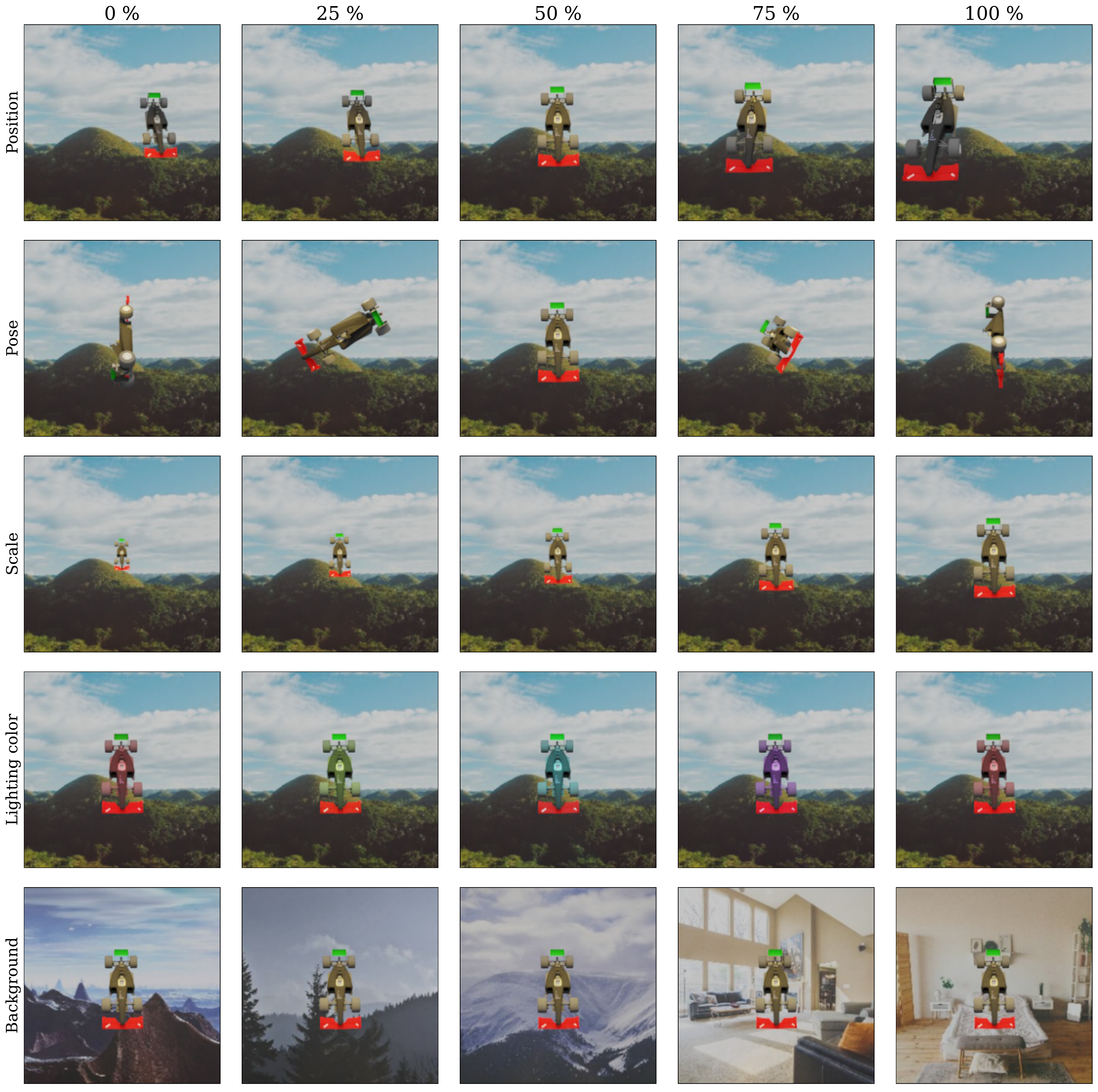}
    \caption{Examples from the dataset illustrating the different factors of variation.}
    \label{fig:data-samples-2}
\end{figure}
\begin{figure}[!thbp]
    \centering
    \includegraphics[width=1\textwidth]{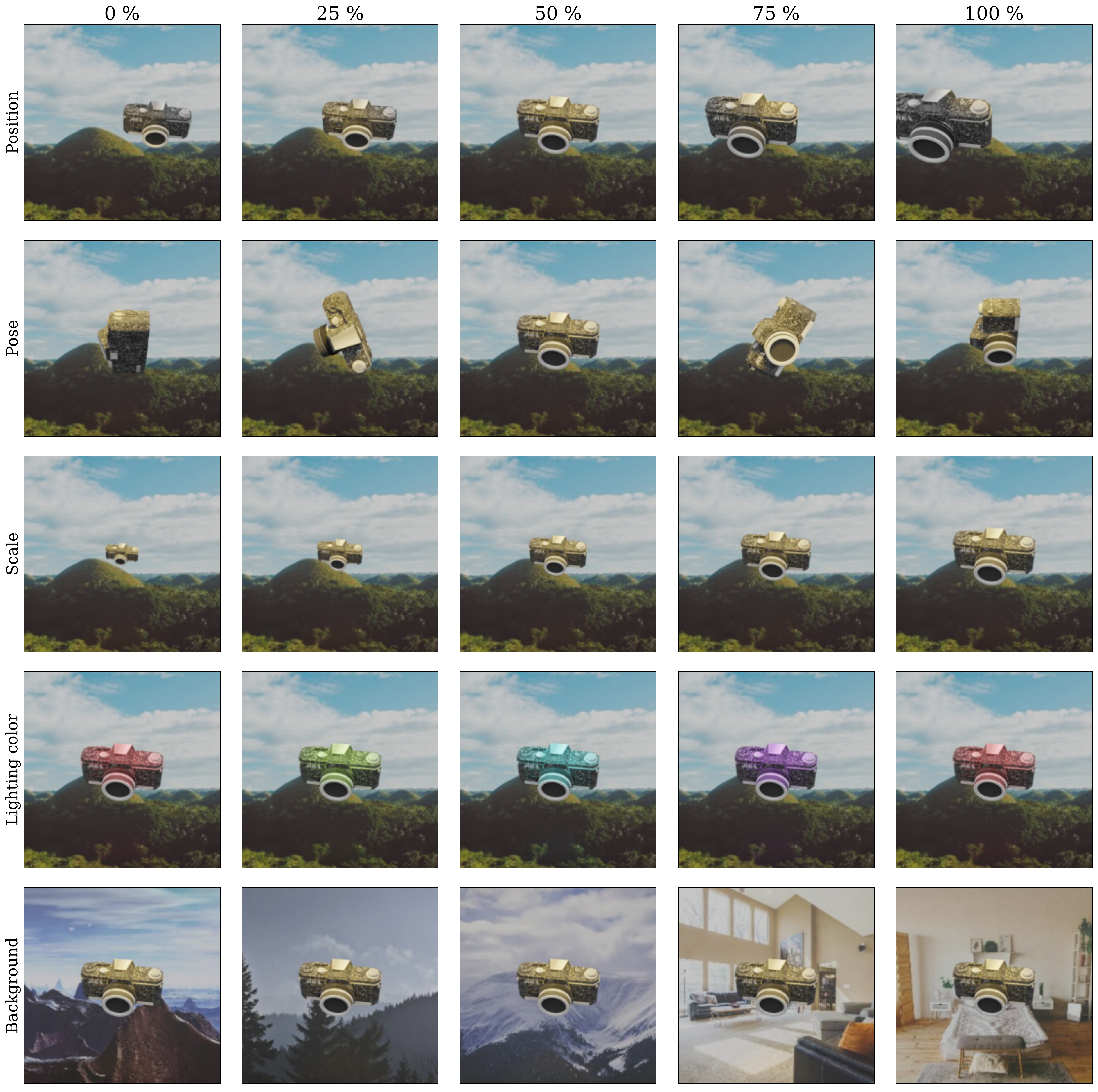}
    \caption{Examples from the dataset illustrating the different factors of variation.}
    \label{fig:data-samples-3}
\end{figure}
\begin{figure}[!thbp]
    \centering
    \includegraphics[width=1\textwidth]{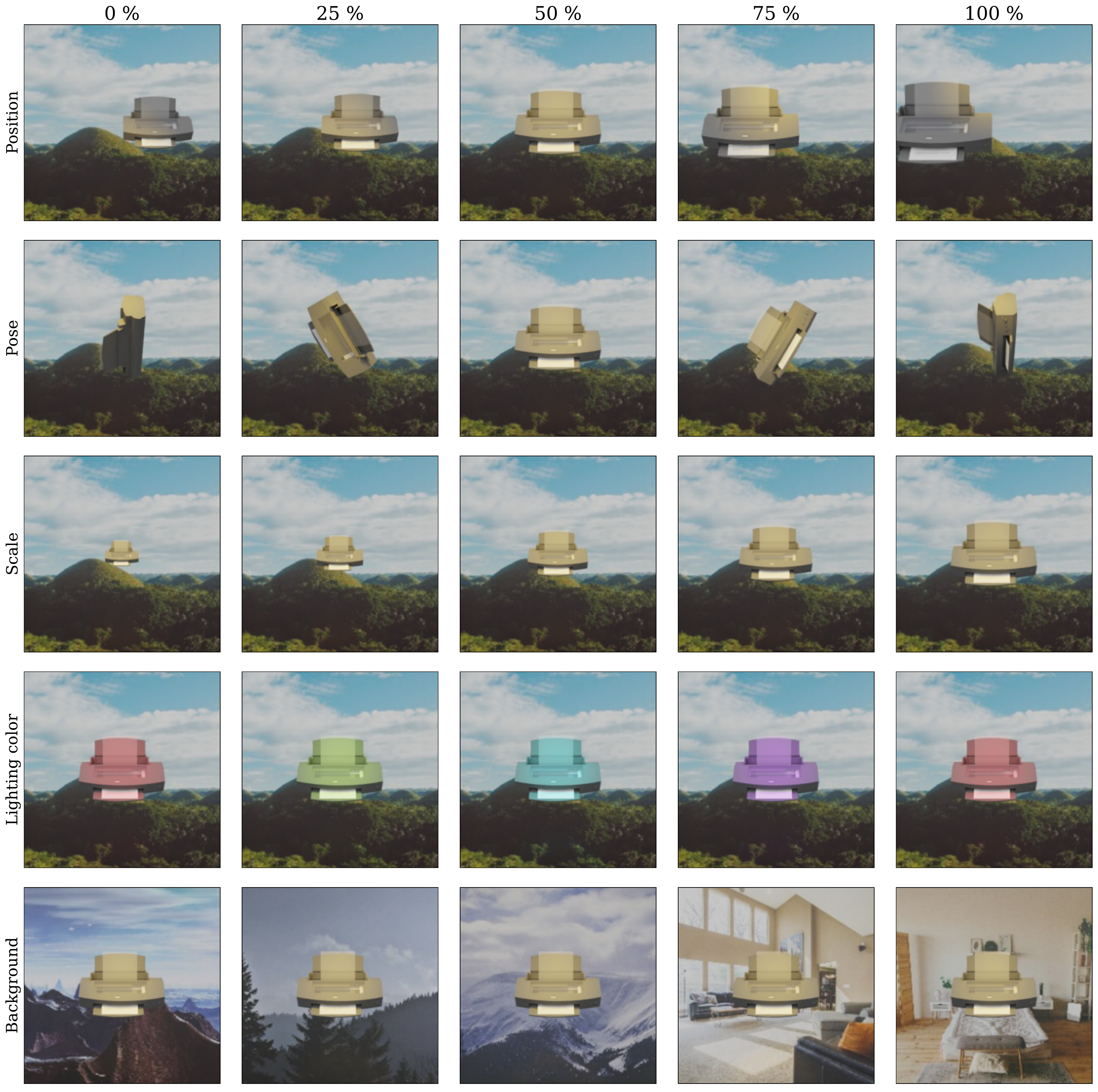}
    \caption{Examples from the dataset illustrating the different factors of variation.}
    \label{fig:data-samples-4}
\end{figure}
\begin{figure}[!thbp]
    \centering
    \includegraphics[width=1\textwidth]{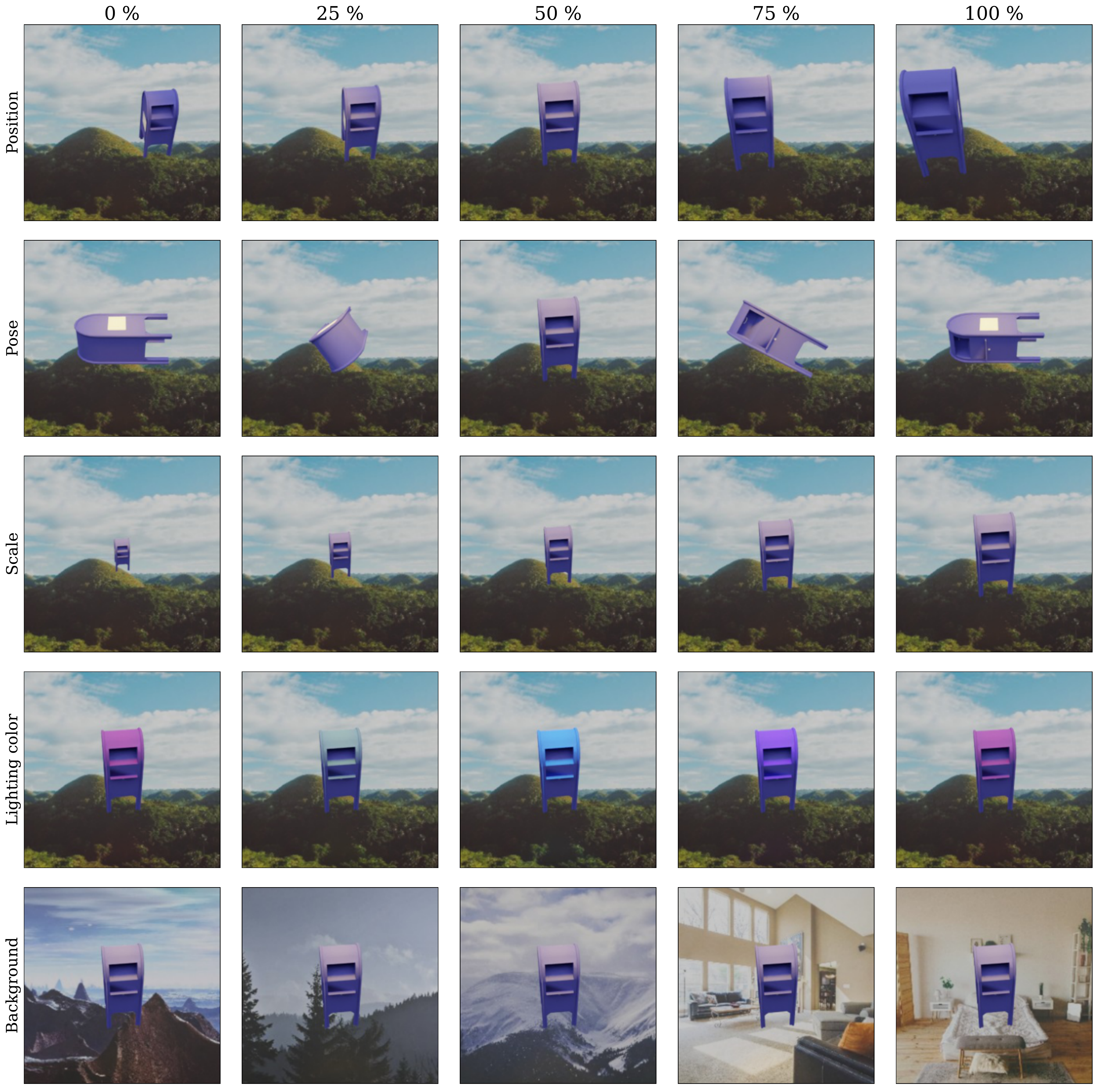}
    \caption{Examples from the dataset illustrating the different factors of variation.}
    \label{fig:data-samples-5}
\end{figure}
\begin{figure}[!thbp]
    \centering
    \includegraphics[width=1\textwidth]{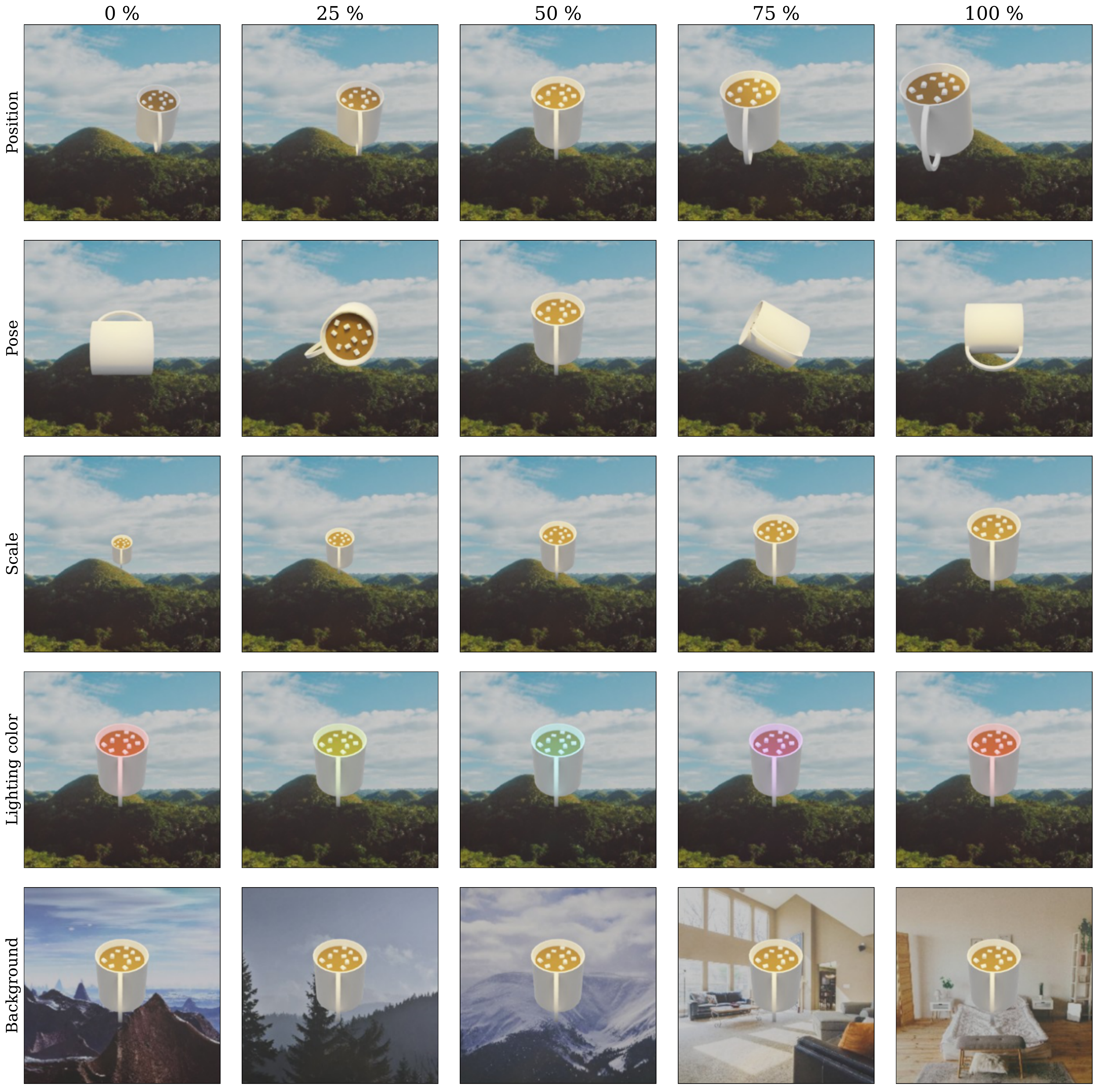}
    \caption{Examples from the dataset illustrating the different factors of variation.}
    \label{fig:data-samples-6}
\end{figure}
\begin{figure}[!thbp]
    \centering
    \includegraphics[width=1\textwidth]{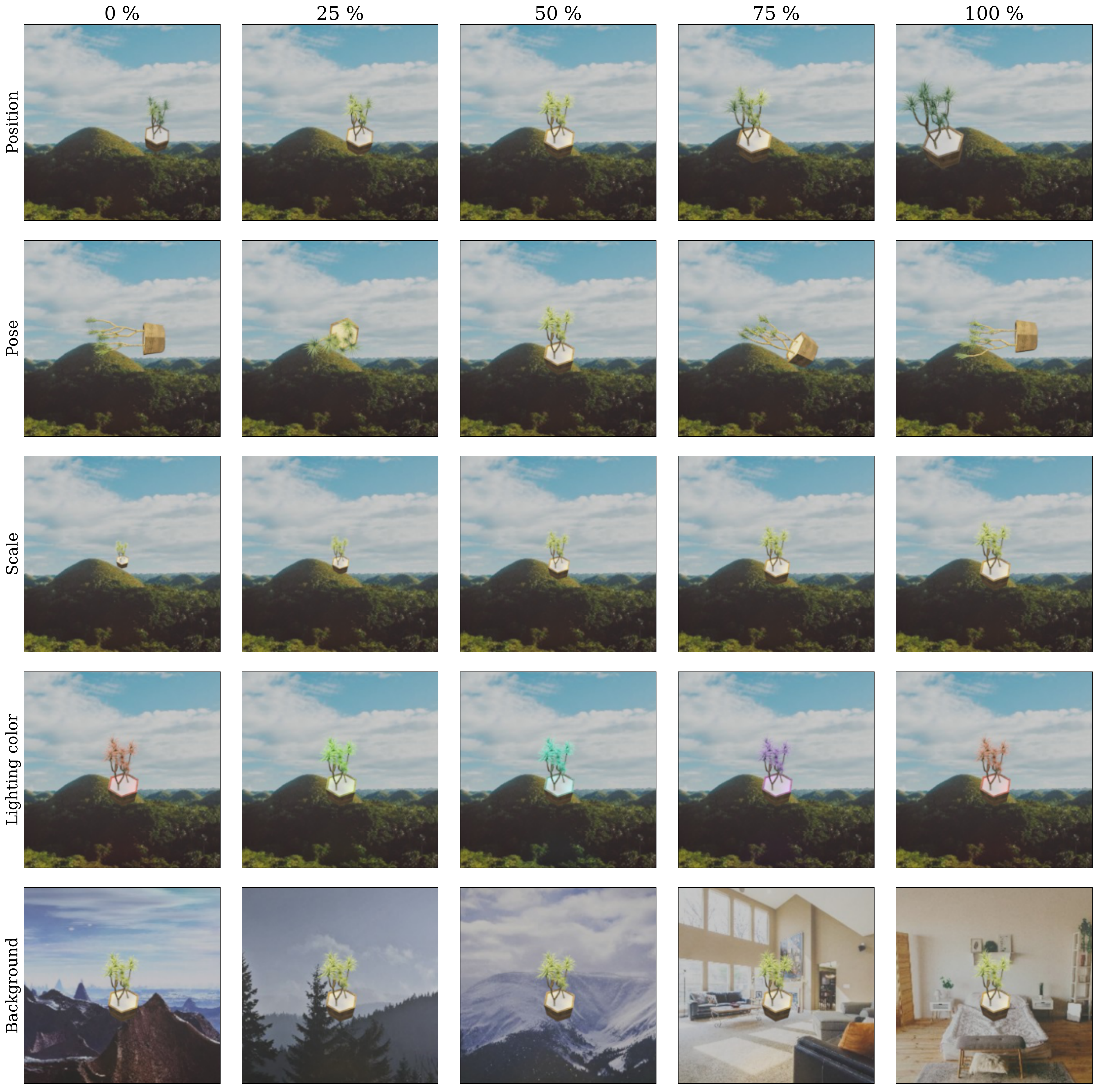}
    \caption{Examples from the dataset illustrating the different factors of variation.}
    \label{fig:data-samples-7}
\end{figure}
\begin{figure}[!thbp]
    \centering
    \includegraphics[width=1\textwidth]{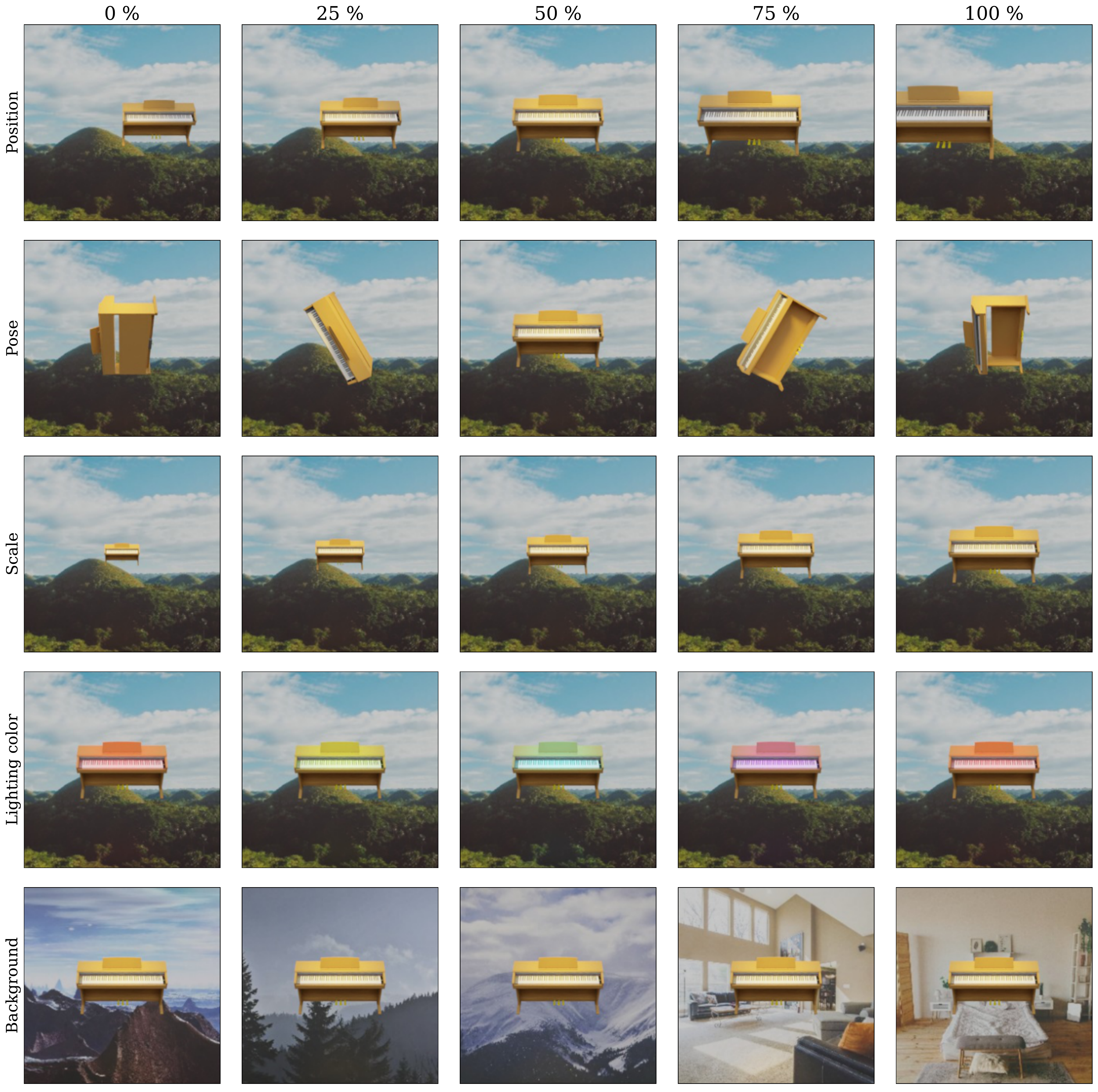}
    \caption{Examples from the dataset illustrating the different factors of variation.}
    \label{fig:data-samples-8}
\end{figure}
\begin{figure}[!thbp]
    \centering
    \includegraphics[width=1\textwidth]{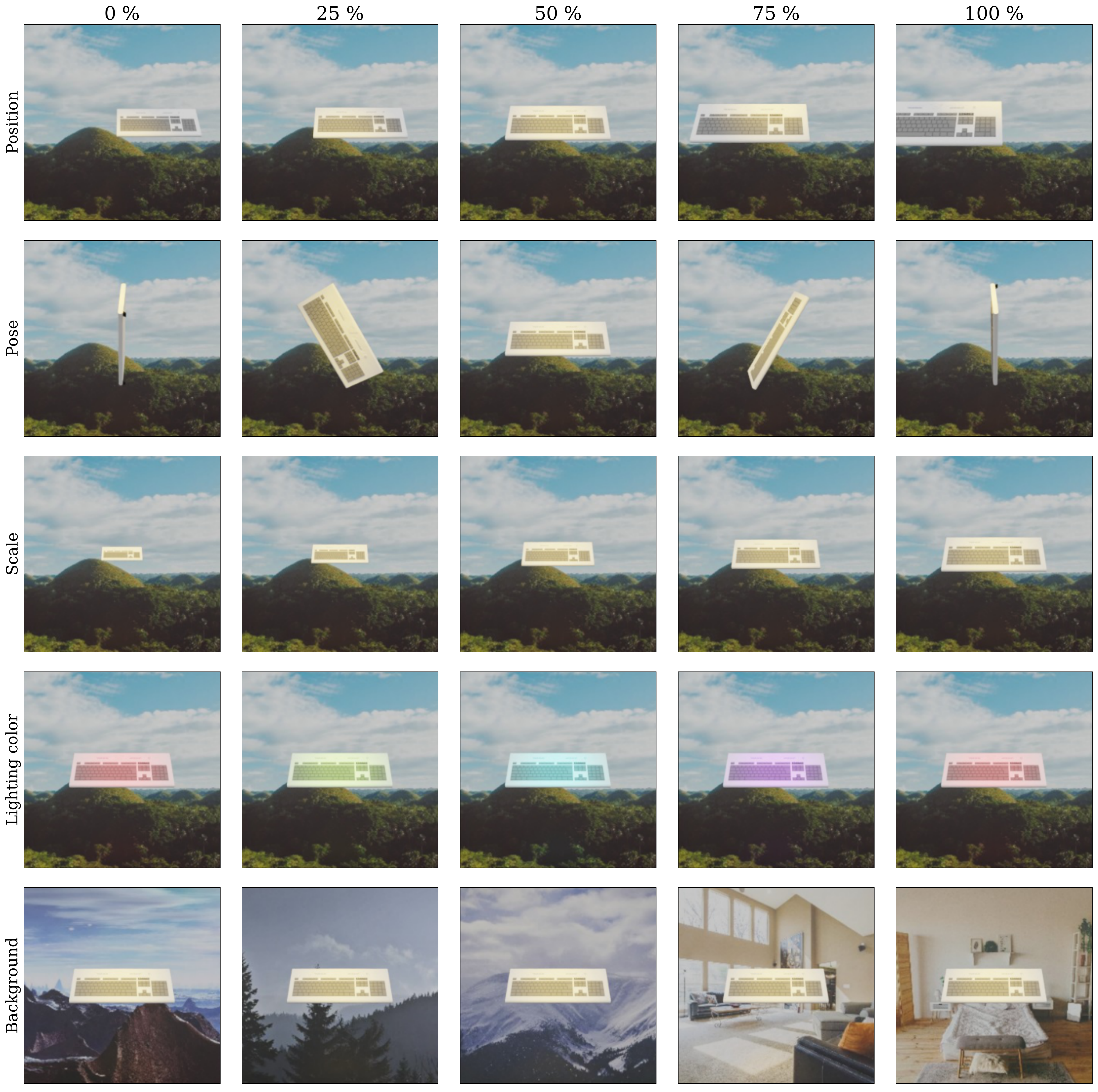}
    \caption{Examples from the dataset illustrating the different factors of variation.}
    \label{fig:data-samples-9}
\end{figure}

\section{Aggregated performances for linear vs finetuning and 21k vs 1k}
\label{app:diverse_1k_v_21k}
\begin{figure}[!thbp]
    \centering
    \includegraphics[width=1\textwidth]{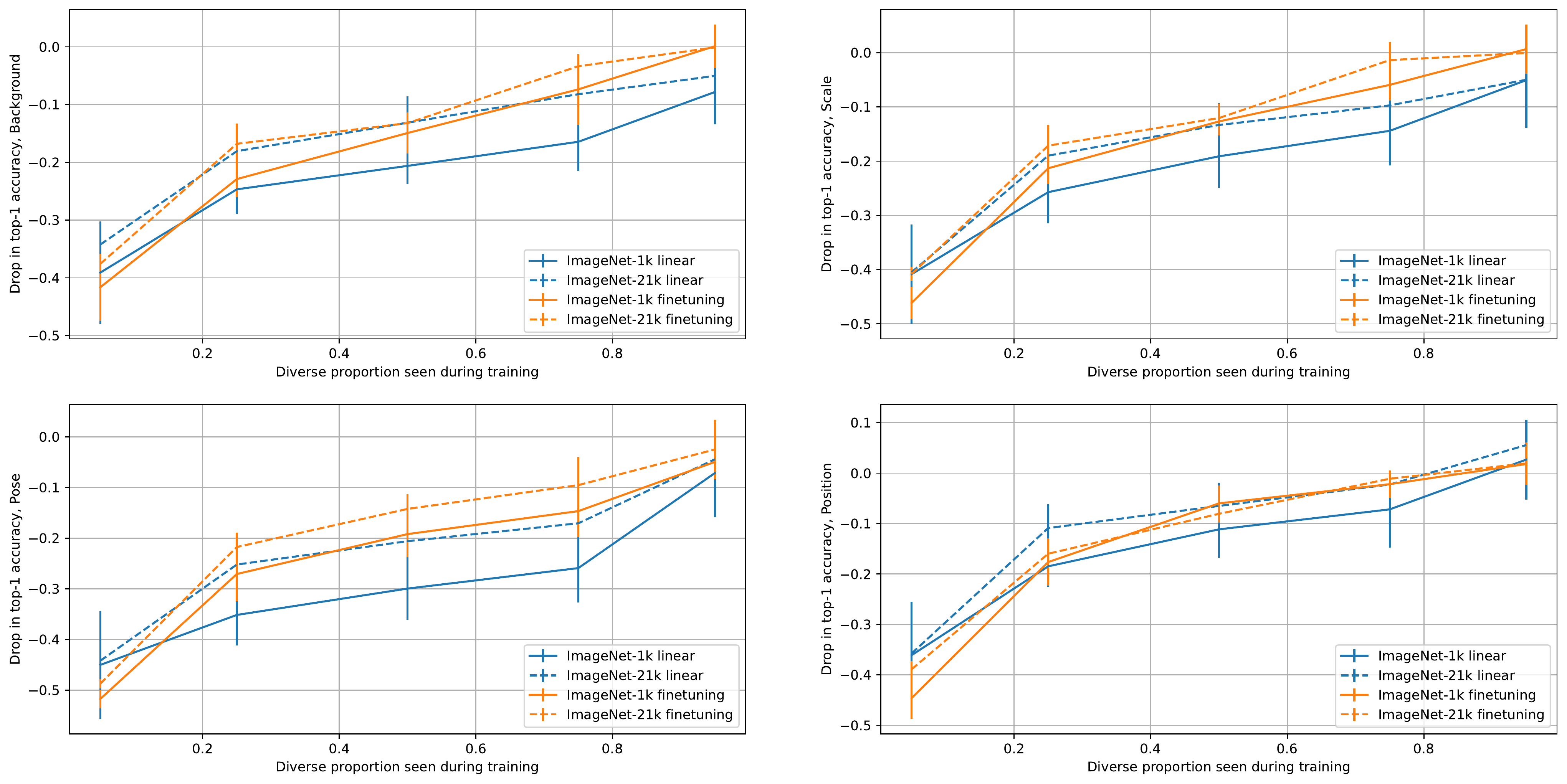}
    \caption{Drops in performance averaged over all methods when varying the proportion of varying examples seen during training.}
    \label{fig:agg-perf}
\end{figure}

As can be seen in figure~\ref{fig:agg-perf}, finetuning usually leads to lower drops in performance with high variability rates during training. However linear evaluation is more robust when diversity was not encountered during training. Pretraining on ImageNet21k always improves robustness compared to ImageNet-1k pretraining, whether in finetuning or in linear evaluation. It is worth noting that for translation robustness, all settings exhibit similar performance, and finetuning only benefits ImageNet-1k pretraining.

\subsection{CLIP Zero Shot Classification}
\label{clip_zero_shot}

We also evaluate CLIP's robustness using zero-shot classification. We assess both the standard Open AI CLIP model as well as CLIP trained on 2B LAION images. We prompt the model using "a photo of a []". CLIP with LAION-2B accuracies are 31.9\% for canonical, 15.9\% when pose varies, 18.2\% when scale varies, 31.9\% when lighting color, and 26.8\% background varies. CLIP with trained on 400M images has canonical 30.1\%, pose 16.0\%, scale 18.4\%, lighting color 28.3\%, and background 23.6\% accuracy.

We examine other prompts ("[], an inanimate object", "a photo of a [], an inanimate object", "[], a household item or vehicle") and observe similar classification performance (25-31\% accuracy) using these variants. 

\section{Cross Factor effects when varying all instances}

\begin{figure}
    \centering
    \includegraphics[width=\textwidth]{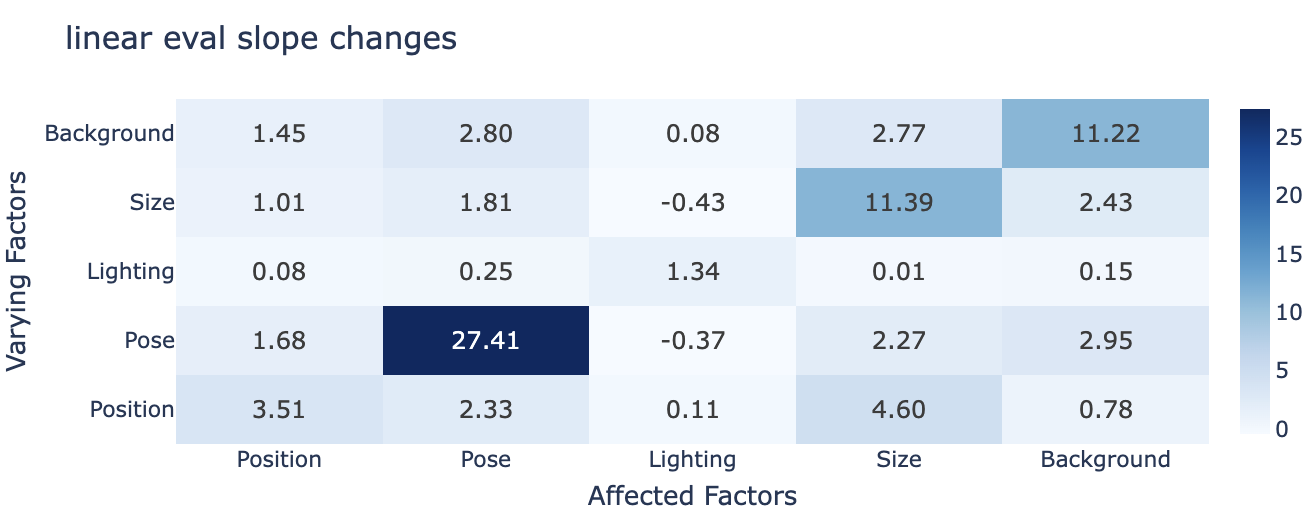}
    \caption{Cross factor changes when the given factor is varying for linear evaluation}
    \label{fig:linear_eval_slopes_image_sampling}
\end{figure}

\begin{figure}
    \centering
    \includegraphics[width=\textwidth]{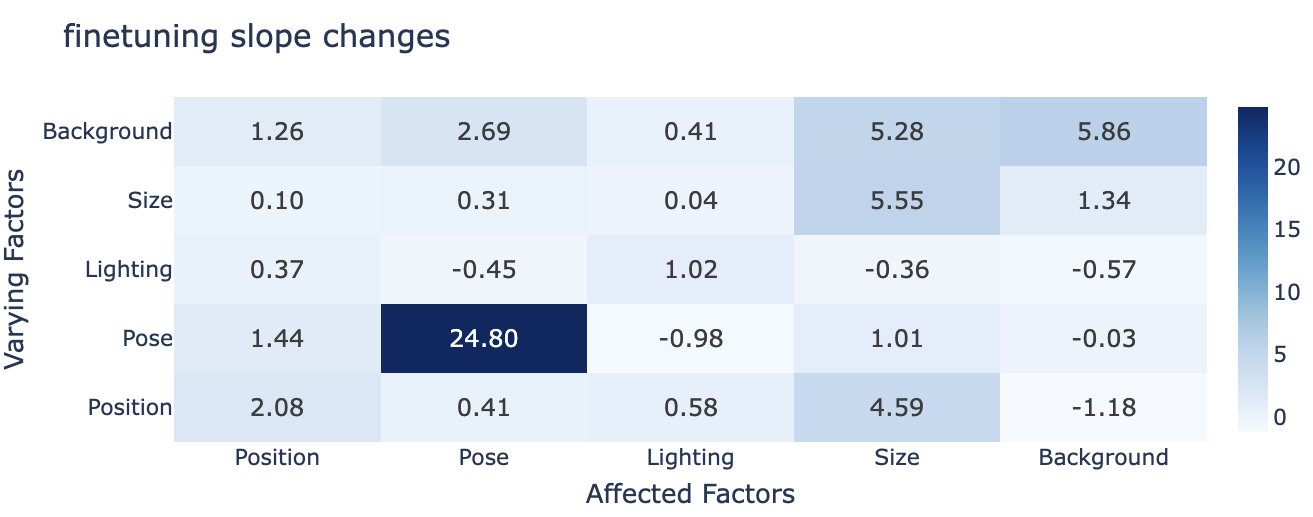}
    \caption{Cross factor changes when the given factor is varying for finetuning}
    \label{fig:finetuning_slopes_image_sampling}
\end{figure}

The cross factor effects when all instances vary with increasing diversity levels are shown in Figures \ref{fig:linear_eval_slopes_image_sampling} and \ref{fig:finetuning_slopes_image_sampling}.

\section{Varying a subset of instances during training}
\begin{figure}[h!]
\begin{subfigure}[t]{\textwidth}
\includegraphics[width=\textwidth]{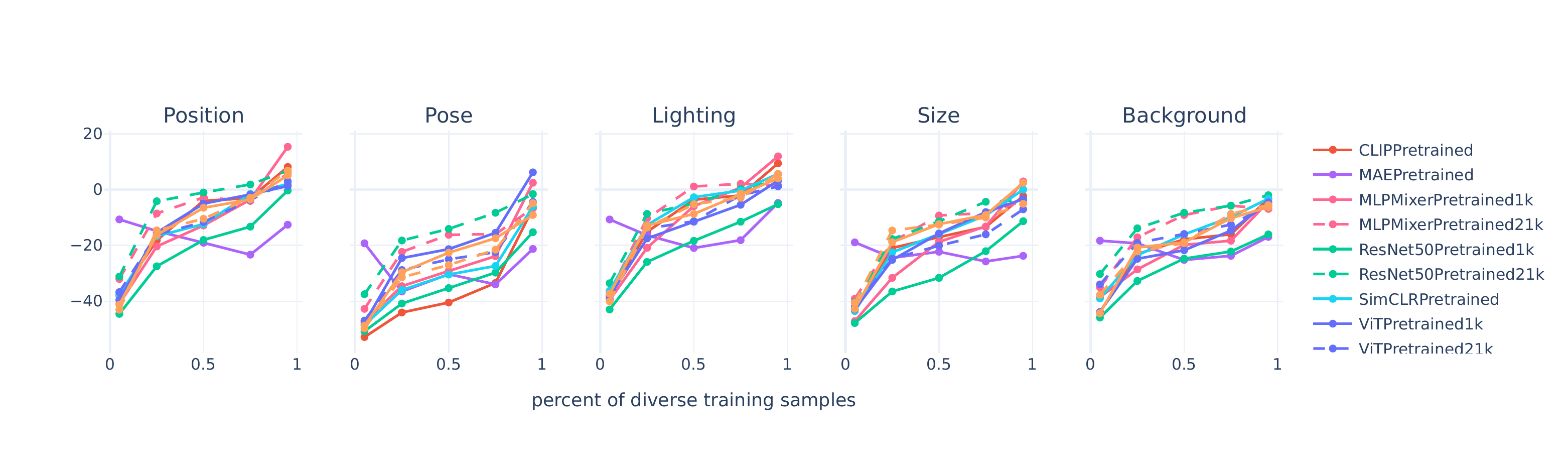}
\end{subfigure}
\begin{subfigure}[t]{\textwidth}
\includegraphics[width=\textwidth]{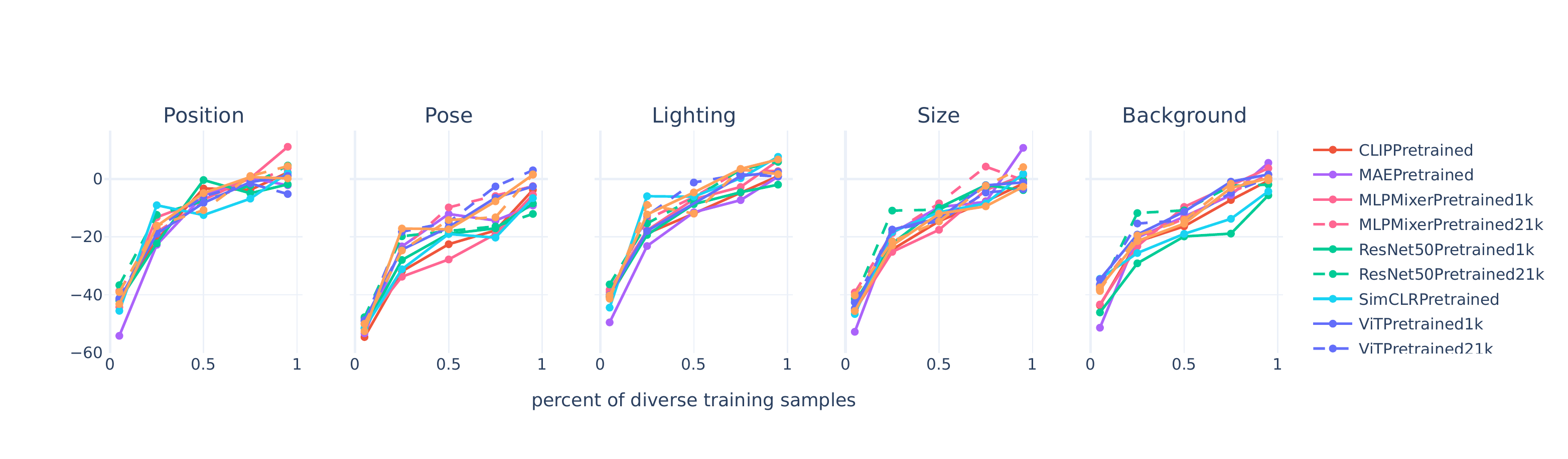}
    \caption{Training with increasing percentage of variability across all instances using finetuning}
\end{subfigure}
\caption{Training with increasing percentage of instances seen varying during training using linear evaluation (top) and finetuning (bottom).}
\label{fig:data_diversity}
\end{figure}

\begin{figure}[h]
\begin{subfigure}{\textwidth}
\includegraphics[width=\textwidth]{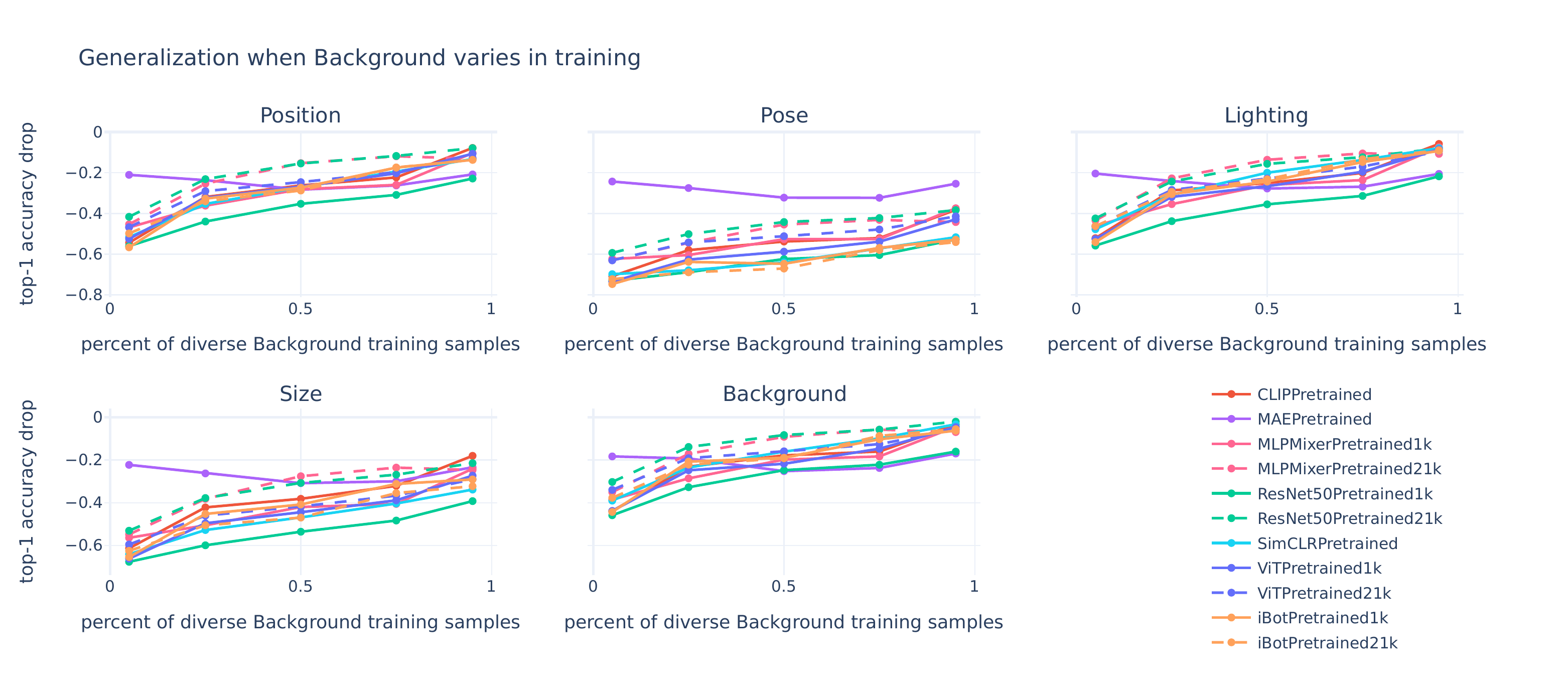} 
\end{subfigure}
\begin{subfigure}{\textwidth}
\includegraphics[width=\textwidth]{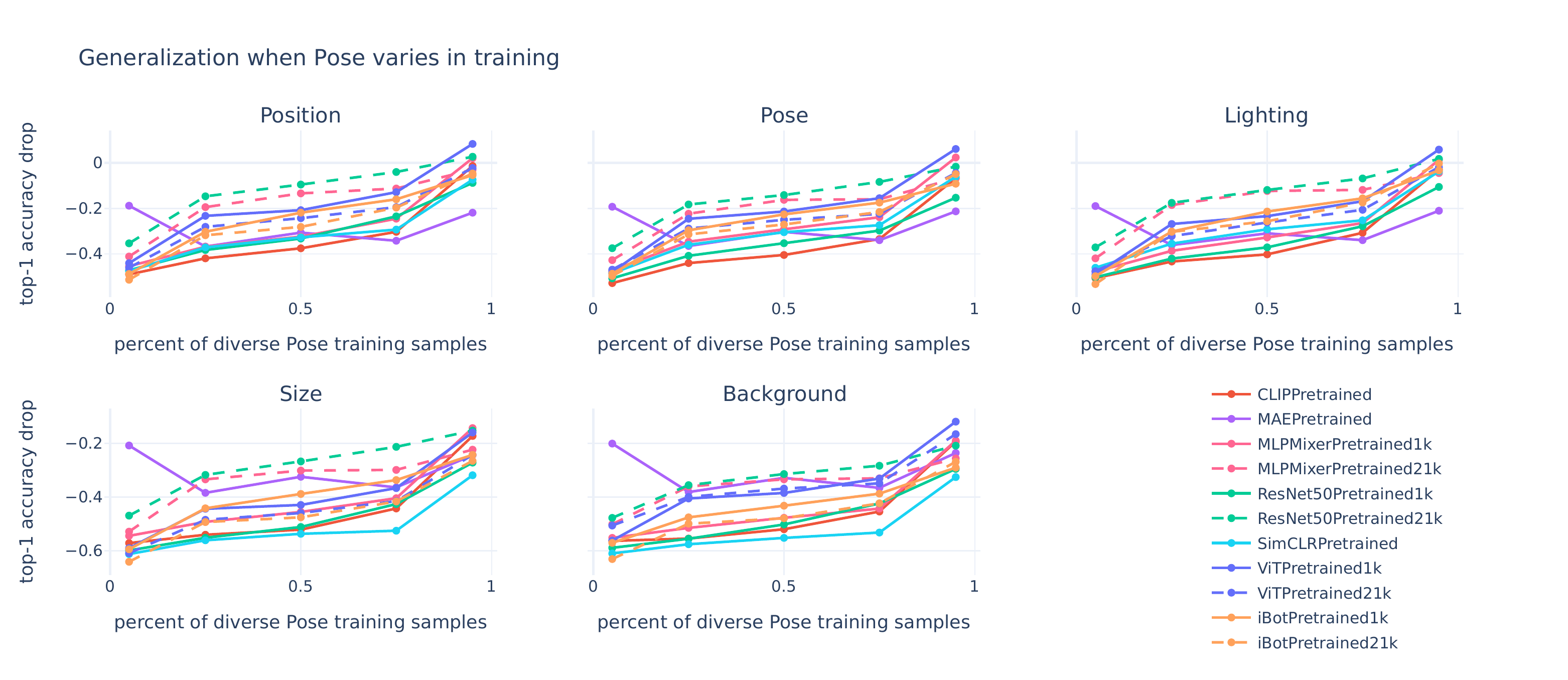} 
\end{subfigure}
\begin{subfigure}{\textwidth}
\includegraphics[width=\textwidth]{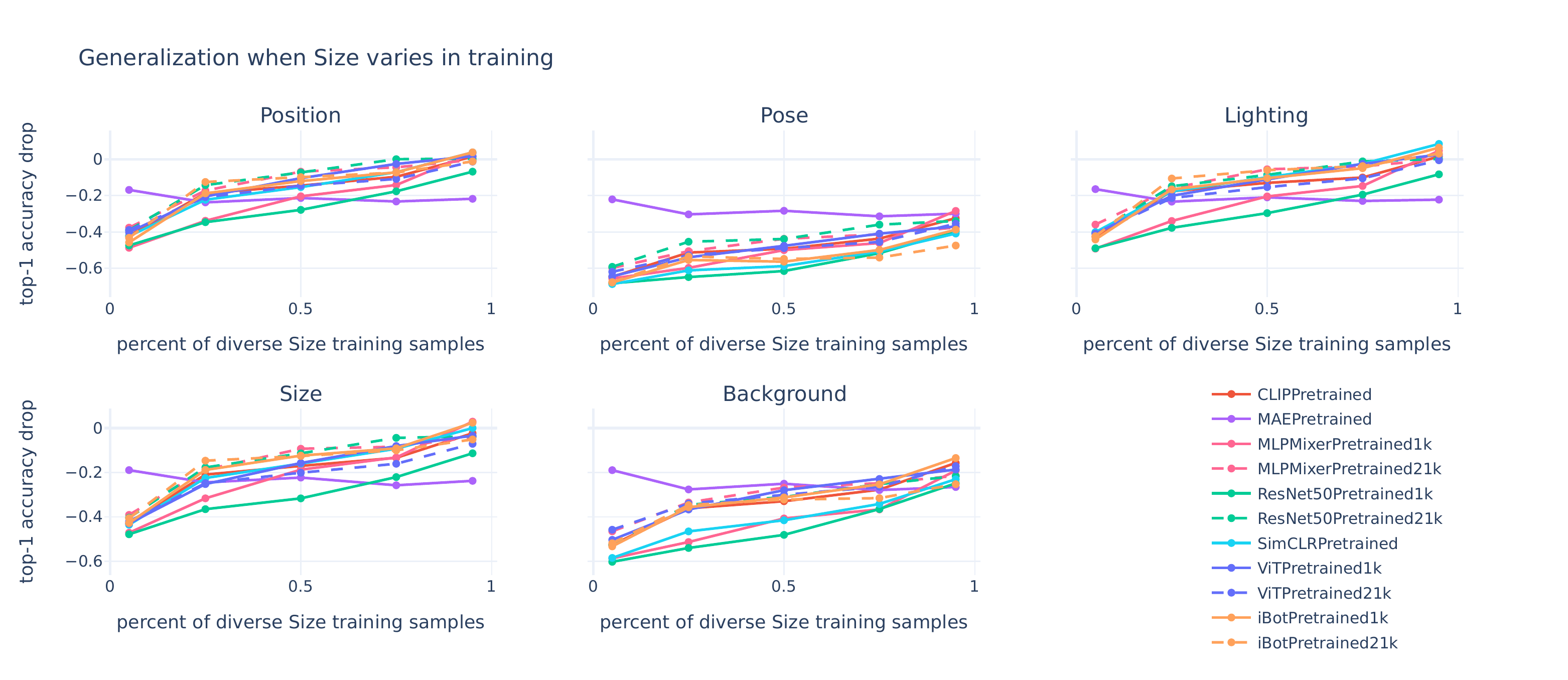} 
\end{subfigure}
\caption{Linear Evaluation Effect of Variability in Training (part 1)}
\label{app_fig:linear_diverse_factor}
\end{figure}

We show the effect of increasing the number of instances seen varying during training. In Figure \ref{fig:data_diversity} we show the effect of each factor. We break down the effect by factor in figures \ref{app_fig:linear_diverse_factor} for linear evaluation and finetuning 
\ref{app_fig:finetuning_diverse_factor}. 
In addition, we show the overall accuracy in tables \ref{app:linear eval_Translationdiverse_training},
\ref{app:linear eval_Background pathdiverse_training}, \ref{app:linear eval_Rotationdiverse_training},
\ref{app:linear eval_Spot huediverse_training}. The val canonical column corresponds to held-out accuracy for canonical and the
val diverse corresponds to the accuracy for a changing factor.

\begin{figure}[h]
\begin{subfigure}{\textwidth}
\includegraphics[width=\textwidth]{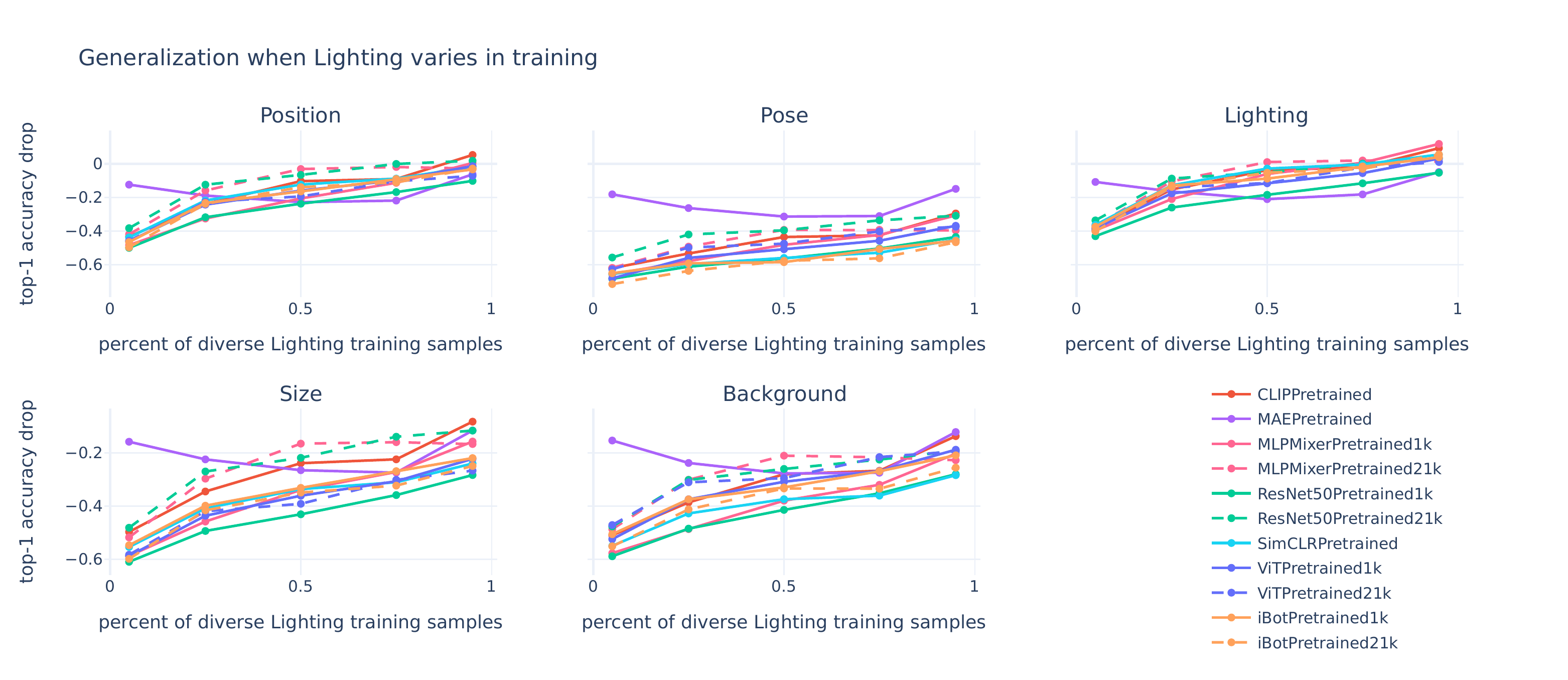} 
\end{subfigure}
\begin{subfigure}{\textwidth}
\includegraphics[width=\textwidth]{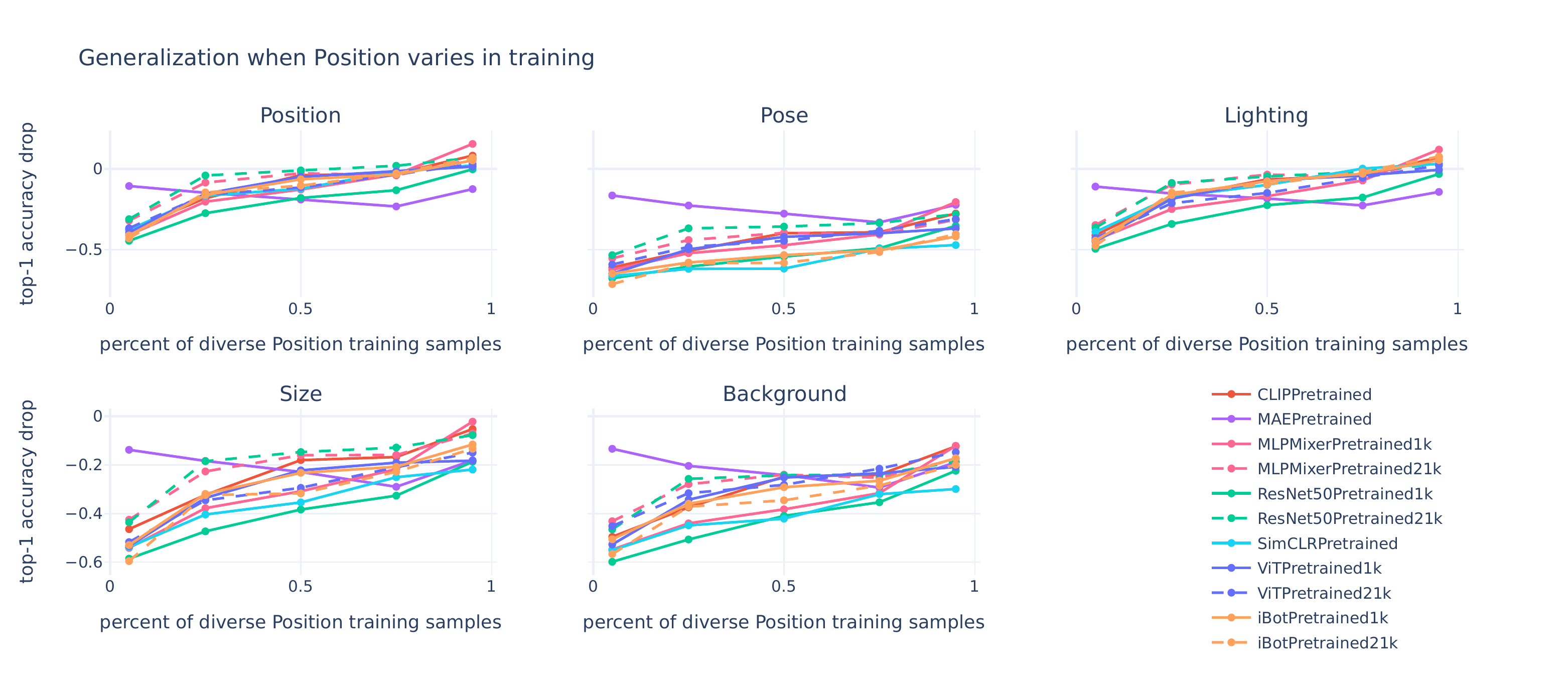} 
\end{subfigure}
\caption{Linear Evaluation Effect of Variability in Training (part 2)}
\label{app_fig:linear_diverse_factor_2}
\end{figure}

\begin{table}
\centering
\caption{Position varying linear eval top-1 accuracy across multiple percentages of varying training instances.}
\label{app:linear eval_Translationdiverse_training}
\begin{adjustbox}{width=\textwidth}
\begin{tabular}{lrrrrrrrrrrrrrrr}
\toprule
{} & \multicolumn{5}{l}{train\_canonical\_top\_1\_accuracy} & \multicolumn{5}{l}{val\_canonical\_top\_1\_accuracy} & \multicolumn{5}{l}{val\_diverse\_Translation\_top\_1\_accuracy} \\
train\_prop\_to\_vary &                           0.05 &   0.25 &   0.50 &   0.75 &   0.95 &                         0.05 &   0.25 &   0.50 &   0.75 &   0.95 &                                   0.05 &   0.25 &   0.50 &   0.75 &   0.95 \\
model                 &                                &        &        &        &        &                              &        &        &        &        &                                        &        &        &        &        \\
\midrule
CLIPPretrained        &                         92.24\% & 91.37\% & 92.25\% & 94.01\% & 96.33\% &                       77.78\% & 75.93\% & 68.52\% & 68.52\% & 58.89\% &                                 39.82\% & 57.70\% & 64.51\% & 65.45\% & 66.98\% \\
MAEPretrained         &                         52.54\% & 55.93\% & 60.44\% & 67.91\% & 83.79\% &                       20.37\% & 27.78\% & 33.33\% & 38.89\% & 27.78\% &                                  9.67\% & 12.91\% & 14.26\% & 15.56\% & 15.18\% \\
MLPMixerPretrained1k  &                         94.66\% & 93.98\% & 94.58\% & 95.75\% & 97.25\% &                       77.78\% & 74.07\% & 72.22\% & 66.67\% & 48.15\% &                                 36.86\% & 53.70\% & 59.35\% & 63.05\% & 63.46\% \\
MLPMixerPretrained21k &                         94.95\% & 95.09\% & 95.19\% & 96.02\% & 97.13\% &                       75.93\% & 75.93\% & 74.07\% & 77.78\% & 70.37\% &                                 43.89\% & 67.36\% & 71.25\% & 73.69\% & 76.18\% \\
ResNet50Pretrained1k  &                         95.00\% & 94.58\% & 94.83\% & 95.79\% & 97.23\% &                       88.89\% & 90.74\% & 87.04\% & 83.33\% & 70.37\% &                                 44.35\% & 63.26\% & 68.99\% & 70.04\% & 70.01\% \\
ResNet50Pretrained21k &                         95.51\% & 95.30\% & 95.90\% & 96.27\% & 97.39\% &                       77.78\% & 72.22\% & 74.07\% & 74.07\% & 70.37\% &                                 46.61\% & 68.04\% & 73.02\% & 75.90\% & 77.13\% \\
SimCLRPretrained      &                         96.13\% & 95.74\% & 96.22\% & 96.82\% & 97.50\% &                       81.48\% & 79.63\% & 81.48\% & 72.22\% & 70.00\% &                                 43.73\% & 62.96\% & 69.10\% & 70.63\% & 72.02\% \\
ViTPretrained1k       &                         95.79\% & 96.18\% & 96.37\% & 96.80\% & 97.75\% &                       88.89\% & 83.33\% & 77.78\% & 77.41\% & 75.93\% &                                 49.34\% & 67.92\% & 72.74\% & 75.65\% & 77.22\% \\
ViTPretrained21k      &                         95.43\% & 95.01\% & 95.62\% & 96.39\% & 97.50\% &                       83.33\% & 83.33\% & 83.33\% & 77.78\% & 72.22\% &                                 46.59\% & 67.21\% & 71.71\% & 74.02\% & 75.17\% \\
iBotPretrained1k      &                         96.67\% & 96.43\% & 96.49\% & 97.01\% & 97.66\% &                       81.48\% & 81.48\% & 79.63\% & 79.63\% & 72.22\% &                                 40.27\% & 65.23\% & 73.10\% & 76.06\% & 77.33\% \\
iBotPretrained21k     &                         96.84\% & 96.30\% & 96.44\% & 96.92\% & 97.61\% &                       90.74\% & 85.19\% & 87.04\% & 81.48\% & 72.22\% &                                 47.69\% & 70.55\% & 76.57\% & 78.53\% & 79.04\% \\
\bottomrule
\end{tabular}
 \end{adjustbox}

\end{table}

\begin{table}
\centering
\caption{Pose linear eval top-1  accuracy across multiple percentages of varying training instances.}
\label{app:linear eval_Rotationdiverse_training}
\begin{adjustbox}{width=\textwidth}
\begin{tabular}{lrrrrrrrrrrrrrrr}
\toprule
{} & \multicolumn{5}{l}{train\_canonical\_top\_1\_accuracy} & \multicolumn{5}{l}{val\_canonical\_top\_1\_accuracy} & \multicolumn{5}{l}{val\_diverse\_Rotation\_top\_1\_accuracy} \\
train\_prop\_to\_vary &                           0.05 &   0.25 &   0.50 &   0.75 &   0.95 &                         0.05 &   0.25 &   0.50 &   0.75 &   0.95 &                                0.05 &   0.25 &   0.50 &   0.75 &   0.95 \\
model                 &                                &        &        &        &        &                              &        &        &        &        &                                     &        &        &        &        \\
\midrule
CLIPPretrained        &                         91.70\% & 91.43\% & 92.42\% & 93.98\% & 96.28\% &                       81.48\% & 85.19\% & 85.19\% & 79.63\% & 55.56\% &                              28.64\% & 41.16\% & 44.72\% & 46.17\% & 48.85\% \\
MAEPretrained         &                         53.90\% & 57.24\% & 61.28\% & 68.44\% & 83.90\% &                       25.93\% & 44.44\% & 38.89\% & 42.59\% & 29.63\% &                               6.68\% &  7.93\% &  8.58\% &  8.63\% &  8.36\% \\
MLPMixerPretrained1k  &                         94.24\% & 93.76\% & 94.74\% & 95.72\% & 97.28\% &                       74.07\% & 74.07\% & 72.22\% & 70.37\% & 46.30\% &                              26.84\% & 39.47\% & 43.04\% & 46.51\% & 48.73\% \\
MLPMixerPretrained21k &                         94.49\% & 94.81\% & 95.03\% & 95.88\% & 97.13\% &                       79.63\% & 77.78\% & 77.78\% & 79.63\% & 72.22\% &                              36.90\% & 55.46\% & 61.57\% & 63.67\% & 65.50\% \\
ResNet50Pretrained1k  &                         94.72\% & 94.56\% & 94.89\% & 95.74\% & 97.18\% &                       85.19\% & 87.04\% & 85.19\% & 81.48\% & 68.52\% &                              34.44\% & 46.22\% & 49.90\% & 51.73\% & 53.24\% \\
ResNet50Pretrained21k &                         95.51\% & 94.96\% & 95.69\% & 96.12\% & 97.30\% &                       77.78\% & 75.93\% & 75.93\% & 72.22\% & 66.67\% &                              40.29\% & 57.68\% & 61.83\% & 63.91\% & 65.04\% \\
SimCLRPretrained      &                         95.74\% & 95.63\% & 96.16\% & 96.80\% & 97.51\% &                       81.48\% & 83.33\% & 83.33\% & 83.33\% & 62.96\% &                              32.94\% & 47.35\% & 52.88\% & 55.97\% & 56.91\% \\
ViTPretrained1k       &                         95.71\% & 95.76\% & 96.09\% & 96.79\% & 97.67\% &                       87.04\% & 79.63\% & 81.48\% & 79.63\% & 59.26\% &                              39.13\% & 55.09\% & 60.14\% & 64.14\% & 65.42\% \\
ViTPretrained21k      &                         95.23\% & 94.89\% & 95.68\% & 96.42\% & 97.45\% &                       85.19\% & 83.33\% & 83.33\% & 83.33\% & 66.67\% &                              38.25\% & 54.50\% & 58.34\% & 60.94\% & 62.28\% \\
iBotPretrained1k      &                         96.39\% & 96.22\% & 96.41\% & 96.96\% & 97.56\% &                       83.33\% & 81.48\% & 81.48\% & 79.63\% & 72.22\% &                              34.62\% & 51.88\% & 58.88\% & 62.19\% & 63.11\% \\
iBotPretrained21k     &                         96.44\% & 96.09\% & 96.42\% & 96.92\% & 97.61\% &                       90.74\% & 88.89\% & 90.74\% & 87.04\% & 72.22\% &                              41.03\% & 57.48\% & 63.69\% & 65.49\% & 67.34\% \\
\bottomrule
\end{tabular}
 \end{adjustbox}

\end{table}

\begin{table}
\centering
\caption{Spot hue linear eval top-1 accuracy across multiple percentages of varying training instances.}
\label{app:linear eval_Spot huediverse_training}
\begin{adjustbox}{width=\textwidth}
\begin{tabular}{lrrrrrrrrrrrrrrr}
\toprule
{} & \multicolumn{5}{l}{train\_canonical\_top\_1\_accuracy} & \multicolumn{5}{l}{val\_canonical\_top\_1\_accuracy} & \multicolumn{5}{l}{val\_diverse\_Spot hue\_top\_1\_accuracy} \\
train\_prop\_to\_vary &                           0.05 &   0.25 &   0.50 &   0.75 &   0.95 &                         0.05 &   0.25 &   0.50 &   0.75 &   0.95 &                                0.05 &   0.25 &   0.50 &   0.75 &   0.95 \\
model                 &                                &        &        &        &        &                              &        &        &        &        &                                     &        &        &        &        \\
\midrule
CLIPPretrained        &                         92.49\% & 91.60\% & 92.46\% & 94.06\% & 96.32\% &                       77.78\% & 75.93\% & 70.37\% & 70.37\% & 59.26\% &                              39.61\% & 60.78\% & 66.72\% & 68.27\% & 68.64\% \\
MAEPretrained         &                         52.26\% & 55.43\% & 60.36\% & 67.78\% & 83.62\% &                       22.22\% & 31.48\% & 37.04\% & 37.04\% & 20.37\% &                              11.51\% & 15.09\% & 16.10\% & 18.91\% & 15.58\% \\
MLPMixerPretrained1k  &                         95.17\% & 94.20\% & 94.98\% & 95.86\% & 97.30\% &                       79.63\% & 77.78\% & 70.37\% & 66.67\% & 57.04\% &                              40.09\% & 56.90\% & 64.43\% & 67.35\% & 68.93\% \\
MLPMixerPretrained21k &                         94.89\% & 95.29\% & 95.44\% & 96.13\% & 97.20\% &                       81.48\% & 79.63\% & 72.22\% & 74.07\% & 76.30\% &                              44.68\% & 69.30\% & 73.32\% & 76.12\% & 77.75\% \\
ResNet50Pretrained1k  &                         95.26\% & 94.81\% & 95.03\% & 95.80\% & 97.23\% &                       87.04\% & 88.89\% & 87.04\% & 83.33\% & 77.78\% &                              44.08\% & 62.96\% & 68.67\% & 71.81\% & 72.54\% \\
ResNet50Pretrained21k &                         95.91\% & 95.45\% & 96.00\% & 96.28\% & 97.41\% &                       77.78\% & 75.93\% & 75.93\% & 72.22\% & 71.11\% &                              44.22\% & 67.22\% & 70.95\% & 72.38\% & 74.39\% \\
SimCLRPretrained      &                         96.30\% & 95.91\% & 96.27\% & 96.84\% & 97.59\% &                       79.63\% & 75.93\% & 74.07\% & 74.07\% & 66.67\% &                              43.36\% & 63.22\% & 71.32\% & 73.72\% & 72.26\% \\
ViTPretrained1k       &                         95.99\% & 96.36\% & 96.54\% & 96.95\% & 97.80\% &                       90.74\% & 87.04\% & 83.33\% & 81.48\% & 74.07\% &                              52.53\% & 69.53\% & 71.81\% & 76.01\% & 77.28\% \\
ViTPretrained21k      &                         95.57\% & 95.39\% & 95.84\% & 96.51\% & 97.59\% &                       85.19\% & 81.48\% & 83.33\% & 77.78\% & 75.93\% &                              46.01\% & 67.50\% & 71.97\% & 75.75\% & 77.06\% \\
iBotPretrained1k      &                         96.50\% & 96.55\% & 96.74\% & 97.12\% & 97.70\% &                       79.63\% & 81.48\% & 81.48\% & 77.78\% & 74.07\% &                              42.32\% & 68.61\% & 72.75\% & 76.28\% & 78.04\% \\
iBotPretrained21k     &                         97.01\% & 96.42\% & 96.65\% & 97.04\% & 97.64\% &                       88.89\% & 87.04\% & 83.33\% & 83.33\% & 75.93\% &                              48.83\% & 73.37\% & 78.09\% & 80.39\% & 81.55\% \\
\bottomrule
\end{tabular}
 \end{adjustbox}

\end{table}

\begin{table}
\centering
\caption{Scale linear eval top-1 accuracy across multiple percentages of varying training instances}
\label{app:linear eval_Scalediverse_training}
\begin{adjustbox}{width=\textwidth}
\begin{tabular}{lrrrrrrrrrrrrrrr}
\toprule
{} & \multicolumn{5}{l}{train\_canonical\_top\_1\_accuracy} & \multicolumn{5}{l}{val\_canonical\_top\_1\_accuracy} & \multicolumn{5}{l}{val\_diverse\_Scale\_top\_1\_accuracy} \\
train\_prop\_to\_vary &                           0.05 &   0.25 &   0.50 &   0.75 &   0.95 &                         0.05 &   0.25 &   0.50 &   0.75 &   0.95 &                             0.05 &   0.25 &   0.50 &   0.75 &   0.95 \\
model                 &                                &        &        &        &        &                              &        &        &        &        &                                  &        &        &        &        \\
\midrule
CLIPPretrained        &                         91.81\% & 91.51\% & 92.26\% & 93.90\% & 96.25\% &                       79.63\% & 72.22\% & 74.07\% & 70.37\% & 61.11\% &                           36.80\% & 51.29\% & 57.05\% & 56.99\% & 58.80\% \\
MAEPretrained         &                         51.98\% & 56.15\% & 60.87\% & 68.07\% & 84.12\% &                       25.93\% & 35.19\% & 33.33\% & 37.04\% & 35.19\% &                            7.01\% & 10.74\% & 11.05\% & 11.29\% & 11.44\% \\
MLPMixerPretrained1k  &                         94.27\% & 93.55\% & 94.47\% & 95.59\% & 97.17\% &                       79.63\% & 77.78\% & 70.37\% & 68.52\% & 53.70\% &                           32.50\% & 46.15\% & 51.66\% & 55.30\% & 56.62\% \\
MLPMixerPretrained21k &                         94.86\% & 94.98\% & 95.08\% & 95.93\% & 97.09\% &                       79.63\% & 79.63\% & 75.93\% & 75.93\% & 74.07\% &                           40.55\% & 60.87\% & 66.63\% & 67.52\% & 70.03\% \\
ResNet50Pretrained1k  &                         94.89\% & 94.57\% & 94.84\% & 95.66\% & 97.16\% &                       87.04\% & 90.74\% & 90.74\% & 83.33\% & 72.22\% &                           39.22\% & 54.23\% & 59.10\% & 61.27\% & 60.89\% \\
ResNet50Pretrained21k &                         95.51\% & 95.28\% & 95.77\% & 96.08\% & 97.35\% &                       81.48\% & 77.78\% & 77.78\% & 74.07\% & 74.07\% &                           41.11\% & 60.06\% & 66.41\% & 69.69\% & 70.33\% \\
SimCLRPretrained      &                         95.91\% & 95.53\% & 96.09\% & 96.66\% & 97.48\% &                       83.33\% & 77.41\% & 77.78\% & 72.22\% & 62.96\% &                           39.79\% & 54.93\% & 61.81\% & 62.93\% & 62.96\% \\
ViTPretrained1k       &                         95.65\% & 96.01\% & 96.18\% & 96.69\% & 97.69\% &                       87.04\% & 83.33\% & 79.63\% & 75.93\% & 72.22\% &                           44.10\% & 58.13\% & 63.91\% & 67.85\% & 68.82\% \\
ViTPretrained21k      &                         95.06\% & 94.87\% & 95.47\% & 96.30\% & 97.38\% &                       83.33\% & 83.33\% & 83.33\% & 81.48\% & 72.22\% &                           41.40\% & 58.53\% & 63.25\% & 65.43\% & 65.14\% \\
iBotPretrained1k      &                         96.58\% & 96.37\% & 96.48\% & 96.98\% & 97.60\% &                       84.07\% & 77.78\% & 79.63\% & 77.78\% & 66.67\% &                           41.34\% & 58.93\% & 67.24\% & 68.66\% & 69.08\% \\
iBotPretrained21k     &                         96.92\% & 96.18\% & 96.39\% & 96.86\% & 97.53\% &                       85.19\% & 77.78\% & 81.48\% & 81.48\% & 75.93\% &                           44.59\% & 63.11\% & 68.90\% & 71.45\% & 70.82\% \\
\bottomrule
\end{tabular}
 \end{adjustbox}

\end{table}

\begin{table}
\centering
\caption{Background path linear eval top-1 accuracy across multiple percentages of varying training instances}
\label{app:linear eval_Background pathdiverse_training}
\begin{adjustbox}{width=\textwidth}
\begin{tabular}{lrrrrrrrrrrrrrrr}
\toprule
{} & \multicolumn{5}{l}{train\_canonical\_top\_1\_accuracy} & \multicolumn{5}{l}{val\_canonical\_top\_1\_accuracy} & \multicolumn{5}{l}{val\_diverse\_Background path\_top\_1\_accuracy} \\
train\_prop\_to\_vary &                           0.05 &   0.25 &   0.50 &   0.75 &   0.95 &                         0.05 &   0.25 &   0.50 &   0.75 &   0.95 &                                       0.05 &   0.25 &   0.50 &   0.75 &   0.95 \\
model                 &                                &        &        &        &        &                              &        &        &        &        &                                            &        &        &        &        \\
\midrule
CLIPPretrained        &                         90.37\% & 91.88\% & 92.73\% & 94.36\% & 96.70\% &                       85.19\% & 77.78\% & 77.78\% & 77.78\% & 66.67\% &                                     41.30\% & 54.70\% & 59.96\% & 61.75\% & 63.41\% \\
MAEPretrained         &                         51.51\% & 56.16\% & 62.31\% & 67.08\% & 84.32\% &                       27.78\% & 31.48\% & 37.04\% & 37.04\% & 29.63\% &                                      9.47\% & 12.21\% & 11.84\% & 13.30\% & 12.67\% \\
MLPMixerPretrained1k  &                         92.84\% & 93.82\% & 94.81\% & 95.41\% & 97.38\% &                       75.93\% & 77.78\% & 72.22\% & 74.07\% & 61.11\% &                                     37.39\% & 49.19\% & 52.41\% & 55.75\% & 56.74\% \\
MLPMixerPretrained21k &                         95.42\% & 95.33\% & 95.62\% & 96.52\% & 97.55\% &                       83.33\% & 81.48\% & 75.93\% & 75.93\% & 77.78\% &                                     48.30\% & 64.37\% & 66.79\% & 70.15\% & 70.84\% \\
ResNet50Pretrained1k  &                         94.35\% & 95.12\% & 94.83\% & 95.93\% & 97.46\% &                       90.74\% & 92.59\% & 88.89\% & 88.89\% & 83.33\% &                                     44.89\% & 59.87\% & 64.15\% & 66.70\% & 67.29\% \\
ResNet50Pretrained21k &                         95.20\% & 96.40\% & 95.89\% & 96.65\% & 97.67\% &                       81.48\% & 79.63\% & 77.78\% & 77.78\% & 75.93\% &                                     51.21\% & 65.75\% & 69.47\% & 72.01\% & 73.90\% \\
SimCLRPretrained      &                         95.50\% & 95.98\% & 96.14\% & 96.45\% & 97.52\% &                       83.33\% & 83.33\% & 81.48\% & 77.78\% & 72.22\% &                                     44.33\% & 59.82\% & 65.39\% & 67.93\% & 68.93\% \\
ViTPretrained1k       &                         95.72\% & 96.42\% & 96.27\% & 96.57\% & 97.95\% &                       94.44\% & 88.89\% & 88.89\% & 87.04\% & 77.78\% &                                     50.62\% & 64.09\% & 67.14\% & 72.33\% & 73.19\% \\
ViTPretrained21k      &                         94.91\% & 95.22\% & 96.09\% & 96.45\% & 97.71\% &                       83.33\% & 83.33\% & 83.33\% & 83.33\% & 77.78\% &                                     49.36\% & 64.21\% & 67.42\% & 70.77\% & 72.13\% \\
iBotPretrained1k      &                         96.42\% & 96.58\% & 96.60\% & 97.40\% & 97.28\% &                       88.89\% & 83.33\% & 87.04\% & 81.48\% & 79.63\% &                                     44.58\% & 62.59\% & 68.10\% & 71.20\% & 73.36\% \\
iBotPretrained21k     &                         96.16\% & 96.14\% & 96.38\% & 97.37\% & 97.27\% &                       88.89\% & 90.74\% & 90.74\% & 83.33\% & 81.48\% &                                     51.20\% & 68.58\% & 71.57\% & 74.62\% & 75.94\% \\
\bottomrule
\end{tabular}
 \end{adjustbox}

\end{table}

\begin{figure}[h]
\begin{subfigure}{\textwidth}
\includegraphics[width=\textwidth]{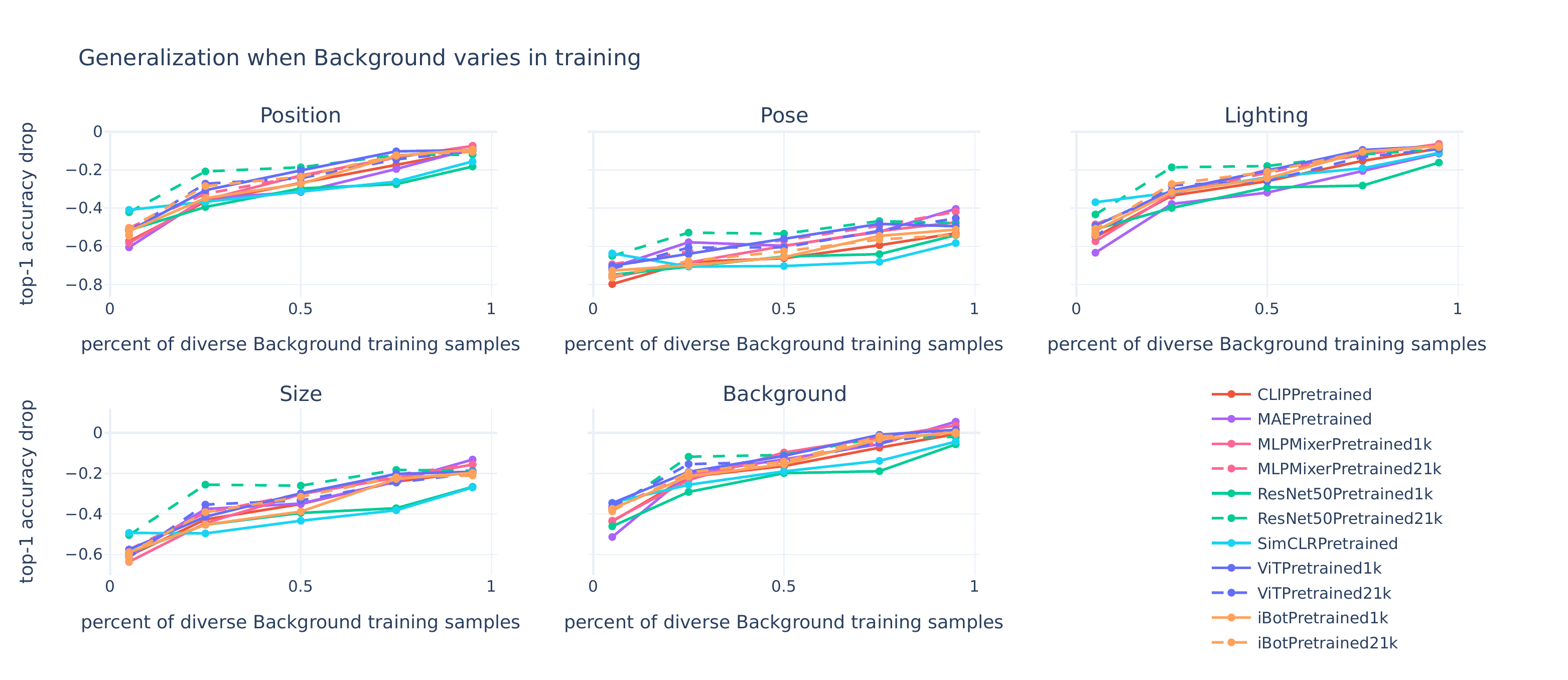} 
\end{subfigure}
\begin{subfigure}{\textwidth}
\includegraphics[width=\textwidth]{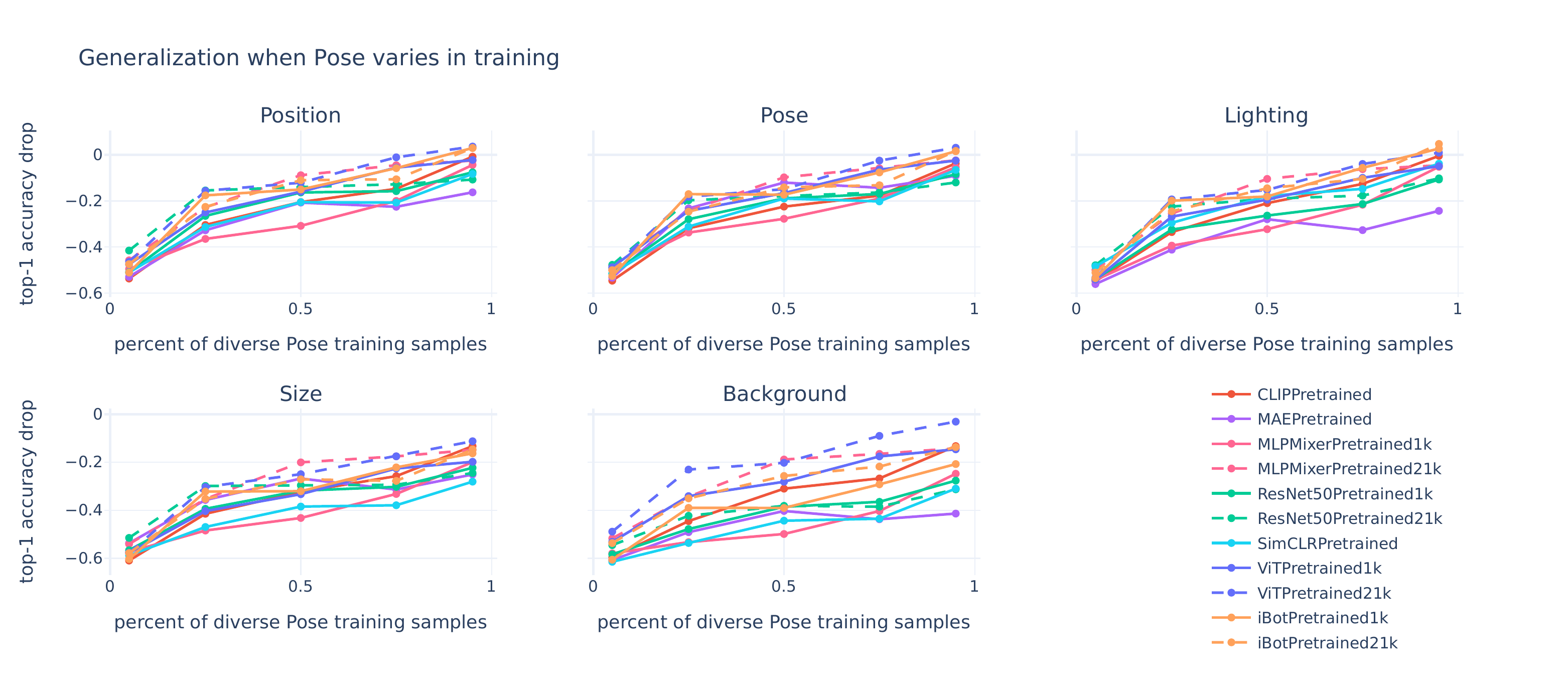}
\end{subfigure}
\begin{subfigure}{\textwidth}
\includegraphics[width=\textwidth]{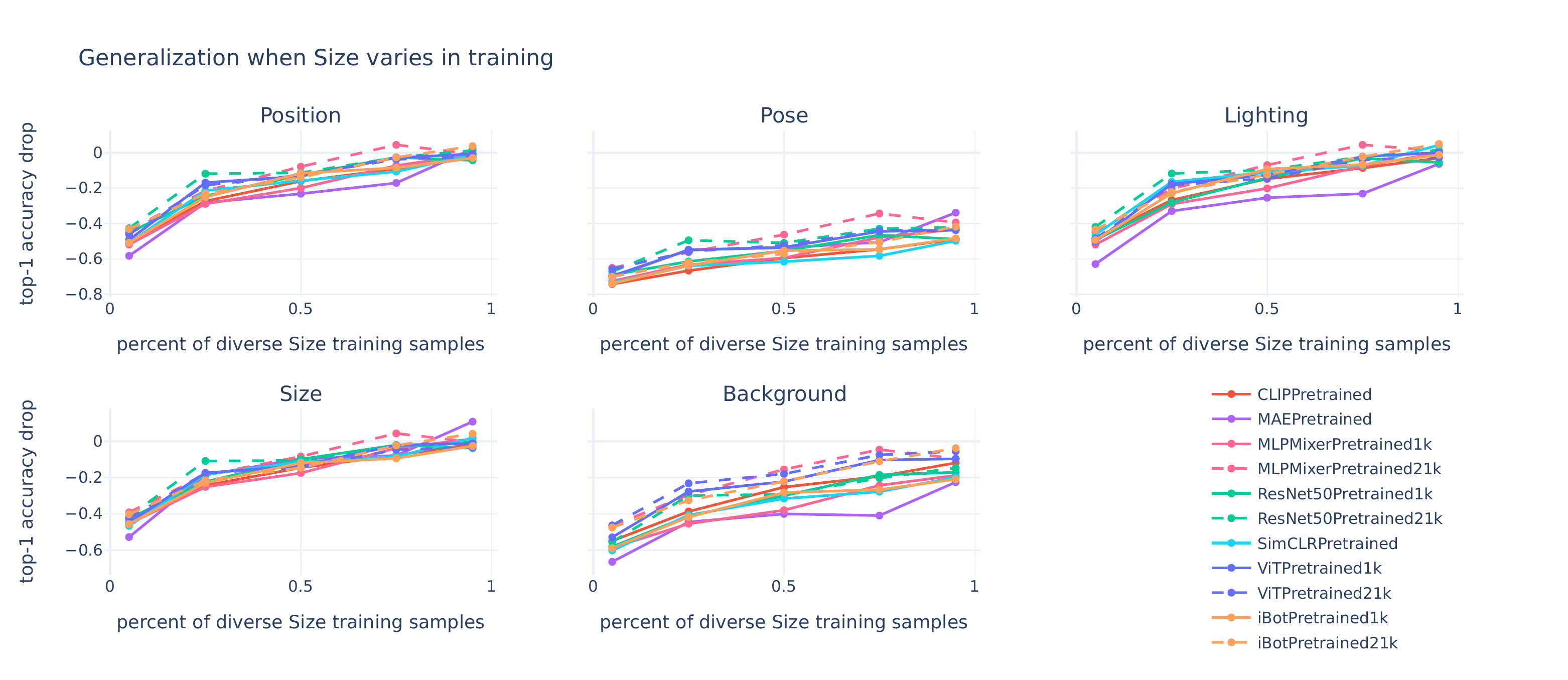}
\end{subfigure}
\caption{Finetuning Effect of Variability in Training (part 1)}
\label{app_fig:finetuning_diverse_factor}
\end{figure}

\begin{figure}[h]
\begin{subfigure}{\textwidth}
\includegraphics[width=\textwidth]{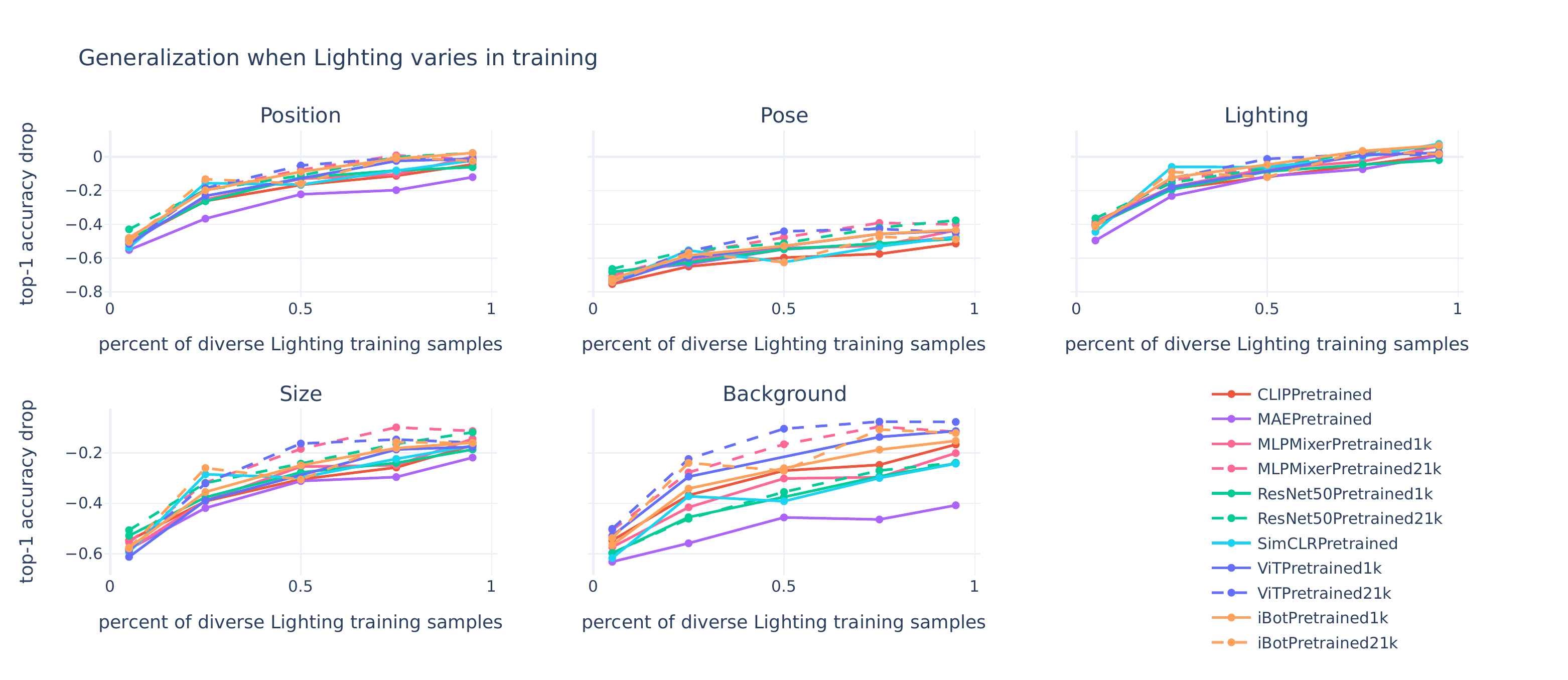}
\end{subfigure}
\begin{subfigure}{\textwidth}
\includegraphics[width=\textwidth]{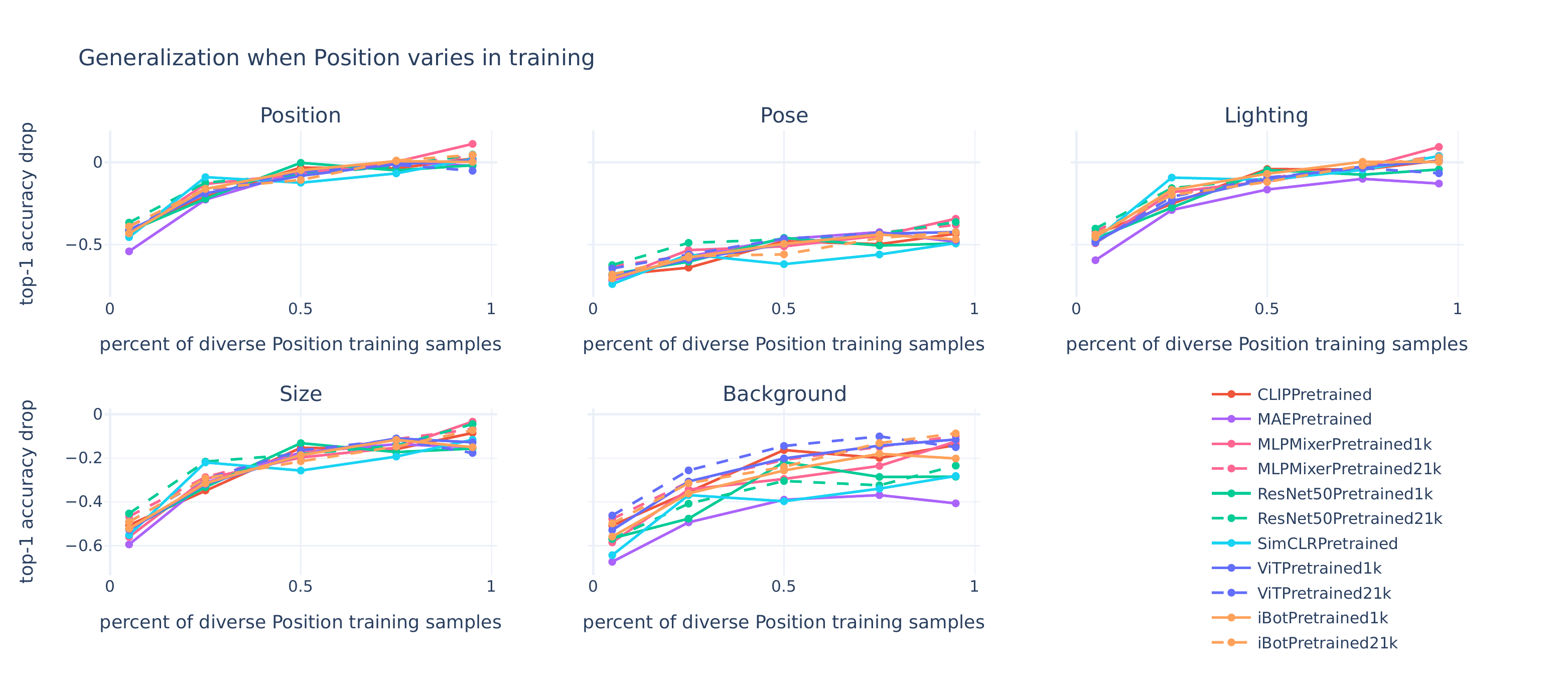}
\end{subfigure}
\caption{Finetuning Effect of Variability in Training (part 2)}
\label{app_fig:finetuning_diverse_factor_2}
\end{figure}

\begin{table}
\centering
\caption{Position finetuning top-1 accuracy across multiple percentages of varying training instances}
\label{finetuning_Translationdiverse_training}
\begin{adjustbox}{width=\textwidth}
\begin{tabular}{lrrrrrrrrrrrrrrr}
\toprule
{} & \multicolumn{5}{l}{train\_canonical\_top\_1\_accuracy} & \multicolumn{5}{l}{val\_canonical\_top\_1\_accuracy} & \multicolumn{5}{l}{val\_diverse\_Translation\_top\_1\_accuracy} \\
train\_prop\_to\_vary &                           0.05 &   0.25 &   0.50 &   0.75 &   0.95 &                         0.05 &   0.25 &   0.50 &   0.75 &   0.95 &                                   0.05 &   0.25 &   0.50 &   0.75 &   0.95 \\
model                 &                                &        &        &        &        &                              &        &        &        &        &                                        &        &        &        &        \\
\midrule
CLIPPretrained        &                         97.49\% & 97.00\% & 97.25\% & 97.55\% & 96.80\% &                       87.04\% & 90.74\% & 77.78\% & 81.48\% & 75.93\% &                                 45.33\% & 69.74\% & 74.53\% & 77.72\% & 78.19\% \\
MAEPretrained         &                         96.75\% & 96.63\% & 97.20\% & 97.46\% & 97.91\% &                       83.33\% & 74.07\% & 66.67\% & 64.81\% & 72.22\% &                                 29.17\% & 51.35\% & 61.05\% & 65.16\% & 70.07\% \\
MLPMixerPretrained1k  &                         96.95\% & 96.73\% & 97.17\% & 97.24\% & 97.88\% &                       88.89\% & 77.78\% & 77.78\% & 74.07\% & 64.81\% &                                 44.65\% & 64.56\% & 70.67\% & 74.34\% & 75.95\% \\
MLPMixerPretrained21k &                         97.71\% & 97.69\% & 97.90\% & 97.98\% & 98.35\% &                       85.19\% & 88.89\% & 85.19\% & 81.48\% & 77.78\% &                                 46.53\% & 70.54\% & 77.27\% & 79.78\% & 81.68\% \\
ResNet50Pretrained1k  &                         97.97\% & 97.92\% & 97.91\% & 97.86\% & 98.27\% &                       87.04\% & 87.04\% & 72.22\% & 81.48\% & 81.48\% &                                 45.05\% & 64.83\% & 71.87\% & 76.56\% & 79.63\% \\
ResNet50Pretrained21k &                         97.54\% & 97.62\% & 97.74\% & 97.69\% & 98.20\% &                       88.89\% & 83.33\% & 83.33\% & 83.33\% & 77.78\% &                                 52.22\% & 70.96\% & 75.78\% & 81.10\% & 82.45\% \\
SimCLRPretrained      &                         97.40\% & 97.57\% & 97.68\% & 97.81\% & 98.12\% &                       90.74\% & 77.78\% & 87.04\% & 83.33\% & 77.78\% &                                 45.21\% & 68.73\% & 74.59\% & 76.55\% & 79.86\% \\
ViTPretrained1k       &                         97.80\% & 97.88\% & 98.00\% & 97.92\% & 98.28\% &                       90.74\% & 90.74\% & 85.19\% & 81.48\% & 81.48\% &                                 49.30\% & 71.82\% & 76.95\% & 80.39\% & 82.58\% \\
ViTPretrained21k      &                         97.80\% & 97.59\% & 97.89\% & 97.91\% & 98.25\% &                       87.04\% & 88.89\% & 81.48\% & 81.48\% & 87.04\% &                                 45.83\% & 71.80\% & 75.51\% & 79.96\% & 81.86\% \\
iBotPretrained1k      &                         97.77\% & 97.55\% & 97.64\% & 97.77\% & 98.06\% &                       88.89\% & 85.19\% & 79.63\% & 75.93\% & 79.63\% &                                 45.69\% & 68.94\% & 74.76\% & 76.63\% & 79.83\% \\
iBotPretrained21k     &                         97.97\% & 97.83\% & 97.88\% & 97.92\% & 98.13\% &                       88.89\% & 87.04\% & 88.89\% & 81.48\% & 79.63\% &                                 49.84\% & 70.97\% & 78.13\% & 82.53\% & 84.07\% \\
\bottomrule
\end{tabular}
 \end{adjustbox}

\end{table}

\begin{table}
\centering
\caption{Pose finetuning top-1 accuracy across multiple percentages of varying training instances}
\label{finetuning_Rotationdiverse_training}
\begin{adjustbox}{width=\textwidth}
\begin{tabular}{lrrrrrrrrrrrrrrr}
\toprule
{} & \multicolumn{5}{l}{train\_canonical\_top\_1\_accuracy} & \multicolumn{5}{l}{val\_canonical\_top\_1\_accuracy} & \multicolumn{5}{l}{val\_diverse\_Rotation\_top\_1\_accuracy} \\
train\_prop\_to\_vary &                           0.05 &   0.25 &   0.50 &   0.75 &   0.95 &                         0.05 &   0.25 &   0.50 &   0.75 &   0.95 &                                0.05 &   0.25 &   0.50 &   0.75 &   0.95 \\
model                 &                                &        &        &        &        &                              &        &        &        &        &                                     &        &        &        &        \\
\midrule
CLIPPretrained        &                         97.60\% & 97.01\% & 97.30\% & 97.58\% & 97.97\% &                       88.89\% & 88.89\% & 85.19\% & 83.33\% & 72.22\% &                              34.29\% & 57.01\% & 62.67\% & 65.58\% & 68.45\% \\
MAEPretrained         &                         96.87\% & 96.75\% & 97.25\% & 97.52\% & 97.95\% &                       75.93\% & 68.52\% & 62.96\% & 68.52\% & 62.96\% &                              22.64\% & 45.24\% & 50.87\% & 54.18\% & 53.82\% \\
MLPMixerPretrained1k  &                         96.75\% & 96.58\% & 97.08\% & 97.25\% & 97.86\% &                       83.33\% & 85.19\% & 83.33\% & 77.78\% & 64.81\% &                              33.29\% & 51.43\% & 55.55\% & 58.78\% & 59.35\% \\
MLPMixerPretrained21k &                         97.68\% & 97.65\% & 97.90\% & 97.90\% & 98.35\% &                       87.04\% & 87.04\% & 77.78\% & 77.78\% & 75.93\% &                              38.93\% & 62.76\% & 67.97\% & 71.90\% & 73.47\% \\
ResNet50Pretrained1k  &                         98.05\% & 97.81\% & 97.89\% & 97.93\% & 98.24\% &                       84.81\% & 81.48\% & 81.48\% & 81.48\% & 75.93\% &                              33.29\% & 53.51\% & 62.56\% & 64.45\% & 67.47\% \\
ResNet50Pretrained21k &                         97.52\% & 97.45\% & 97.65\% & 97.61\% & 98.18\% &                       85.19\% & 79.63\% & 83.33\% & 87.04\% & 85.19\% &                              37.45\% & 59.85\% & 65.39\% & 70.66\% & 73.13\% \\
SimCLRPretrained      &                         97.37\% & 97.43\% & 97.53\% & 97.74\% & 98.07\% &                       87.04\% & 87.04\% & 83.33\% & 87.04\% & 75.93\% &                              35.50\% & 55.87\% & 64.34\% & 66.79\% & 69.34\% \\
ViTPretrained1k       &                         97.94\% & 97.87\% & 97.98\% & 97.95\% & 98.31\% &                       88.89\% & 87.04\% & 85.19\% & 77.78\% & 75.93\% &                              40.13\% & 62.66\% & 68.53\% & 71.32\% & 73.36\% \\
ViTPretrained21k      &                         97.68\% & 97.61\% & 97.90\% & 97.97\% & 98.27\% &                       88.89\% & 79.63\% & 83.33\% & 74.07\% & 70.74\% &                              39.75\% & 61.42\% & 68.39\% & 71.51\% & 73.76\% \\
iBotPretrained1k      &                         97.49\% & 97.55\% & 97.56\% & 97.75\% & 98.02\% &                       88.89\% & 77.78\% & 83.33\% & 75.93\% & 68.52\% &                              36.30\% & 60.69\% & 65.98\% & 68.22\% & 70.00\% \\
iBotPretrained21k     &                         98.00\% & 97.79\% & 97.96\% & 98.01\% & 98.19\% &                       88.89\% & 87.04\% & 83.33\% & 85.19\% & 72.22\% &                              38.86\% & 62.35\% & 69.18\% & 71.96\% & 73.84\% \\
\bottomrule
\end{tabular}
 \end{adjustbox}

\end{table}

\begin{table}
\centering
\caption{Spot hue finetuning top-1 accuracy across multiple percentages of varying training instances}
\label{finetuning_Spot huediverse_training}
\begin{adjustbox}{width=\textwidth}
\begin{tabular}{lrrrrrrrrrrrrrrr}
\toprule
{} & \multicolumn{5}{l}{train\_canonical\_top\_1\_accuracy} & \multicolumn{5}{l}{val\_canonical\_top\_1\_accuracy} & \multicolumn{5}{l}{val\_diverse\_Spot hue\_top\_1\_accuracy} \\
train\_prop\_to\_vary &                           0.05 &   0.25 &   0.50 &   0.75 &   0.95 &                         0.05 &   0.25 &   0.50 &   0.75 &   0.95 &                                0.05 &   0.25 &   0.50 &   0.75 &   0.95 \\
model                 &                                &        &        &        &        &                              &        &        &        &        &                                     &        &        &        &        \\
\midrule
CLIPPretrained        &                         97.66\% & 97.13\% & 97.40\% & 97.62\% & 97.98\% &                       90.74\% & 88.89\% & 87.04\% & 87.04\% & 81.48\% &                              49.99\% & 70.06\% & 75.12\% & 82.36\% & 82.48\% \\
MAEPretrained         &                         97.04\% & 96.67\% & 97.20\% & 97.41\% & 97.95\% &                       79.63\% & 77.78\% & 72.22\% & 74.07\% & 68.52\% &                              30.10\% & 54.63\% & 60.74\% & 66.78\% & 69.34\% \\
MLPMixerPretrained1k  &                         97.32\% & 96.92\% & 97.20\% & 97.36\% & 97.89\% &                       87.04\% & 83.33\% & 77.78\% & 79.63\% & 72.22\% &                              48.80\% & 65.60\% & 70.82\% & 76.88\% & 78.59\% \\
MLPMixerPretrained21k &                         97.80\% & 97.83\% & 97.99\% & 97.99\% & 98.42\% &                       88.89\% & 87.04\% & 81.48\% & 75.93\% & 79.63\% &                              49.81\% & 73.15\% & 75.23\% & 79.01\% & 82.35\% \\
ResNet50Pretrained1k  &                         98.02\% & 97.99\% & 98.03\% & 97.98\% & 98.28\% &                       88.89\% & 87.04\% & 83.33\% & 81.48\% & 77.78\% &                              48.48\% & 67.72\% & 74.65\% & 76.90\% & 75.83\% \\
ResNet50Pretrained21k &                         97.68\% & 97.60\% & 97.87\% & 97.79\% & 98.24\% &                       90.74\% & 87.04\% & 83.33\% & 77.78\% & 75.93\% &                              54.35\% & 71.69\% & 76.70\% & 81.03\% & 81.86\% \\
SimCLRPretrained      &                         97.60\% & 97.61\% & 97.72\% & 97.87\% & 98.17\% &                       90.00\% & 76.30\% & 85.19\% & 77.78\% & 74.07\% &                              45.58\% & 70.36\% & 79.09\% & 78.06\% & 81.75\% \\
ViTPretrained1k       &                         97.68\% & 97.93\% &    NaN & 98.00\% & 98.37\% &                       94.44\% & 88.89\% &    NaN & 81.48\% & 81.48\% &                              54.38\% & 70.94\% &    NaN & 82.53\% & 83.94\% \\
ViTPretrained21k      &                         97.77\% & 97.68\% & 97.94\% & 98.00\% & 98.24\% &                       92.59\% & 85.19\% & 77.78\% & 78.15\% & 81.48\% &                              52.49\% & 72.55\% & 76.58\% & 79.75\% & 82.39\% \\
iBotPretrained1k      &                         97.91\% & 97.58\% & 97.76\% & 97.88\% & 98.08\% &                       88.89\% & 81.48\% & 79.63\% & 75.93\% & 74.07\% &                              48.67\% & 69.24\% & 75.01\% & 79.44\% & 80.81\% \\
iBotPretrained21k     &                         98.25\% & 97.87\% & 97.88\% & 97.96\% & 98.13\% &                       90.74\% & 83.33\% & 92.59\% & 79.63\% & 83.33\% &                              49.39\% & 74.42\% & 80.67\% & 83.05\% & 85.02\% \\
\bottomrule
\end{tabular}
 \end{adjustbox}

\end{table}

\begin{table}
\centering
\caption{Scale finetuning top-1 accuracy across multiple percentages of varying training instances}
\label{finetuning_Scalediverse_training}
\begin{adjustbox}{width=\textwidth}
\begin{tabular}{lrrrrrrrrrrrrrrr}
\toprule
{} & \multicolumn{5}{l}{train\_canonical\_top\_1\_accuracy} & \multicolumn{5}{l}{val\_canonical\_top\_1\_accuracy} & \multicolumn{5}{l}{val\_diverse\_Scale\_top\_1\_accuracy} \\
train\_prop\_to\_vary &                           0.05 &   0.25 &   0.50 &   0.75 &   0.95 &                         0.05 &   0.25 &   0.50 &   0.75 &   0.95 &                             0.05 &   0.25 &   0.50 &   0.75 &   0.95 \\
model                 &                                &        &        &        &        &                              &        &        &        &        &                                  &        &        &        &        \\
\midrule
CLIPPretrained        &                         97.68\% & 97.08\% & 97.39\% & 97.63\% & 97.95\% &                       90.74\% & 90.74\% & 87.04\% & 83.33\% & 79.63\% &                           45.47\% & 66.39\% & 72.46\% & 75.63\% & 78.05\% \\
MAEPretrained         &                         96.81\% & 96.64\% & 97.19\% & 97.40\% & 97.91\% &                       81.48\% & 70.37\% & 70.37\% & 70.37\% & 53.70\% &                           28.70\% & 51.73\% & 60.66\% & 62.49\% & 64.47\% \\
MLPMixerPretrained1k  &                         97.23\% & 96.60\% & 97.07\% & 97.24\% & 97.82\% &                       87.04\% & 85.19\% & 83.33\% & 74.07\% & 70.37\% &                           41.62\% & 60.01\% & 65.78\% & 70.24\% & 71.64\% \\
MLPMixerPretrained21k &                         97.80\% & 97.80\% & 97.96\% & 97.95\% & 98.37\% &                       85.19\% & 87.04\% & 81.48\% & 72.22\% & 77.78\% &                           46.02\% & 68.37\% & 73.08\% & 76.51\% & 77.38\% \\
ResNet50Pretrained1k  &                         97.88\% & 97.94\% & 97.95\% & 97.89\% & 98.24\% &                       87.04\% & 85.19\% & 81.48\% & 75.93\% & 79.63\% &                           44.47\% & 62.86\% & 71.55\% & 73.81\% & 75.80\% \\
ResNet50Pretrained21k &                         97.40\% & 97.58\% & 97.84\% & 97.72\% & 98.18\% &                       90.74\% & 79.63\% & 81.48\% & 79.63\% & 79.63\% &                           49.37\% & 68.72\% & 70.89\% & 76.85\% & 79.18\% \\
SimCLRPretrained      &                         97.57\% & 97.54\% & 97.65\% & 97.83\% & 98.10\% &                       88.89\% & 83.33\% & 83.33\% & 83.33\% & 74.44\% &                           42.25\% & 65.04\% & 71.88\% & 74.89\% & 76.25\% \\
ViTPretrained1k       &                         97.80\% & 97.92\% & 98.05\% & 97.92\% & 98.34\% &                       88.89\% & 83.33\% & 85.19\% & 79.63\% & 79.63\% &                           44.17\% & 65.86\% & 71.66\% & 77.46\% & 78.46\% \\
ViTPretrained21k      &                         97.77\% & 97.71\% & 97.85\% & 97.99\% & 98.24\% &                       85.19\% & 85.19\% & 85.19\% & 79.63\% & 79.63\% &                           42.63\% & 67.71\% & 70.61\% & 74.96\% & 76.14\% \\
iBotPretrained1k      &                         97.85\% & 97.61\% & 97.74\% & 97.79\% & 98.04\% &                       90.74\% & 87.04\% & 83.33\% & 83.33\% & 79.63\% &                           45.14\% & 64.09\% & 71.34\% & 73.90\% & 76.99\% \\
iBotPretrained21k     &                         97.88\% & 97.75\% & 97.86\% & 97.97\% & 98.12\% &                       88.89\% & 87.04\% & 87.04\% & 81.48\% & 75.93\% &                           48.70\% & 65.52\% & 72.39\% & 79.16\% & 80.04\% \\
\bottomrule
\end{tabular}
 \end{adjustbox}

\end{table}

\begin{table}
\centering
\caption{Background path finetuning top-1 accuracy across multiple percentages of varying training instances}
\label{finetuning_Background pathdiverse_training}
\begin{adjustbox}{width=\textwidth}
\begin{tabular}{lrrrrrrrrrrrrrrr}
\toprule
{} & \multicolumn{5}{l}{train\_canonical\_top\_1\_accuracy} & \multicolumn{5}{l}{val\_canonical\_top\_1\_accuracy} & \multicolumn{5}{l}{val\_diverse\_Background path\_top\_1\_accuracy} \\
train\_prop\_to\_vary &                           0.05 &   0.25 &   0.50 &   0.75 &   0.95 &                         0.05 &   0.25 &   0.50 &   0.75 &   0.95 &                                       0.05 &   0.25 &   0.50 &   0.75 &   0.95 \\
model                 &                                &        &        &        &        &                              &        &        &        &        &                                            &        &        &        &        \\
\midrule
CLIPPretrained        &                         96.49\% & 97.05\% & 97.52\% & 97.81\% & 98.01\% &                       94.44\% & 88.89\% & 92.59\% & 87.04\% & 81.48\% &                                     50.79\% & 67.24\% & 76.25\% & 79.79\% & 80.92\% \\
MAEPretrained         &                         96.20\% & 96.61\% & 97.31\% & 97.63\% & 97.99\% &                       81.85\% & 72.22\% & 77.78\% & 72.22\% & 62.96\% &                                     30.49\% & 52.04\% & 64.77\% & 66.67\% & 68.56\% \\
MLPMixerPretrained1k  &                         96.16\% & 97.03\% & 97.13\% & 97.66\% & 97.76\% &                       90.74\% & 87.04\% & 81.48\% & 75.93\% & 72.22\% &                                     47.41\% & 63.84\% & 71.85\% & 73.76\% & 75.96\% \\
MLPMixerPretrained21k &                         98.08\% & 98.15\% & 98.09\% & 98.33\% & 98.71\% &                       88.89\% & 92.59\% & 90.74\% & 85.19\% & 79.63\% &                                     51.84\% & 72.00\% & 77.05\% & 80.46\% & 81.10\% \\
ResNet50Pretrained1k  &                         97.71\% & 97.76\% & 97.95\% & 98.04\% & 98.38\% &                       92.59\% & 92.59\% & 90.74\% & 92.59\% & 83.33\% &                                     46.50\% & 63.48\% & 70.90\% & 73.73\% & 77.72\% \\
ResNet50Pretrained21k &                         97.38\% & 98.09\% & 97.72\% & 98.33\% & 98.34\% &                       88.89\% & 83.33\% & 87.04\% & 81.48\% & 85.19\% &                                     50.87\% & 71.59\% & 76.21\% & 79.13\% & 83.24\% \\
SimCLRPretrained      &                         97.38\% & 97.31\% & 97.77\% & 97.55\% & 98.05\% &                       77.78\% & 87.04\% & 88.89\% & 90.74\% & 83.33\% &                                     43.32\% & 61.47\% & 69.94\% & 76.99\% & 79.07\% \\
ViTPretrained1k       &                         98.12\% & 97.76\% & 97.89\% & 97.73\% & 98.44\% &                       88.89\% & 90.74\% & 87.04\% & 81.48\% & 83.33\% &                                     54.25\% & 71.56\% & 75.73\% & 80.58\% & 84.92\% \\
ViTPretrained21k      &                         97.77\% & 97.49\% & 97.95\% & 98.04\% & 98.37\% &                       91.67\% & 88.89\% & 90.74\% & 87.04\% & 81.48\% &                                     55.17\% & 73.53\% & 76.50\% & 82.28\% & 81.94\% \\
iBotPretrained1k      &                         97.47\% & 97.44\% & 97.61\% & 98.15\% & 97.86\% &                       88.89\% & 92.59\% & 90.74\% & 81.48\% & 79.63\% &                                     51.19\% & 71.17\% & 75.32\% & 78.37\% & 79.93\% \\
iBotPretrained21k     &                         97.69\% & 97.91\% & 97.77\% & 98.17\% & 97.93\% &                       93.52\% & 92.59\% & 90.74\% & 85.19\% & 83.33\% &                                     55.00\% & 73.11\% & 76.62\% & 83.56\% & 82.90\% \\
\bottomrule
\end{tabular}
 \end{adjustbox}

\end{table}

\section{All Factor Gaps}

\begin{figure}[t]
\includegraphics[width=\textwidth]{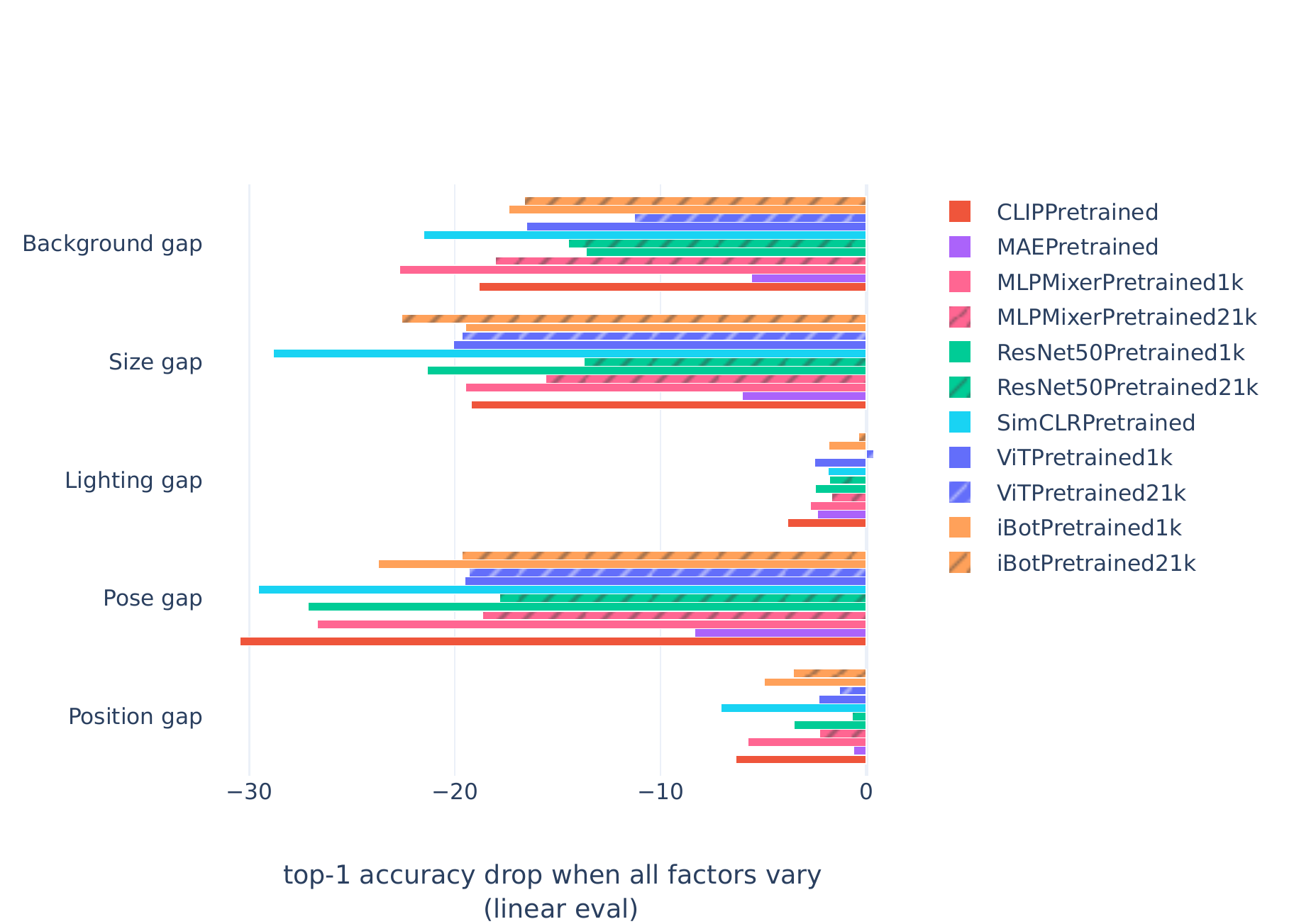}
    \centering
    \caption{Generalization gaps when all factors vary during training with linear evaluation}
    \label{fig:all_factor_linear}
\end{figure}

\begin{figure}[t]
\includegraphics[width=\textwidth]{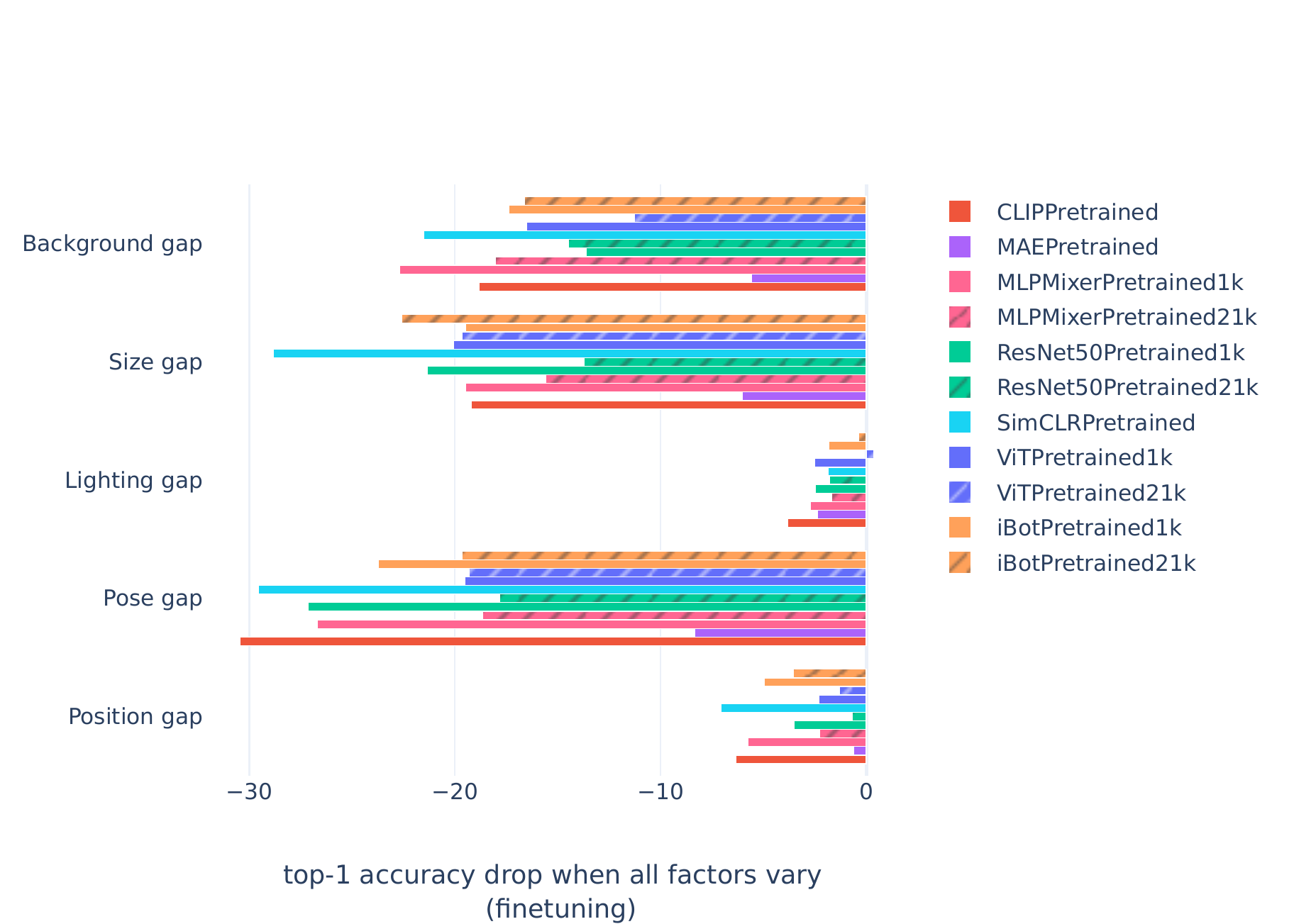}
    \centering
    \caption{Generalization gaps when all factors vary during training with finetuning}
    \label{fig:all_factor_finetuning}
\end{figure}

We also study the setting where all factors vary during training. In Figures \ref{fig:all_factor_linear} and \ref{fig:all_factor_finetuning} we show the generalization gaps when all factors vary for linear evaluation and finetuning.

\section{Class generalization}

In addition to the finetning results, we include here linear evaluation results for class generalization gaps \ref{linear eval_class_generalization}.

\begin{table}
\centering
\caption{Linear eval class generalization top-1 accuracy gaps: shows validation top-1 accuracy difference between classes (27 randomly selected) seen with diversity and those not.}
\label{linear eval_class_generalization}
\begin{adjustbox}{width=\textwidth}
\begin{tabular}{lrrrrrr}
\toprule
       model &  Position gap &  Pose gap &  Lighting color gap &  Size gap &  Background gap &  Average gap \\
\midrule
        CLIP &        -41.69 &    -46.51 &              -43.73 &    -41.51 &          -48.09 &       -44.31 \\
         MAE &         -3.32 &     -6.64 &               -7.69 &     -5.68 &          -14.61 &        -7.59 \\
  MLPMixer1k &        -39.07 &    -43.70 &              -43.42 &    -36.32 &          -47.79 &       -42.06 \\
 MLPMixer21k &        -43.07 &    -57.34 &              -42.90 &    -44.84 &          -52.09 &       -48.05 \\
 ResNet50-1k &        -44.97 &    -51.06 &              -43.84 &    -42.18 &          -55.67 &       -47.54 \\
ResNet50-21k &        -46.34 &    -56.89 &              -44.06 &    -47.26 &          -57.81 &       -50.47 \\
      ViT-1k &        -47.03 &    -58.69 &              -45.24 &    -48.67 &          -52.77 &       -50.48 \\
     ViT-21k &        -50.41 &    -56.94 &              -49.17 &    -49.76 &          -55.23 &       -52.30 \\
     iBot-1k &        -46.90 &    -61.71 &              -46.35 &    -51.56 &          -59.88 &       -53.28 \\
    iBot-21k &        -53.00 &    -64.94 &              -50.99 &    -56.05 &          -64.83 &       -57.96 \\
     Average &        -41.58 &    -50.44 &              -41.74 &    -42.38 &          -50.88 &       -45.40 \\
\bottomrule
\end{tabular}
 \end{adjustbox}
\end{table}

\section{Cross factor effects}

\begin{figure}[h]
\includegraphics[width=\textwidth]{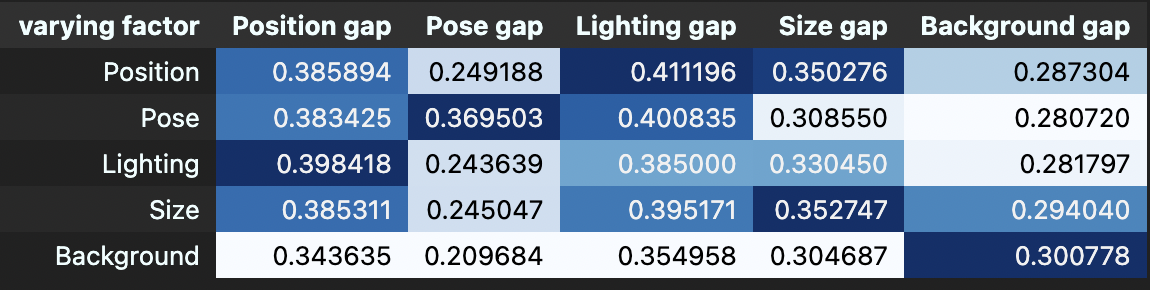}
    \caption{Spill over effects: shows the average slope across models when a given factor varies during linear evaluation}
    \label{appfig:spill_over_effect_slope_linear_eval}
\end{figure}

\begin{figure}[h]
\includegraphics[width=\textwidth]{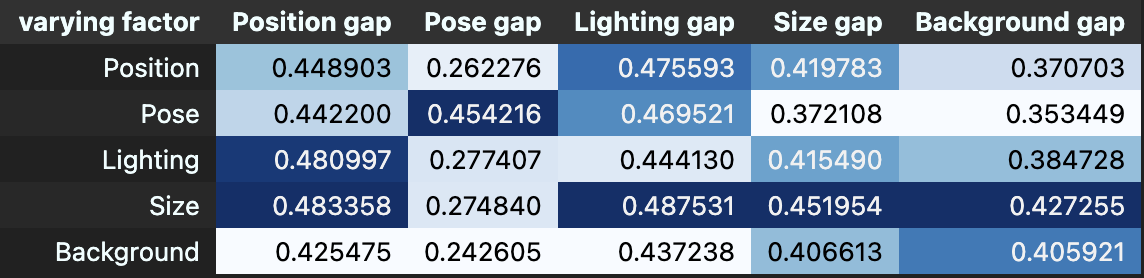}
    \caption{Spill over effects: shows the average slope across models when a given factor varies for finetuning}
        \label{appfig:spill_over_effect_slope_finetuning}
\end{figure}

\begin{figure}[h]
\includegraphics[width=\textwidth]{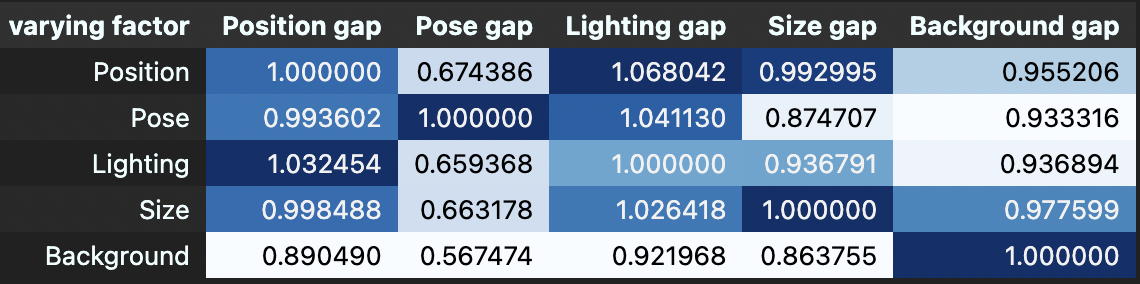}
    \caption{Normalized Spill over effects: shows the average slope across models when a given factor varies during linear evaluation. Normalization is across rows by dividing the diagonal value to isolate how much more a given spill-over effect than the intended.}
    \label{appfig:spill_over_effect_slope_linear_eval_normalized}
\end{figure}

\begin{figure}[h]
\includegraphics[width=\textwidth]{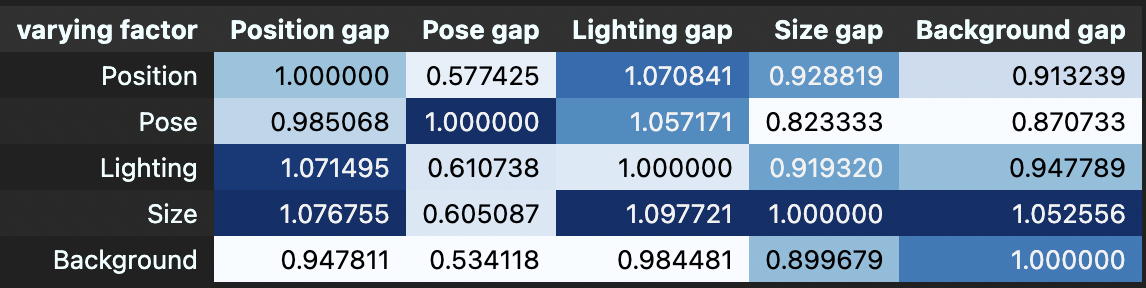}
    \caption{Normalized Spill over effects: shows the average slope across models when a given factor varies for finetuning.  Normalization is across rows by dividing the diagonal value to isolate how much more a given spill-over effect than the intended.}
    \label{appfig:spill_over_effect_slope_finetuning_normalized}
\end{figure}

We study the effect of varying a factor on the generalization gaps of other factors. 
In Figures \ref{appfig:spill_over_effect_slope_linear_eval} and \ref{appfig:spill_over_effect_slope_finetuning}
we show the slopes of the generalization gaps as the number of varying training instances increases during training.
We see how varying one factor can also close the robustenss gap of other factors. We also show normalized versions of these plots 
in \ref{appfig:spill_over_effect_slope_linear_eval_normalized} and \ref{appfig:spill_over_effect_slope_finetuning_normalized}.

\section{Effect of class similarity on models' ability to generalize variation across classes}

\begin{figure}[h]
\includegraphics[width=\textwidth]{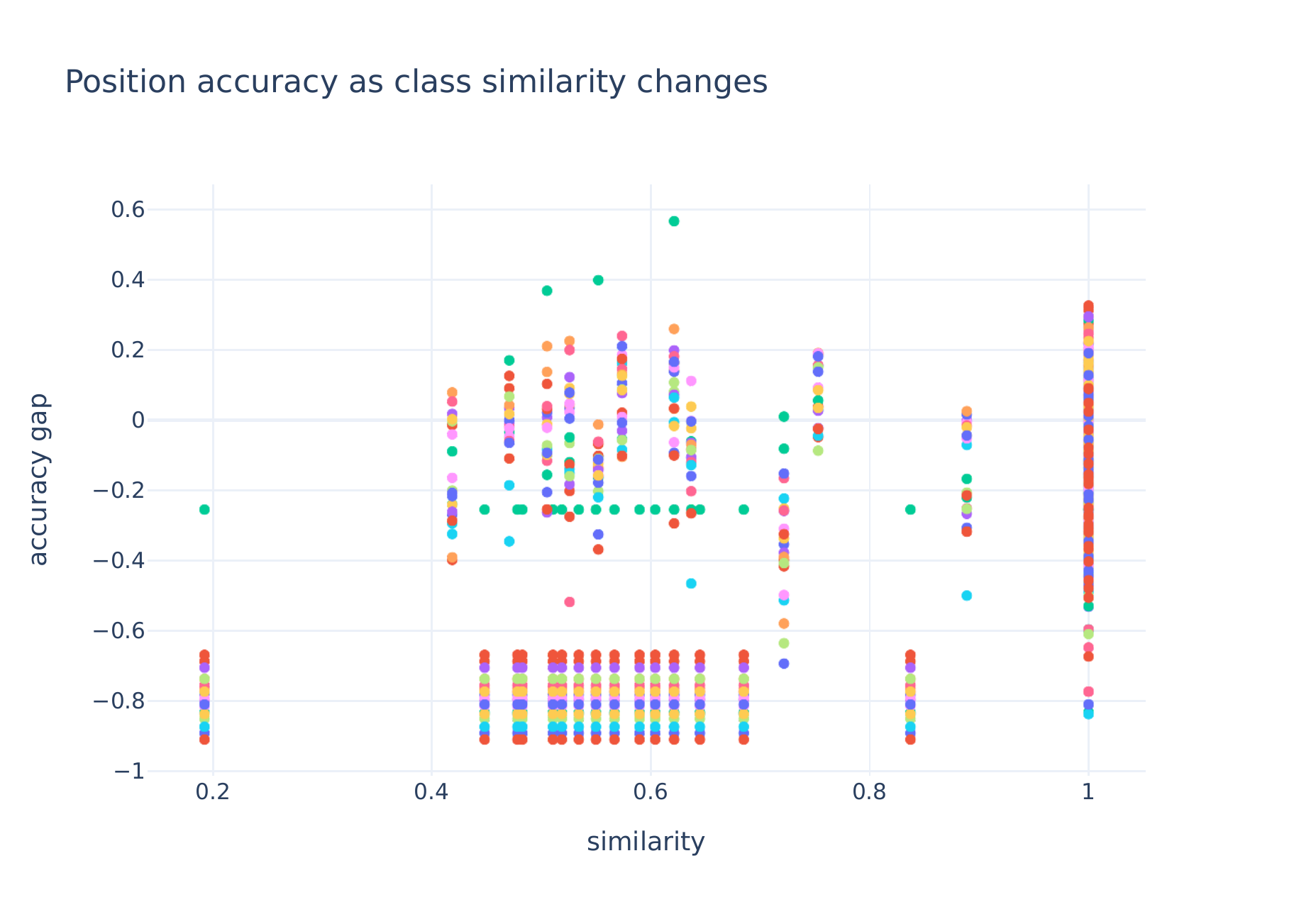}
    \caption{Position gap as class similarity to nearest neighbor increases to classes seen varying during training.}
    \label{appfig:class_sim_position}
\end{figure}

We study the effect of class similarity by measuring the generalization gaps per class for each factor relative to the 
class's similarity to the nearest class seen varying during training.
If models' are able to generalize variation across classes, we might expect models generalize variation better 
when the class is similar to one seen varying during training. In Figures \ref{appfig:class_sim_position},
\ref{appfig:class_sim_position}, \ref{appfig:class_sim_lighting}, and \ref{appfig:class_sim_scale}.

\begin{figure}[h]
\includegraphics[width=\textwidth]{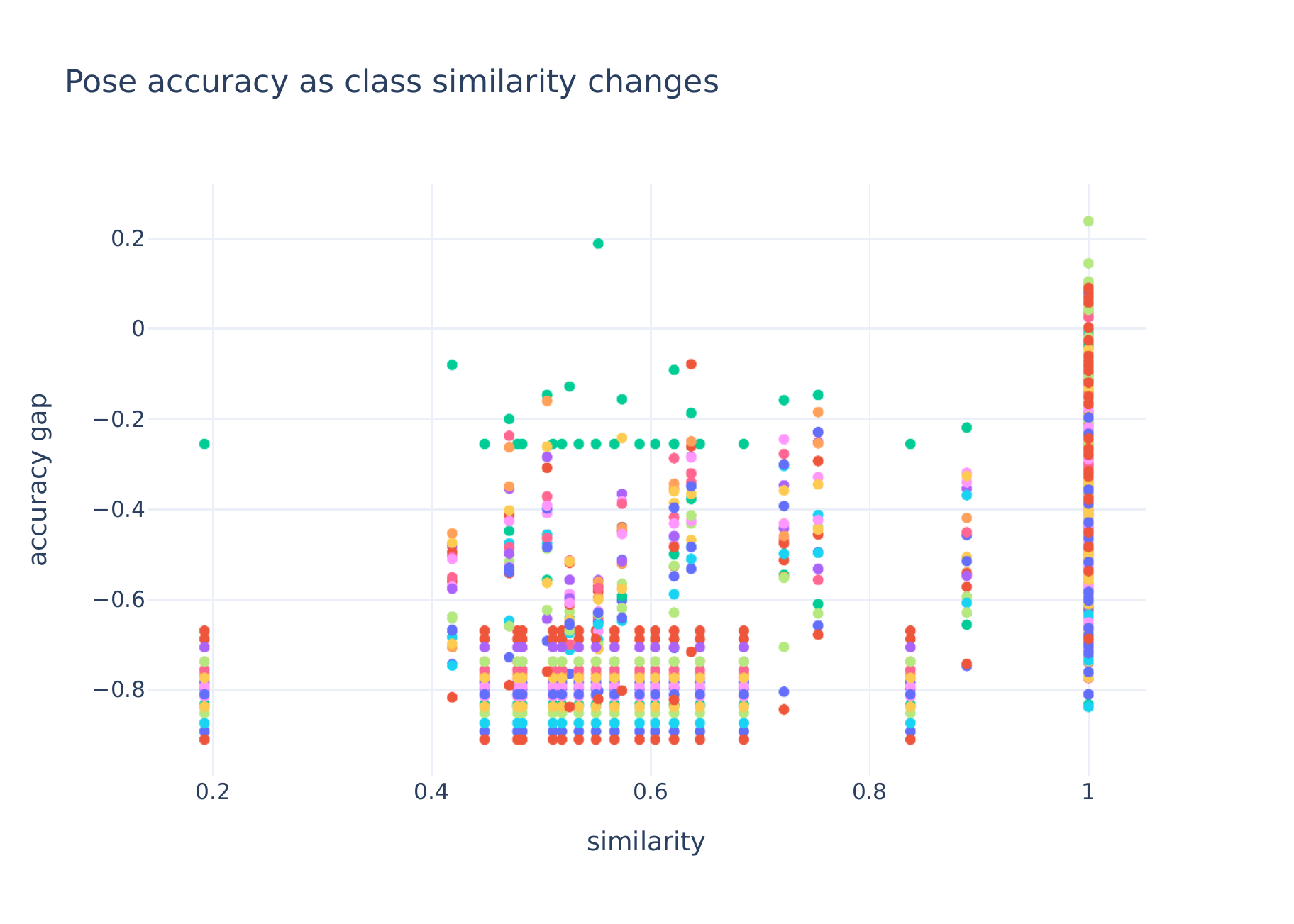}
    \caption{Pose gap as class similarity to nearest neighbor increases to classes seen varying during training.}
    \label{appfig:class_sim_pose}
\end{figure}

\begin{figure}[h]
\includegraphics[width=\textwidth]{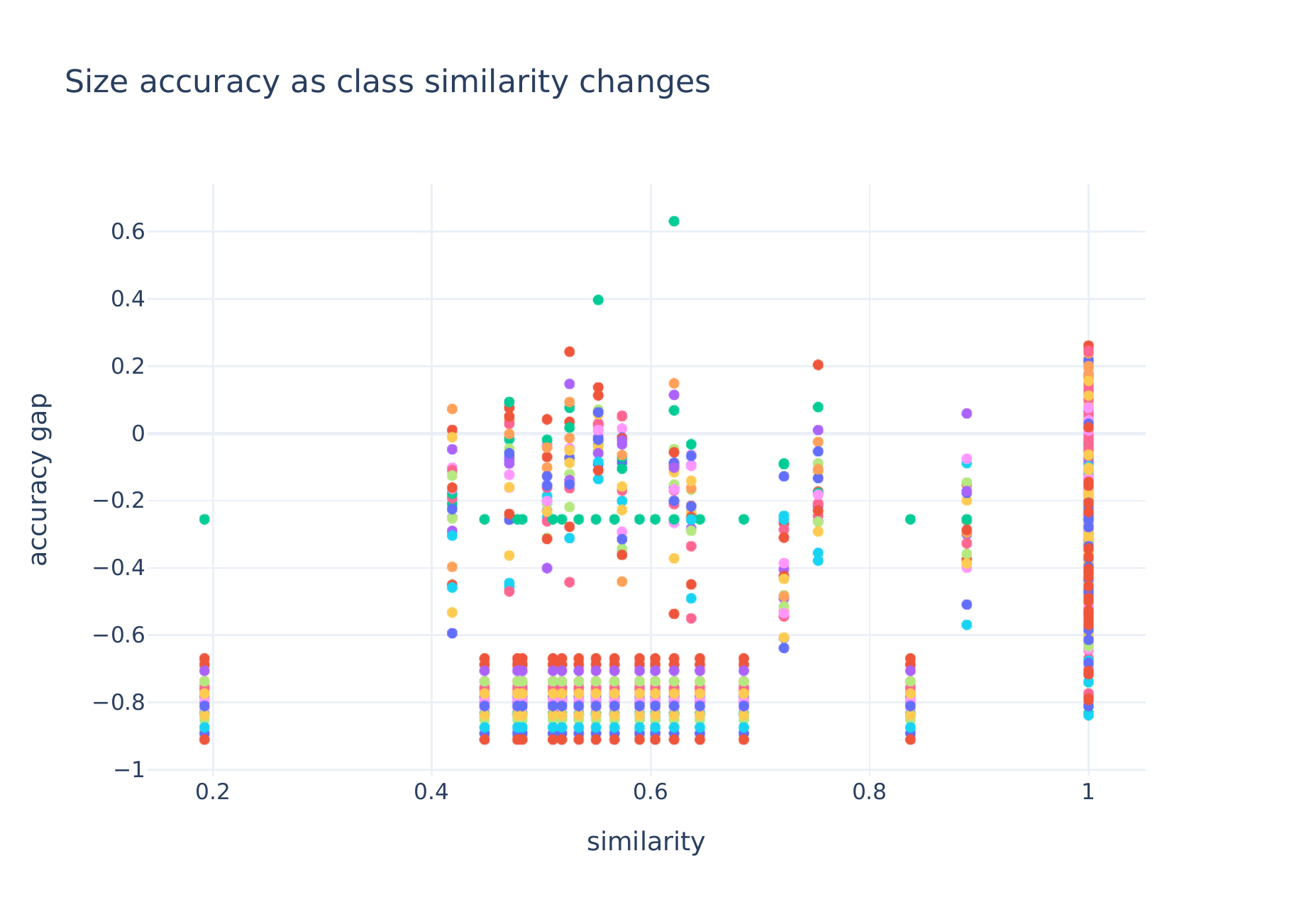}
    \caption{Scale gap as class similarity to nearest neighbor increases to classes seen varying during training.}
    \label{appfig:class_sim_scale}
\end{figure}

\begin{figure}[h]
\includegraphics[width=\textwidth]{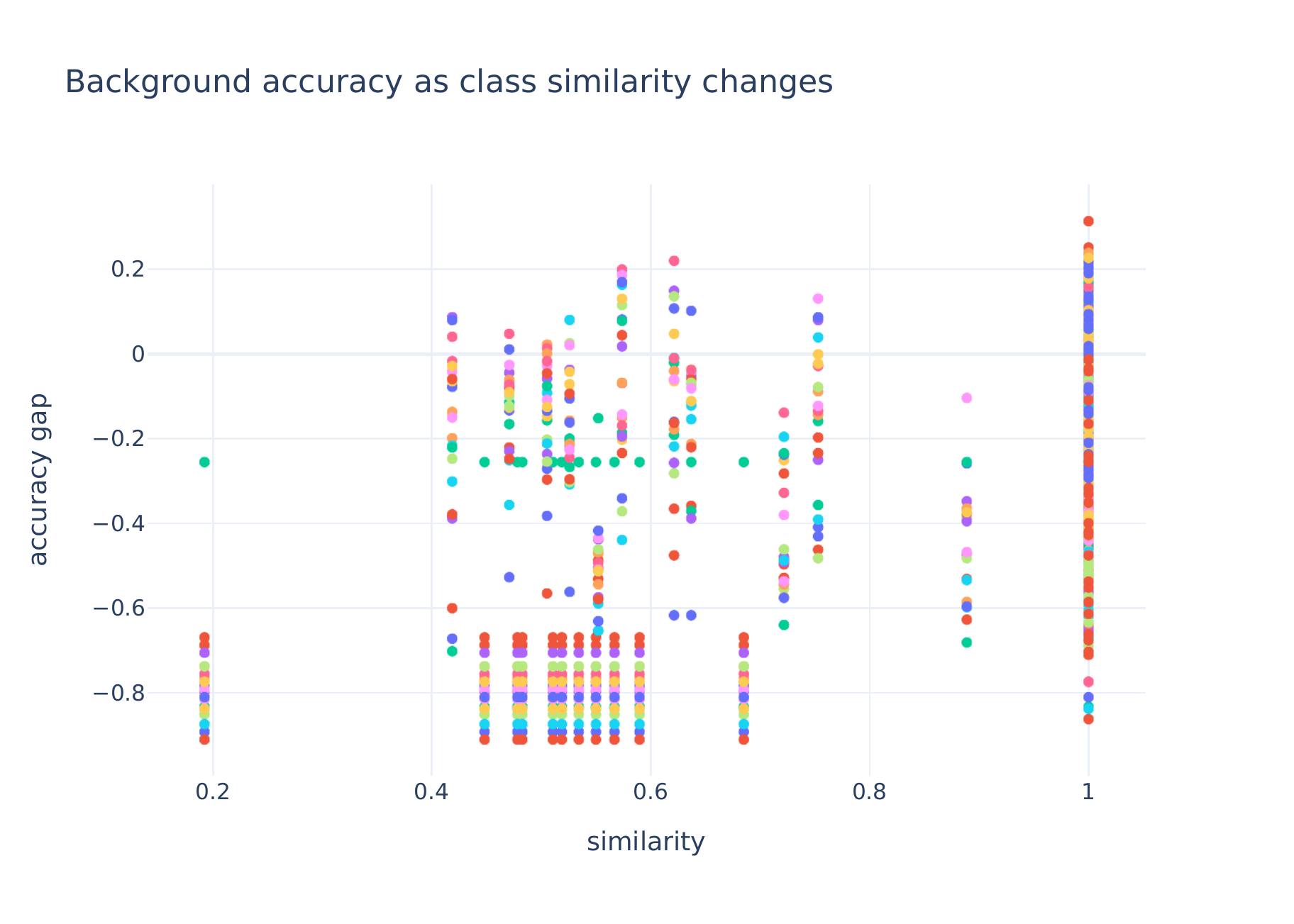}
    \caption{Background gap as class similarity to nearest neighbor increases to classes seen varying during training.}
\end{figure}

\begin{figure}[h]
\includegraphics[width=\textwidth]{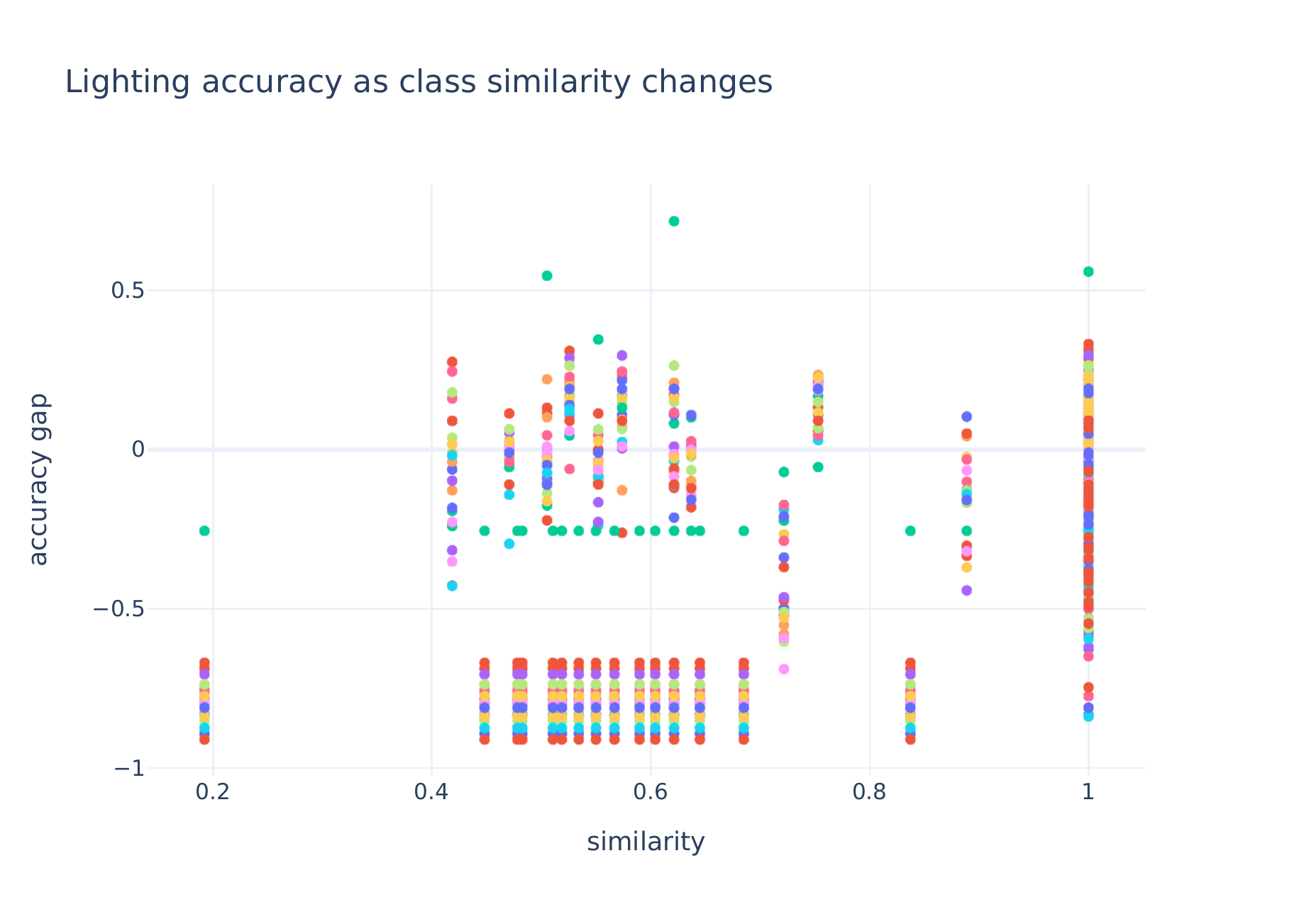}
    \caption{Lighting color gap as class similarity to nearest neighbor increases to classes seen varying during training.}
    \label{appfig:class_sim_lighting}
\end{figure}

\section{Experiments details}

Tables \ref{linear eval_canonical} and \ref{finetuning_canonical} show results for the best  after 10k steps of training with adam on 6 log scale learning rates (1e-2 to 1e-6) cross validated on canonical top-1 accuracy for validation images.

\end{document}